\def\year{2022}\relax
\documentclass[letterpaper]{article} % DO NOT CHANGE THIS
\usepackage{aaai22}  % DO NOT CHANGE THIS
\usepackage{times}  % DO NOT CHANGE THIS
\usepackage{helvet}  % DO NOT CHANGE THIS
\usepackage{courier}  % DO NOT CHANGE THIS
\usepackage[hyphens]{url}  % DO NOT CHANGE THIS
\usepackage{graphicx} % DO NOT CHANGE THIS
\usepackage{amssymb}
\usepackage{multirow}
\usepackage{aaai22}  % DO NOT CHANGE THIS
\usepackage{times}  % DO NOT CHANGE THIS
\usepackage{helvet}  % DO NOT CHANGE THIS
\usepackage{courier}  % DO NOT CHANGE THIS
\usepackage[hyphens]{url}  % DO NOT CHANGE THIS
\usepackage{graphicx}
\usepackage{epstopdf}
\usepackage{amsmath}
\usepackage{lineno}
\usepackage{bbding}
\usepackage{booktabs}

\usepackage{hyperref}
\usepackage{natbib}  % DO NOT CHANGE THIS AND DO NOT ADD ANY OPTIONS TO IT
\usepackage{caption} % DO NOT CHANGE THIS AND DO NOT ADD ANY OPTIONS TO IT
\DeclareCaptionStyle{ruled}{labelfont=normalfont,labelsep=colon,strut=off} % DO NOT CHANGE THIS
\usepackage{algorithm}
\usepackage{algorithmic}
\usepackage{bbding}
\usepackage{pifont}
\usepackage{wasysym}
\usepackage{amssymb}% http://ctan.org/pkg/amssymb
\usepackage{enumitem,amssymb}
\newlist{todolist}{itemize}{2}
\setlist[todolist]{label=$\square$}
\usepackage{pifont}
\usepackage{newfloat}
\usepackage{listings}
\floatstyle{ruled}
\newfloat{listing}{tb}{lst}{}
\floatname{listing}{Listing}
\title{Online-updated High-order Collaborative Networks for Single Image Deraining}
\author {
    Cong Wang\textsuperscript{\rm 1},
    Jinshan Pan\textsuperscript{\rm 2},
    Xiao-Ming Wu\textsuperscript{\rm 1}\thanks{\rm Corresponding author}
}
\affiliations {
    \textsuperscript{\rm 1} Department of Computing, The Hong Kong Polytechnic University\\
    \textsuperscript{\rm 2} School of Computer Science and Engineering, Nanjing University of Science and Technology\\
    supercong94@gmail.com, sdluran@gmail.com, xiao-ming.wu@polyu.edu.hk
}

\usepackage{bibentry}
\begin{document}
% \pagewiselinenumbers
% \switchlinenumbers
\maketitle
\begin{abstract}
Single image deraining is an important and challenging task for some downstream artificial intelligence applications such as video surveillance and self-driving systems.
Most of the existing deep-learning-based methods constrain the network to generate derained images but few of them explore features from intermediate layers, different levels, and different modules which are beneficial for rain streaks removal.
In this paper, we propose a high-order collaborative network with multi-scale compact constraints and a bidirectional scale-content similarity mining module to exploit features from deep networks externally and internally for rain streaks removal.
Externally, we design a deraining framework with three sub-networks trained in a collaborative manner, where the bottom network transmits intermediate features to the middle network which also receives shallower rainy features from the top network and sends back features to the bottom network.
Internally, we enforce multi-scale compact constraints on the intermediate layers of deep networks to learn useful features via a Laplacian pyramid.
Further, we develop a bidirectional scale-content similarity mining module to explore features at different scales in a down-to-up and up-to-down manner.
To improve the model performance on real-world images, we propose an online-update learning approach, which uses real-world rainy images to fine-tune the network and update the deraining results in a self-supervised manner.
Extensive experiments demonstrate that our proposed method performs favorably against eleven state-of-the-art methods on five public synthetic datasets and one real-world dataset.
The source code will be available at \url{https://supercong94.wixsite.com/supercong94}.
\end{abstract}

\section{Introduction}
% Outdoor images and videos taken in bad
% weather conditions such as rain, haze, and snow, often have limited visibility, which degrades the performance of various applications such as image retrieval and video surveillance.
% Hence, it is crucial to restore the degraded images and videos to increase the visibility of the scene
% and improve the performance of these applications.
% In this paper, we mainly focus on the rainy condition in a single image and propose an effective method for image deraining. %algorithm.
% Clear images play a significant role for some artificial intelligence applications as these images can provide reliable signals.
% However, images captured in rainy conditions may make these applications fail to work due to vision degradation.
% Hence, removing rain streaks from a single image is an essential task.
Outdoor images taken in rainy conditions have limited visibility, which degrades the performance of various applications such as video surveillance and self-driving systems.
Hence, it is essential to recover the degraded images to improve scene visibility and the performance of downstream applications. This paper considers single image deraining.

%solve the single image deraining problem by proposing an effective method.
\begin{figure}[!t]
\centering
\begin{tabular}{c}
\includegraphics[width = 0.99\linewidth]{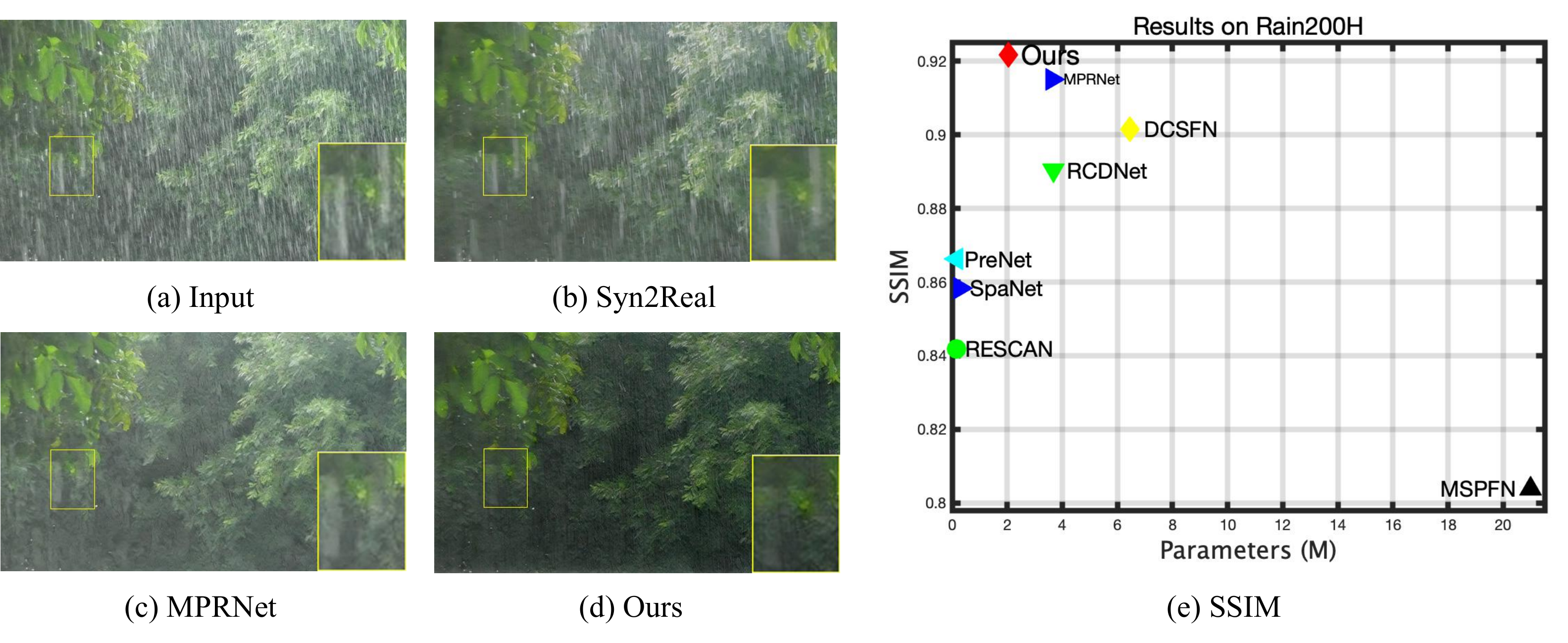}
\end{tabular}
% \vspace{-2mm}
\caption{Results on the real-world image (b-d) and the synthetic Rain200H dataset (e). Our proposed method generates a better deraining result on the real-world image and achieves the best performance-parameter trade-off on the synthetic dataset.
}
\label{fig:introduction}
% \vspace{-3.5mm}
\end{figure}

A rainy image $O$ can be modeled as a linear combination of a rain-free image $B$ and a rain streaks image $R$:
\begin{equation}
O = B + R.
\label{eq:rainy formation}
\end{equation}
Many deraining approaches have been developed based on this simple rainy image model, including prior-based methods and deep-learning-based methods. Prior-based methods usually explore empirical statistical properties of rain streaks and rain-free images, such as image decomposition~\cite{derain_id_kang}, sparse coding~\cite{derain_dsc_luo,derain_zhang_Sparse_and_Low-Rank}, low-rank representation~\cite{derain_lowrank}, and Gaussian mixture model~\cite{derain_lp_li}.
%
%Although these methods can achieve decent deraining results to some extent, the priors are based on empirical statistical observations and do not hold when real-world complex rainy conditions deviate from the simplified assumptions.
%Although these methods can achieve decent deraining results to some extent,
However, since the priors are based on empirical statistical observations, they do not hold when real-world complex rainy conditions deviate from the simplified assumptions.
Recent years have witnessed the successful application of deep learning methods to image deraining~\cite{derain_ddn_fu,derain_jorder_yang,derain_rescan_li,derain_zhang_did,dualcnn-pan,spic-wang-derain,cvpr20_jiang_mspfn,mm20_wang_dcsfn,derain_mprnet_cvpr21,icassp-wang-derain,PhysicsGAN-pan}.
%and achieved significant progress.
%
%Most of these state-of-the-art deep-learning-based methods develop various kinds of networks based on joint learning
Most of these methods develop joint learning networks~\cite{derain_zhang_did,derain_cvpr19_hu,cvpr20_jiang_mspfn} or explore multi-scale architectures~\cite{mm20_wang_dcsfn,icme2020_zhu_phy,derain_mprnet_cvpr21} for image deraining.
The methods based on joint learning networks include density-guided multi-stream networks~\cite{derain_zhang_did} and progressive networks motivated by patch similarity and guided by multi-scale architectures~\cite{cvpr20_jiang_mspfn}.
While these methods are able to remove rain streaks with the guiding networks, they do not perform well when the guiding process is not accurately estimated.
In addition, although multi-scale architectures have been demonstrated to be effective, existing methods usually adopt a straightforward way to fuse features from different scales and do not explore the properties of multi-scale features.
Last but not least, existing methods commonly use synthetic datasets to train the deep models due to the lack of well-constructed real-world datasets. However, the gap between synthetic and real data limits the performance of these methods in real-world applications.

Due to the difficulty of single image deraining, how to fully utilize convolutional features from both shallow and deep layers of a deep model and explore multi-scale features are important for rain streaks removal. In addition, it is essential to develop an effective algorithm to improve the deraining performance on real-world images.

To this end, we propose a high-order collaborative network with multi-scale compact constraints and a bidirectional scale-content mining module to remove rain streaks.
The high-order collaborative design allows exploring features from the shallower and deeper layers of different sub-networks collaboratively, while the multi-scale compact constraints are used to effectively learn features from intermediate convolutional layers and the bidirectional scale-content mining module is embedded in an encoder-decoder network to explore features of different scales.

Specifically, the high-order collaborative design contains three sub-networks (bottom, middle, and up) which are learned in a collaborative manner, where the bottom network transmits intermediate features to the middle network which also receives shallower rainy features from the top network and sends back features to the bottom network. The middle and top networks provide deep supervision for the bottom network to better learn and transmit shallow rainy features to higher layers to facilitate deraining. To generate more useful features from a deep network internally, we propose to enforce multi-scale compact constraints on the intermediate layers to learn better features with Laplacian pyramid images.
In addition, we develop a bidirectional scale-content similarity mining module in a down-to-up and up-to-down manner to capture long-range dependencies between features at different scales, which is embedded in an encoder-decoder architecture to explore useful features.

Finally, to improve the model performance on real-world images, we propose a simple yet effective online-update learning approach, to fine-tune the model trained on synthetic datasets using real-world data in a self-supervised manner with a KL-Divergence loss function. The proposed network design and online-update learning approach enable our model to achieve state-of-the-art deraining performance, especially on real-world images, as illustrated in Fig.~\ref{fig:introduction}.

The main contributions of this paper include:
\begin{itemize}
\item We propose a collaborative deraining framework with multi-scale compact constraints to control the learning process in an external and internal manner, with a new bidirectional scale-content similarity mining module to adaptively learn richer feature representations.

\item We present a simple yet effective online-update learning approach to fine-tune the model trained on synthetic datasets to adapt to real rainy conditions in a self-supervised manner for real-world image deraining.

\item We conduct extensive experiments and ablation studies to evaluate the proposed method. The results demonstrate that our method performs favorably against state-of-the-art methods with fewer parameters on both synthetic and real-world datasets.
\end{itemize}
\section{Related Work}
In this section, we briefly review recent works on image deraining, which are based on deep learning, as well as some image restoration methods based on similarity mining.
\subsection{Single Image Deraining}
%
% As mentioned above, there are two solutions for single image deraining: prior-based methods and deep learning-based methods.

% Prior-based methods usually design some regularities according to the property of rain streams and rain-free images to restrain the solution spaces.
% Among them, image decomposition~\cite{derain_id_kang}, sparse coding~\cite{derain_dsc_luo}, low-rank presentation~\cite{derain_lowrank}, the combination between sparse and low-rank representation~\cite{derain_zhang_Sparse_and_Low-Rank}, Gaussian mixture model~\cite{derain_lp_li}, histogram of orientation of streaks~\cite{derain_prior}, and joint bi-layer optimization~\cite{derain_zhu_bilayer} are widely known.
% Alough prior-based models can achieve deraining performance to some extent, they are based on some assumptions which do not always hold on so that the deraining results are limited.
% Moreover these prior models need to solve a non-convex optimization problem, which is difficult to handle.

In recent years, deep-learning-based approaches have dominated the research of image deraining due to the strong representation learning ability of deep neural networks.
%and fast implementing process.
\citet{derain_ddn_fu} observe that high-frequency details provide more rain streaks details and less background interference and design a deep detail residual network to learn rain streaks.
Some deraining methods explore the properties of multi-scale images. \citet{mm20_wang_dcsfn} propose a cross-scale framework to fuse features of different scales from sub-networks. \citet{cvpr20_jiang_mspfn} design a multi-scale progressive fusion network to transmit and fuse small-scale features to the original scale based on the similarity of multi-scale images.
Recurrent networks are also used for deraining.
%introduced and designed into the deraining domain.
%~
\citet{derain_rescan_li} develop a recurrent squeeze-and-excitation network with dilation convolution to model channel context relation. \citet{derain_prenet_Ren_2019_CVPR} propose a progressive recurrent network to remove rain streaks stage by stage and analyze the effect of inputs, outputs, and loss functions for image deraining. Some non-local methods are based on attention mechanisms. \citet{derain_nledn_li} propose to embed a non-local module to an encoder-decoder framework to capture long-range feature dependency for improving representation learning. \citet{mm20_wang_jdnet} attempt to combine self-attention and scale-aggregation in a self-calibrated network.
Besides, some semi-supervised approaches have been proposed for removing rain streaks in real-world images.
For example, \citet{Derain-cvpr19-semi} develop a semi-supervised transfer learning approach, and \citet{cvpr20_syn2real} design a Gaussian-process-based model that learns on both synthetic and real data.

Different from the above works, our proposed method aims to explore inner structures and useful features of the deraining network in an external and internal manner to train the network for better rain streaks removal. We also propose an effective online-update learning approach to fine-tune the model trained with synthetic data on real data for real-world image deraining.
\subsection{Similarity Mining}
Similarity mining aims to find the most matched content at the feature level.
\citet{sr_cross_scale_mining_Mei}  study cross-scale feature mining by exploring the inherent properties of images for single image super-resolution.
\citet{mei2020pyramid} mine the similarity from coarser to finer levels in a multi-scale manner for image restoration. Different from these works that only explore single directional mining (from coarser to finer levels), we in this paper design a bidirectional scale-content similarity mining module to adaptively learn richer features for better deraining.
\begin{figure}[!t]
\begin{center}
\begin{tabular}{c}
\includegraphics[width = 0.99\linewidth]{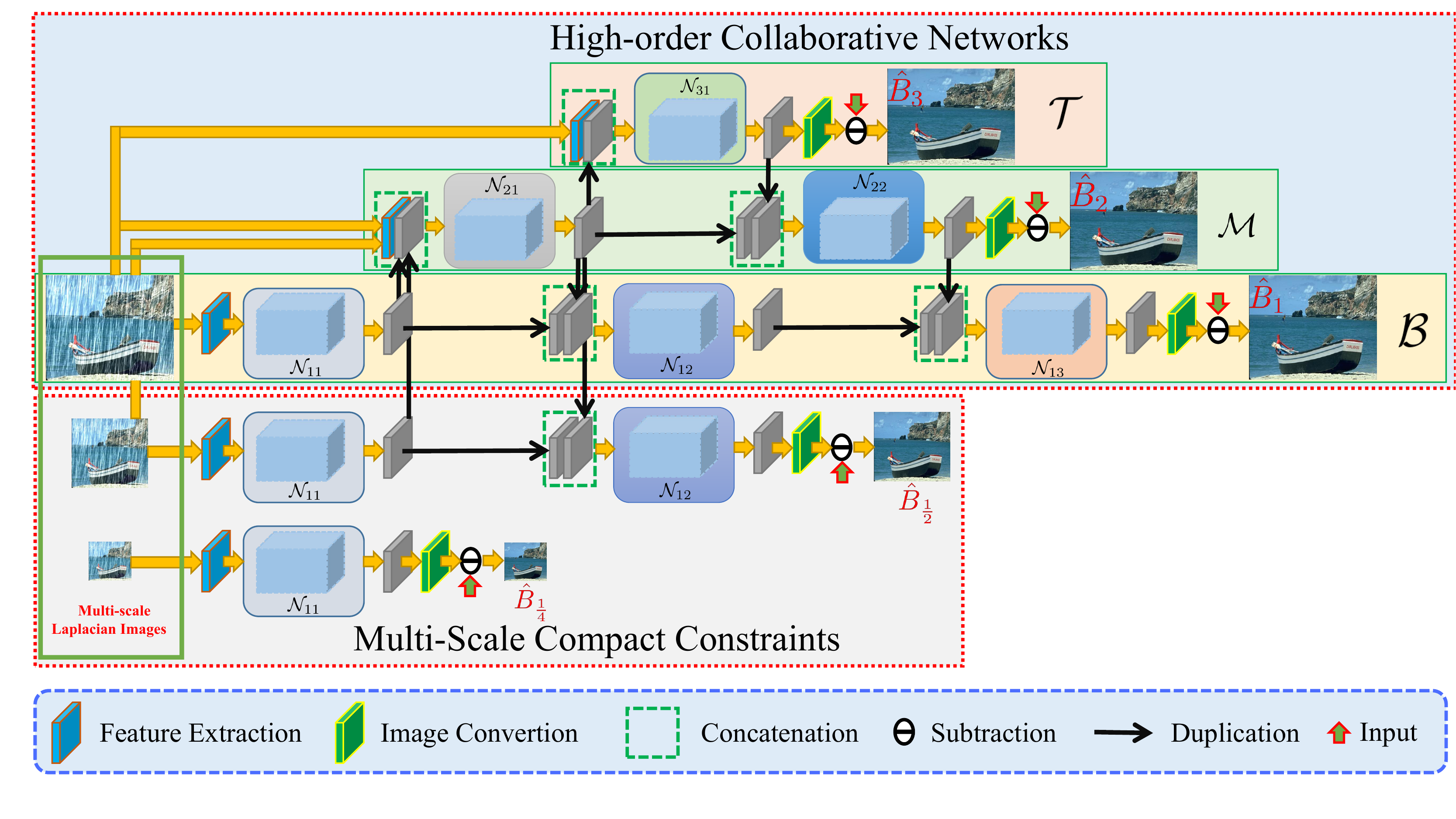}
\end{tabular}
\end{center}
% \vspace{-2mm}
\caption{Proposed high-order collaborative networks with multi-scale compact constraints.
Each network module $\mathcal{N}_{mn}$ has a same encoder-decoder structure with the bidirectional scale-content similarity minding module shown in Fig.~\ref{fig: encoder-decoder}.
}
\label{fig: overall}
\end{figure}
\section{Proposed Method}
In this section, we introduce each element of the proposed method including high-order collaborative network, multi-scale compact constraint, encoder-decoder structure with a bidirectional scale-content similarity mining module, supervised loss function, and online-update learning.
\subsection{High-order Collaborative Networks}\label{sec: High-order Collaborative Progressive Network}
How to exploit features from the intermediate layers and different modules in a collaborative manner is important for image restoration.
Here, we design an effective high-order collaborative network\footnote{We regard two-stream networks as collaborative networks and three-stream networks as high-order.} to assist each sub-network to learn features in an external manner for better image deraining.
The overall network structure of the high-order collaborative network is shown in Fig.~\ref{fig: overall}, which consists of three sub-networks with similar structure.
%Bottom-Middle-Top three-stream sub-networks and these sub-networks share similar structures.
%
The bottom ($\mathcal{B}$), middle ($\mathcal{M}$), and top ($\mathcal{T}$) sub-networks respectively contain three, two, and one encoder-decoder component with a bidirectional scale-content similarity mining module that will be explained in detail in Sec.~\ref{sec: Scale-content Similarity Mining}.
%

% The rainy image is input into $\mathcal{N}_{11}$ of $\mathcal{B}$ to first learn rainy features and the learned features are re-used by the subsequent modules $\mathcal{N}_{12}$ of $\mathcal{B}$ and $\mathcal{N}_{21}$ of $\mathcal{M}$ for further progressive rainy features re-usage and learning.
% Similarly, these features are further learned for sub-parts of different sub-networks in a collaborative manner.
%From this learning scheme,
The three sub-networks can transmit shallower and deeper rainy features to each other in a collaborative manner to improve the performance of the deraining network $\mathcal{B}$.
Note that $\mathcal{M}$ and $\mathcal{T}$ are supervised by ground truth, which can be regarded as a sort of deep supervision in an external manner such that $\mathcal{M}$ and $\mathcal{T}$ can transmit useful features to assist $\mathcal{B}$ to learn better features for deraining.
The three sub-networks are trained by:
\begin{equation}
\begin{array}{ll}
\mathcal{L}_{\text{collaborative}} = \sum_{i=1}^{3}\alpha_{i} \big(-SSIM(\hat{B}_{i}, B)\big),
\end{array}
\label{eq:n0}
\end{equation}
where $\{\hat{B}_{i}\}$ ($ i = 1, 2, 3$) denote the output of networks $\mathcal{B}$, $\mathcal{M}$, and $\mathcal{T}$ respectively, $B$ is the ground truth, and $\alpha_{i}$ are weight parameters.
\subsection{Multi-Scale Compact Constraints}\label{sec: Multi-scale Compact Constraint}
Although the high-order collaborative design enables the bottom sub-network $\mathcal{B}$ to exploit deeper and shallower features from the other two sub-networks, the features of each intermediate convolutional layer are learned without any constraints, making the solution space too large.
%that enlarge the learning range, limiting the deraining performance.
% \JS{the features in each convolutional layer of the deep models can be arbitrary and some of the learned features may not facilitate the image deraining task. CHECK THIS SENTENCE!!}
%
% To regularize the solution spaces, we design multi-scale compact constraints (MSCC) in an internal manner to push the $\mathcal{N}_{11}$, $\mathcal{N}_{12}$, $\mathcal{N}_{21}$ to learn more useful features that facilitate image deraining.

To regularize the solution space of intermediate convolutional layers, we design multi-scale compact constraints (MSCC) in an internal manner to enforce the network modules $\mathcal{N}_{11}$, $\mathcal{N}_{12}$, and $\mathcal{N}_{21}$, which are parts of the sub-networks $\mathcal{B}$ and $\mathcal{M}$, to learn more useful features to facilitate image deraining.
The multi-scale compact constraints are motivated by the multi-scale Laplacian images which can better model image structures than the original scale image. We use them to constrain the intermediate convolutional layers of the network.

We first obtain the Laplacian pyramid images and then use the scaled images to
%to the first two (i.e., $\mathcal{N}_{11}$ and $\mathcal{N}_{12}$) and one networks (i.e., $\mathcal{N}_{21}$) of $\mathcal{B}$ and $\mathcal{M}$
constrain the modules $\mathcal{N}_{11}, \mathcal{N}_{12},  \mathcal{N}_{21}$ as shown in Fig.~\ref{fig: overall}. Specifically,
%respectively as the same operations with the executing high-order collaborative networks.
%
the multi-scale compact constraints are enforced by:
\begin{equation}
\begin{array}{ll}
\mathcal{L}_{\text{mscc}} = \sum_{j=1}^{2}\beta_{j}\big(-SSIM(\hat{B}_{\frac{1}{2^{j}}}, B_{\frac{1}{2^{j}}})\big),
\end{array}
\label{eq:n0}
\end{equation}
where $\hat{B}_{\frac{1}{2^{j}}}$ are the output of intermediate layers as denoted in Fig.~\ref{fig: overall},
%$\frac{1}{2^{j}}$ scale compact constraint, while
$B_{\frac{1}{2^{j}}}$ are the corresponding $\frac{1}{2^{j}}$ scale Laplacian pyramid rain-free images, and
$\beta_{j}$ are weight parameters.
\subsection{Encoder and Decoder with BiSCSM}\label{sec: Scale-content Similarity Mining}
\begin{figure}[!t]
\begin{center}
\begin{tabular}{c}
\includegraphics[width = 0.99\linewidth]{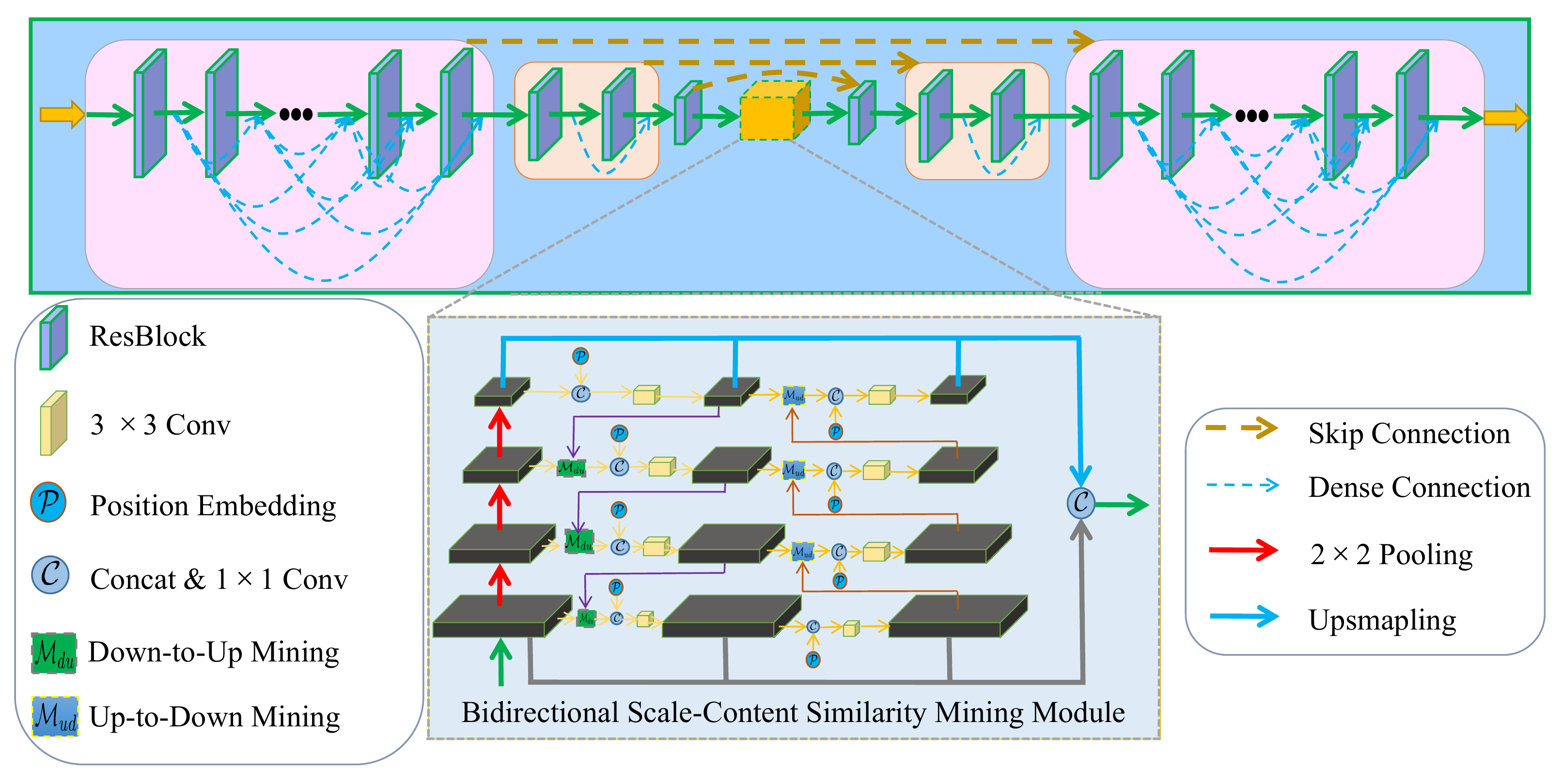}
\end{tabular}
\end{center}
% \vspace{-2mm}
\caption{Proposed encoder and decoder with a bidirectional scale-content similarity mining module. The Down-to-Up and Up-to-Down mining modules are shown in Fig.~\ref{fig: mining module}.
}
\label{fig: encoder-decoder}
% \vspace{-3mm}
\end{figure}

%
%To further make the network for better image deraining, we
Further, we develop a bidirectional scale-content similarity mining module (BiSCSM) to explore similar features from different scales, which
%The proposed BiSCSM
is motivated by \citet{Mei_sisr_mining,sr_cross_scale_mining_Mei}. The architecture is shown in Fig.~\ref{fig: encoder-decoder}.

\begin{figure}[!t]
\begin{center}
\begin{tabular}{cc}
\includegraphics[width = 0.499\linewidth]{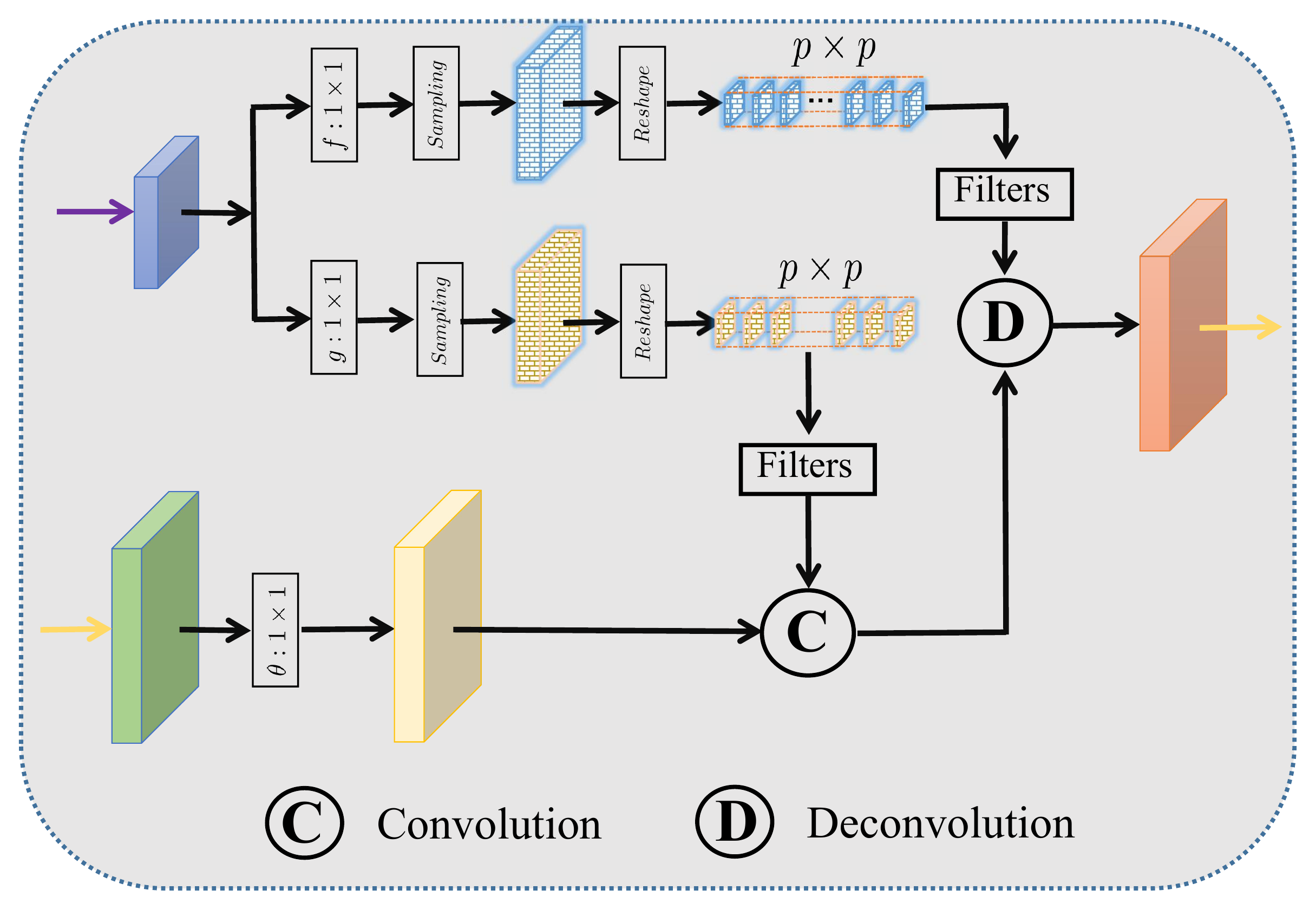} &\hspace{-4mm}
\includegraphics[width = 0.499\linewidth]{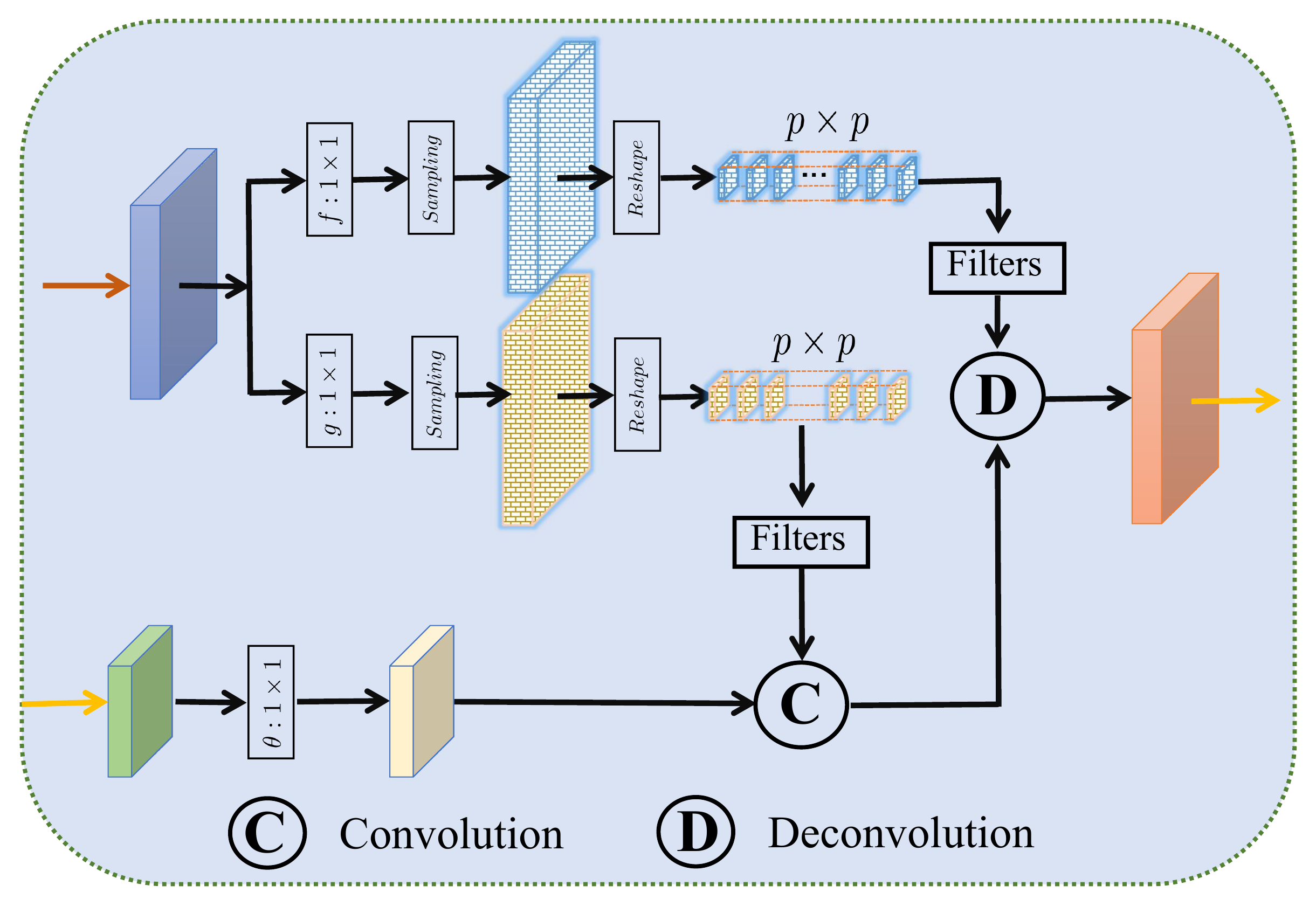}
\\
(a) Down-to-Up Mining&\hspace{-4mm} (b) Up-to-Down Mining
\end{tabular}
\end{center}
\vspace{-2mm}
\caption{The scale-content similarity mining modules.
}
\label{fig: mining module}
% \vspace{-3.5mm}
\end{figure}

The proposed BiSCSM contains Down-to-Up mining and Up-to-Down mining modules. The Up-to-Down mining module is defines as:
\begin{equation}
\begin{array}{ll}
y_{i,j}^{p \times p} = \frac{1}{\sigma(x,z)}\sum_{g,h} \phi(x^{p \times p}_{i,j},z^{p \times p}_{u,v}) \theta (x_{g,h}^{p \times p}) ,
\end{array}
\label{eq:n0}
\end{equation}
where $\phi(x^{p \times p}_{i,j},z^{p \times p}_{u,v}) = e^{(W_{f}x_{i,j}^{p \times p})^{T}(W_{g}z_{u,v}^{p \times p})}$,
$\theta = W_{\theta}x_{g,h}^{p \times p}$,
$\sigma = \sum_{u,v}\phi(x^{p \times p}_{i,j},z^{p \times p}_{u,v})$,
% $W$ denotes a learnable filter.
$y_{i,j}^{p \times p}$ is a $p \times p$ feature patch located at ($i$, $j$), and
$W_{f},W_{g},W_{\theta}$ are learnable filters.
The Down-to-Up mining module is defined similarly as shown in Fig.~\ref{fig: mining module}.
%in a similar way as the Down-to-Up mining module and their diagrams are shown in Fig.~\ref{fig: mining module}.

%Based on the above discussions,
Our proposed BiSCSM is different from the one in \citet{Mei_sisr_mining,sr_cross_scale_mining_Mei} which only mines patch similarity in a single direction. Our Down-to-Up mining and Up-to-Down mining modules enable capturing bidirectional similar content
%from larger-scale to smaller-scale and from smaller-scale to larger-scale
from large to small scales and from small to large scales to mine rich rainy features.
%can be mined richly as much as possible.
The BiSCSM is embedded in an encoder-decoder framework as in Fig.~\ref{fig: encoder-decoder} to learn useful features for rain streaks removal.
%to learn useful features
%that facilitate the rain streaks removal.
%
We also use positional embedding to encode relations among rain streak features.
%To further better learn the rain streaks relation, positional embedding is applied in the BiSCSM.
%
Finally, we fuse the learned features at difference scales.
% Different from \cite{Mei_sisr_mining} that they explore the exemplar mining in a single direction for the image super-resolution, content similarity also exists in the rainy features from bidirectional scale spaces, i.e., mining the content similarity in smaller scale from larger scale and in larger scale from smaller scale.
% Based on this, we design a bidirectional scale mining module in a down-up and up-down manner that can capture long-range dependencies between different scaled features at different levels, which is embedded in a encoder-decoder framework to discover richer rainy features.

\subsection{Supervised Loss Function}\label{Loss Function on Synthetic Datasets}
Based on the above network design, the overall loss function for training on synthetic datasets is:
\begin{equation}
\begin{array}{ll}
\mathcal{L}_{\text{synthetic}} = {\mathcal{L}_{\text{collaborative}}}+ {\mathcal{L}_{\text{mscc}}}.
\end{array}
\label{eq:n0}
\end{equation}

\subsection{Online-update Learning}\label{sec: Online Updating Scheme}
\begin{figure}[!t]
\begin{center}
\begin{tabular}{cc}
\includegraphics[width = 0.99\linewidth]{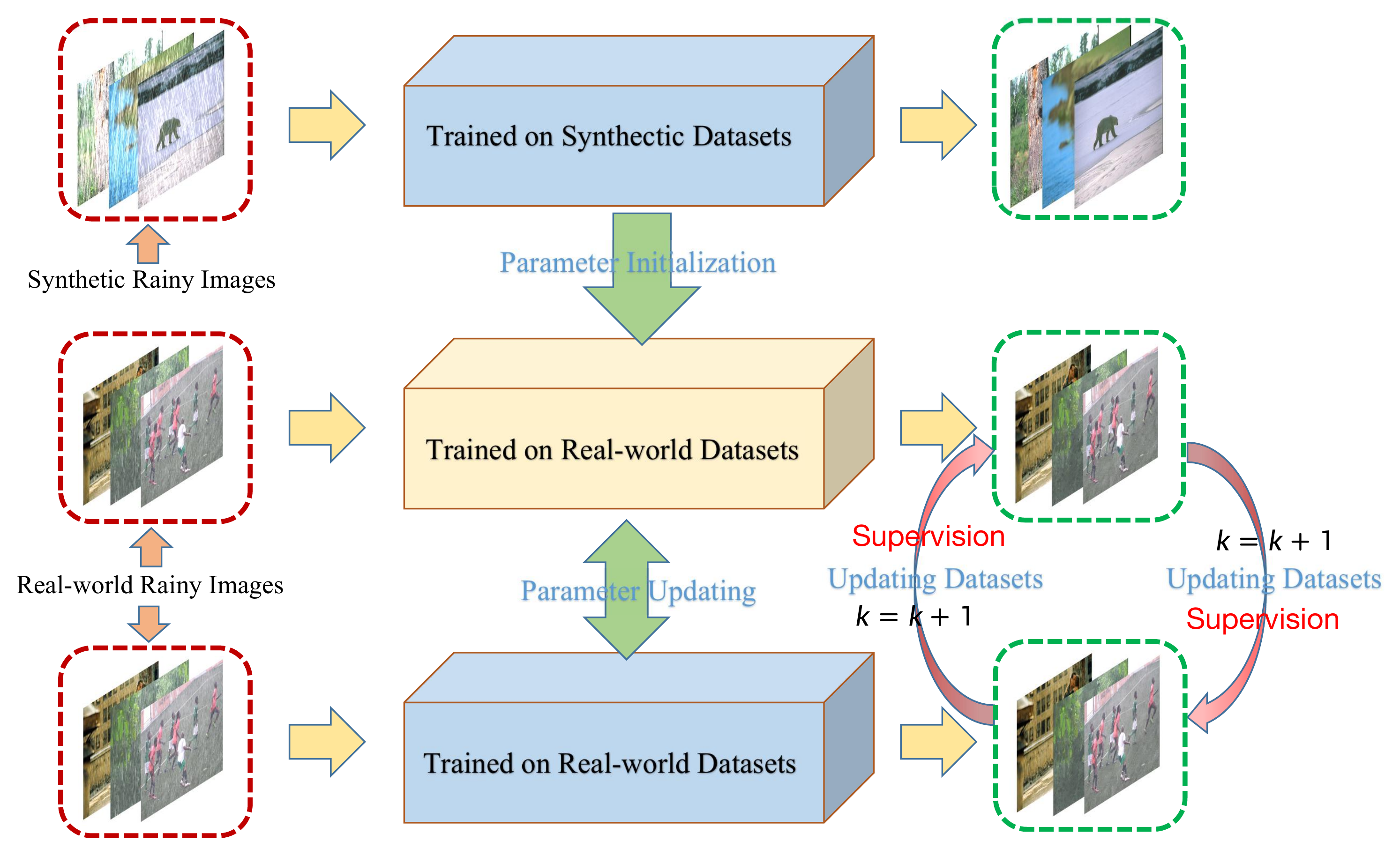}
\end{tabular}
\end{center}
% \vspace{-2mm}
\caption{Proposed online-update learning approach.
}
\label{fig: The illustration of online updating learning manner.}
% \vspace{-2mm}
\end{figure}

Since a deraining model trained on synthetic data normally does not generalize well to real-world images, we propose an online-update learning approach to fine-tune the model on real-world rainy images.
The key challenges are two-fold. The first is that there are no ground-truth images available for training.
%the supervision of the network training.
The second is how to keep the training stable and improve the model performance.
%that when fine-tuning the model on real-world data, we should maintain or improve the deraining performance of this model compared with the deraining output at the previous epoch.
To address these issues, we use the derained results of real-world images generated by the model trained on synthetic data as pseudo ground truth and update the pseudo ground truth at each epoch with the fine-tuned model. The loss function is based on Kullback-Leibler divergence and defined as:
\begin{equation}
\begin{array}{ll}
\mathcal{L}_{\text{real}} = \underbrace{||\hat{B}^{k}- \hat{B}^{k-1}||_{1}}_{\text{content term}} +
\underbrace{{\lambda KL_{Loss}(\hat{R}^{k}, \hat{R}^{k-1}_{\text{random}})}}_{\text{regularization term}},
\end{array}
\label{eq:real-world}
\end{equation}
where $\hat{B}^{k}$ and $\hat{B}^{k-1}$ are the deraining results of a real-world rainy image $O$ generated by the model $\mathcal{B}$ at epoch $k$ and $k-1$ respectively. Note that $\hat{B}^{0}$ is the initial deraining result produced by $\mathcal{B}$ after training on synthetic data.
$\hat{R}^{k}= O-\hat{B}^{k}$ is the estimated rain streaks of $O$. $\hat{R}^{k-1}_{\text{random}} = O_{\text{random}}-\hat{B}_{\text{random}}^{k-1}$ is the estimated rain streaks of a randomly selected real-world rainy image $O_{\text{random}}$ and $\hat{B}_{\text{random}}^{k-1}$ is the deraining result of $O_{\text{random}}$ at epoch $k-1$.
The first term of Eq.~(\ref{eq:real-world}) ensures the content of image background consistent between different epochs and keeps the training stable, and the second term is a regularization term that enforces similarity in rain streaks and improves the deraining performance as the training proceeds (see Fig.~\ref{fig: Results on fine-tune manner.}).
%and $\tilde{B}_{r}^{k-1}$ is the estimated rain-free images at epoch $k-1$.
%
The fine-tuning process is illustrated in
Fig.~\ref{fig: The illustration of online updating learning manner.} and described in Alg.~\ref{alg:B}.
%illustrates the training process of the proposed online-updated learning algorithm.
%
%The training process is described in Alg.~\ref{alg:B}.
%where the texts marked in bold denote our proposed online-updated learning scheme.
%The effectiveness of this learning approach is validated in Sec.~\ref{sec: experiments}.
%
%More analysis and discussion about the KL-Loss and the online-updated scheme are included in supplementary materials.

\begin{algorithm}[!t]
\caption{Online-update Learning on Real-world Data}
\label{alg:B}
\hspace*{0.02in}{Preparation:}
  The initial deraining result $\hat{B}^{0}$ of a real-world image $O$ generated by the model trained on synthetic data.
  %Alg.~\ref{alg:A};
   %Initialize the %parameters of real-world training
   %model with the parameters of the model trained on synthetic datasets.
\\
\hspace*{0.02in}{Input:}
  %Real-world training data
  $\{O, \hat{B}^{0}\}$.
\\
\hspace*{0.02in}{Output:}
Derained image \{$\hat{B}$\}.

\begin{algorithmic}[1]
\STATE {While $k$ $\leq$  $\mathrm{Epoch}_{\mathrm{Real}}$ ~do:}
\\
\STATE {\quad \quad Randomly crop training image pairs \{$O, \hat{B}^{k-1}$\}}
\\
\STATE {\quad \quad Randomly select a real-world rainy image $O_{\text{random}}$ and obtain $\hat{R}_{\text{random}}^{k-1} = O_{\text{random}}-\hat{B}_{\text{random}}^{k-1}$}
\\
\STATE {\quad \quad Update the deraining model by Eq.~(\ref{eq:real-world}})
\\
\STATE {\quad \quad Output the current deraining result: \{$\hat{B}^{k}$\} }
\STATE {\quad \quad Update the pseudo ground truth:\{$\hat{B}^{k-1}$\} $\gets$\{$\hat{B}^{k}$\}}
\STATE {\quad \quad $k$ $\gets$ $k+1$}
\STATE {End while}
\end{algorithmic}
\end{algorithm}
\begin{table*}[!t]

% The best and second best results are marked in {\color{red}{red}} and {\color{blue}{blue}} respectively.}
\centering
\scalebox{0.865}{
\begin{tabular}{l|cccccccccccccccccccccc}
\hline
\multirow{2}{*}{Methods}
& \multicolumn{2}{c}{Rain200H}
& \multicolumn{2}{c}{Rain200L}
& \multicolumn{2}{c}{Rain1200}
& \multicolumn{2}{c}{Rain1400}
& \multicolumn{2}{c}{Rain12}
& \multirow{2}{*}{\# Param}
\\
& PSNR~$\uparrow$ & SSIM~$\uparrow$& PSNR~$\uparrow$ & SSIM~$\uparrow$& PSNR~$\uparrow$ & SSIM~$\uparrow$& PSNR~$\uparrow$ & SSIM~$\uparrow$& PSNR~$\uparrow$ & SSIM~$\uparrow$     \\
\hline
RESCAN (ECCV'18) & 26.661&0.8419 &36.993&0.9788&32.127&0.9028&30.969&0.9117&32.965& 0.9545 &0.15M
\\
NLEDN (MM'18) &27.315&0.8904 &36.487&0.9792&32.473&0.9198&31.014&0.9206 &33.028& 0.9615&1.01M
\\
% ReHEN (MM'19) &25.484 & 0.8584 & 36.075 &0.9774 & 27.099& 0.8082& 29.000& 0.8891&33.217&  0.9546& 0.30M
% \\
SSIR (CVPR'19) &14.420&0.4501&23.476&0.8026&24.427&0.7713&25.772&0.8224&24.138&0.7768&0.06M
\\
PreNet (CVPR'19) & 27.525&0.8663 &34.266&0.9660&30.456&0.8702 &30.984&0.9156 &35.095&0.9400&0.17M
\\
SpaNet (CVPR'19) &25.484 & 0.8584& 36.075 &0.9774 & 27.099& 0.8082& 29.000& 0.8891&33.217&  0.9546&0.28M
\\
DCSFN (MM'20)&28.469&  0.9016 &37.847&  0.9842  &32.275& 0.9228&31.493& 0.9279 &35.803& 0.9683&6.45M
\\
MSPFN (CVPR'20) &25.553& 0.8039 & 30.367& 0.9219 &30.382&0.8860 &31.514& 0.9203 &34.253& 0.9469&21.00M
\\
DRDNet (CVPR'20) &15.102& 0.5028 &37.465&0.9806 &28.386&0.8574 &28.359&0.8574 &25.435& 0.7550 &5.23M
\\
RCDNet (CVPR'20) &28.698& 0.8904 & 38.400 &0.9841 &32.273& 0.9111& 31.016  &0.9164& 31.038  &0.9069 &3.67M
\\
Syn2Real (CVPR'20)&14.495 &  0.4021 &31.035 & 0.9365 &28.812 & 0.8400&28.582 &  0.8586&31.038  &  0.9069 &2.62M
\\
MPRNet (CVPR'21)&29.949 & 0.9151&36.610&0.9785&  33.655 &0.9310  & 32.257 & 0.9325 & 36.578 &0.9696 &3.64M
\\
% DualGCN (AAAI'21) &29.758& 0.9026&39.415 &0.9878&32.033& 0.9163&30.567& 0.9148 &35.805 &0.9687 & 2.73M
% \\
\hline
Ours &\textbf{29.985}&\textbf{0.9218}&\textbf{39.284} & \textbf{0.9875}&\textbf{33.718}& \textbf{0.9327}&\textbf{32.617}& \textbf{0.9334} &\textbf{36.851} &\textbf{0.9714} & 2.04M
\\
\hline
\end{tabular}}
% \vspace{-2mm}
\caption{Quantitative results on five synthetic datasets.
The best results are marked in \textbf{bold}.
$\uparrow$ denotes higher is better.
}
\label{tab: the results in synthetic datasets}
% \vspace{-4mm}
\end{table*}
\begin{figure*}[!t]%\footnotesize
\begin{center}
\begin{tabular}{ccccccccc}
\includegraphics[width = 0.104\linewidth]{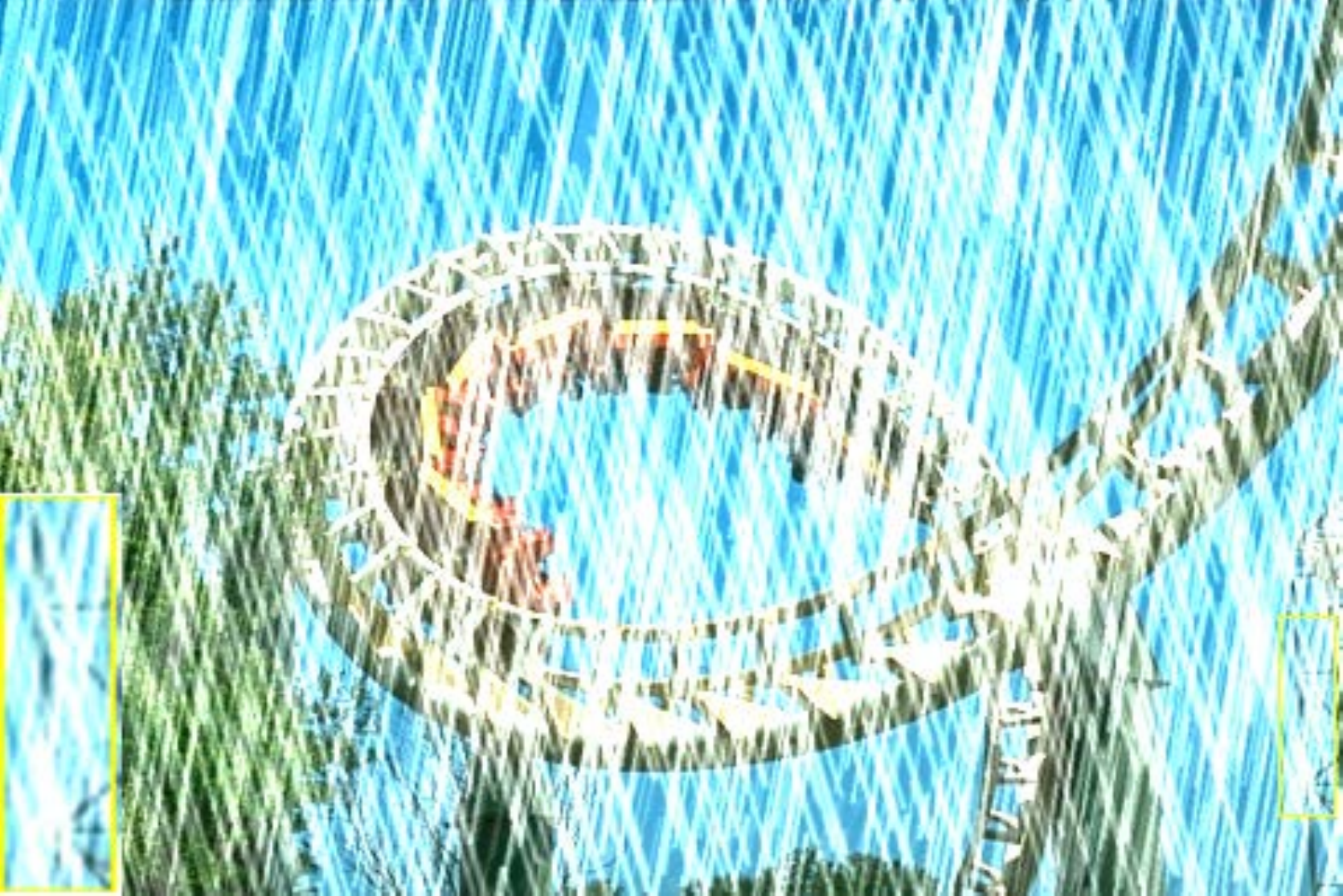} &\hspace{-4mm}
\includegraphics[width = 0.104\linewidth]{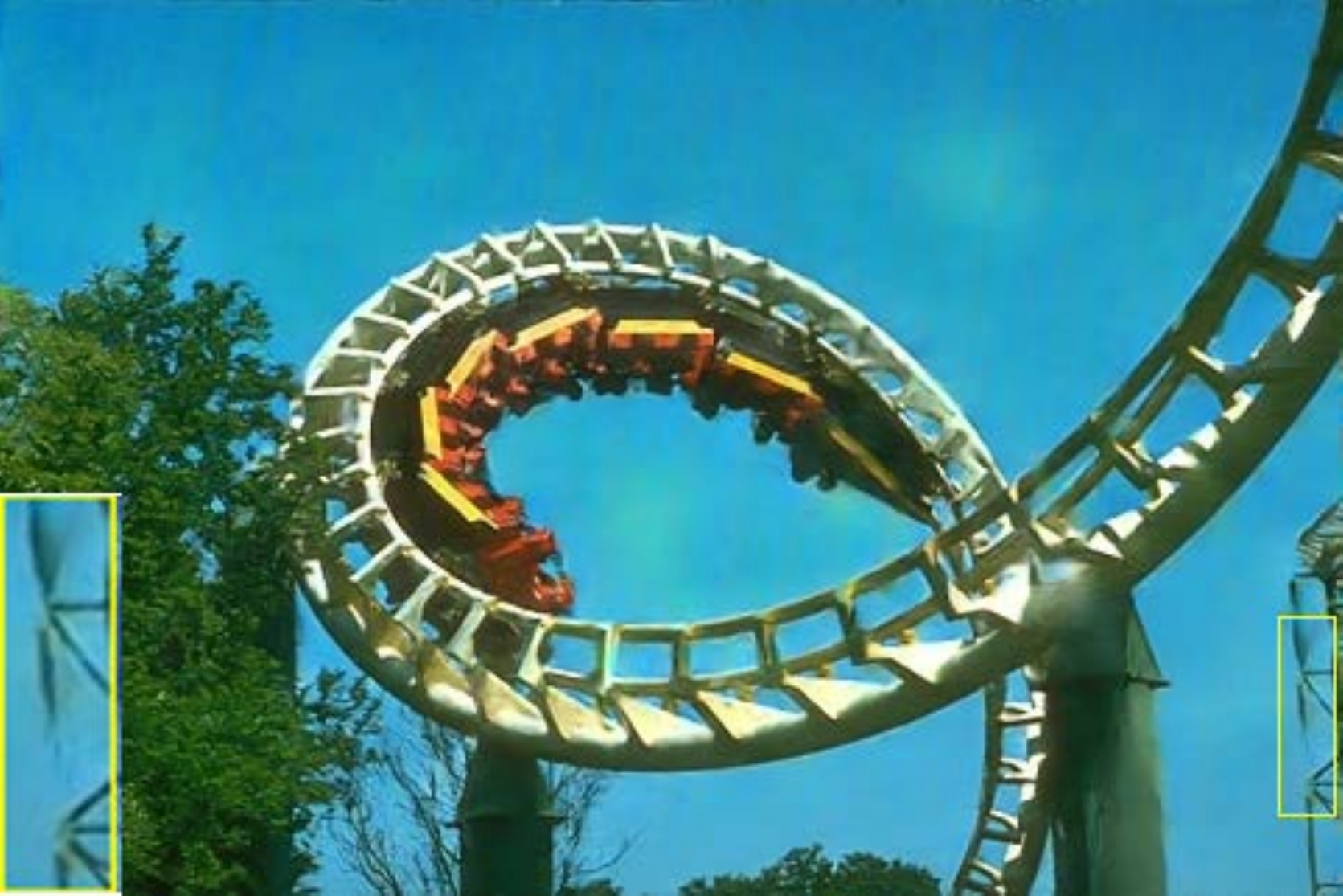} &\hspace{-4mm}
\includegraphics[width = 0.104\linewidth]{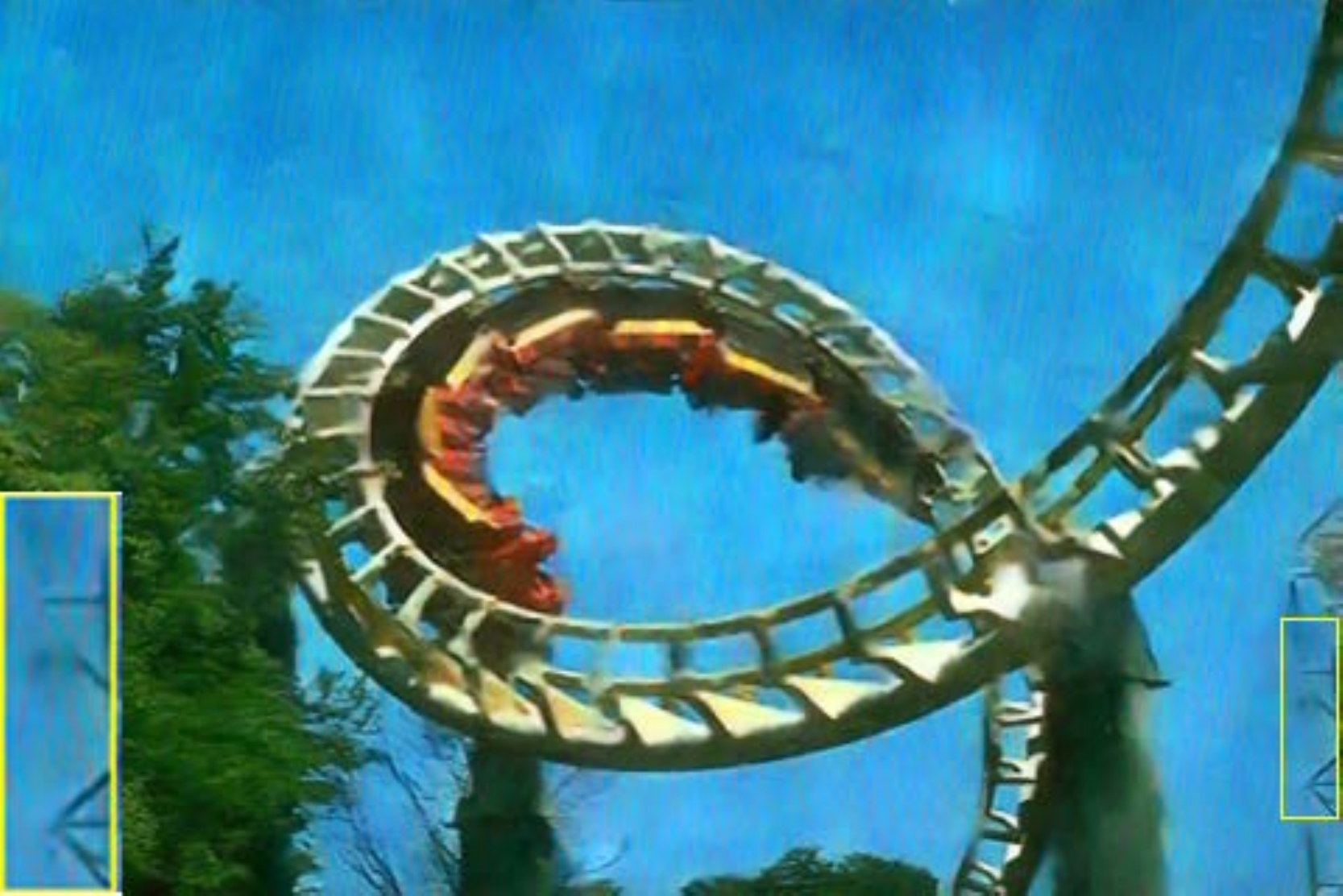} &\hspace{-4mm}
\includegraphics[width = 0.104\linewidth]{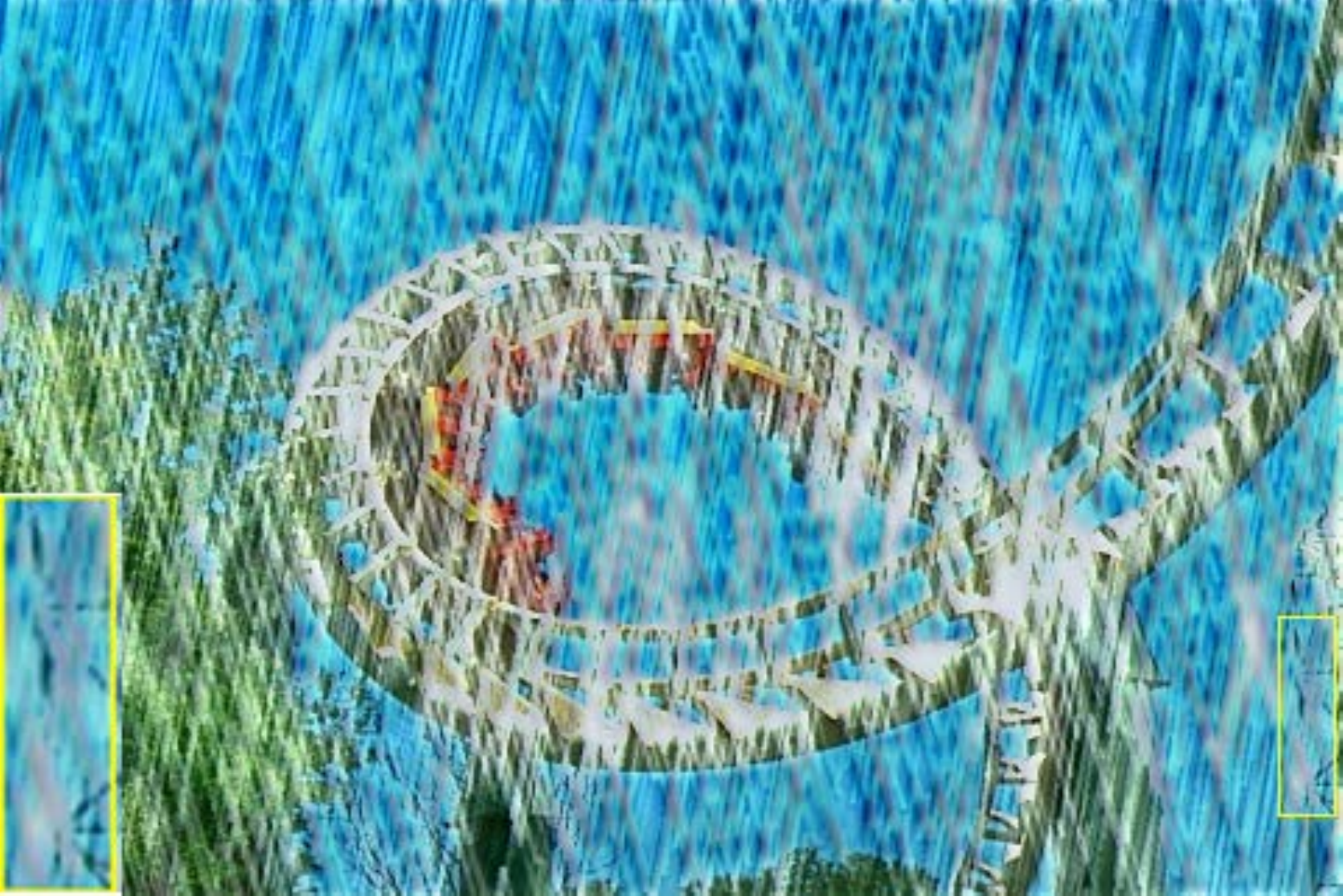} &\hspace{-4mm}
\includegraphics[width = 0.104\linewidth]{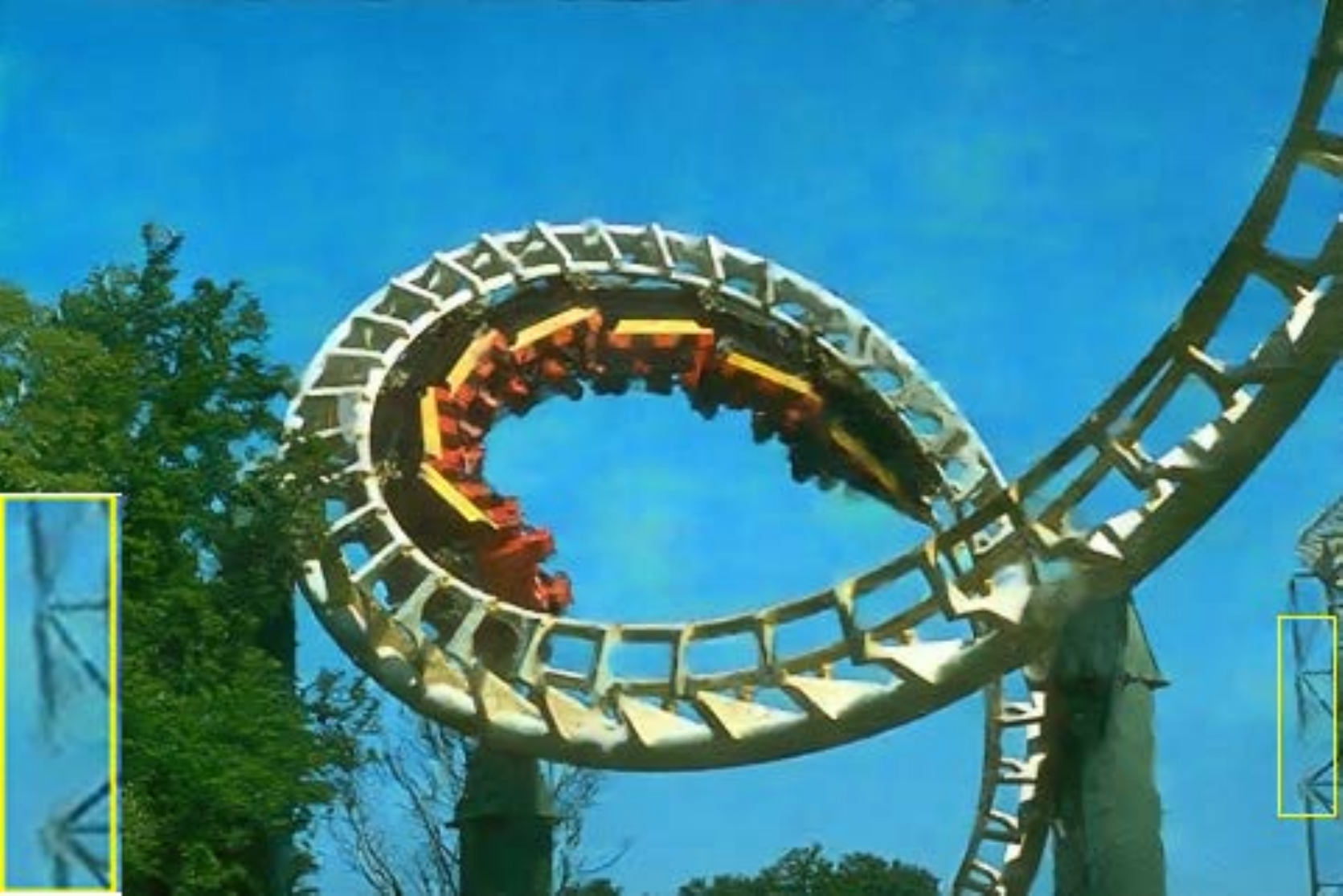}&\hspace{-4mm}
\includegraphics[width = 0.104\linewidth]{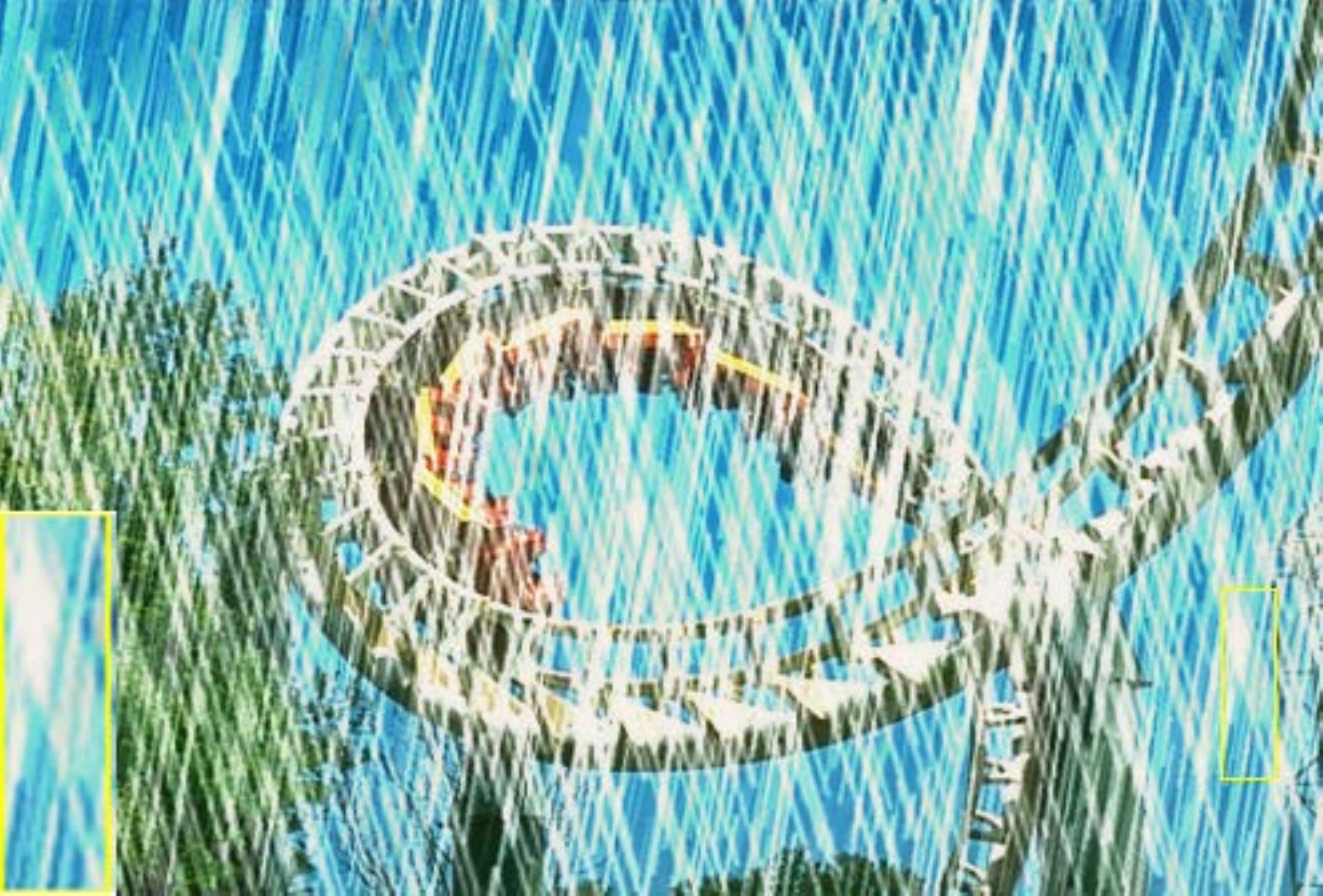}&\hspace{-4mm}
\includegraphics[width = 0.104\linewidth]{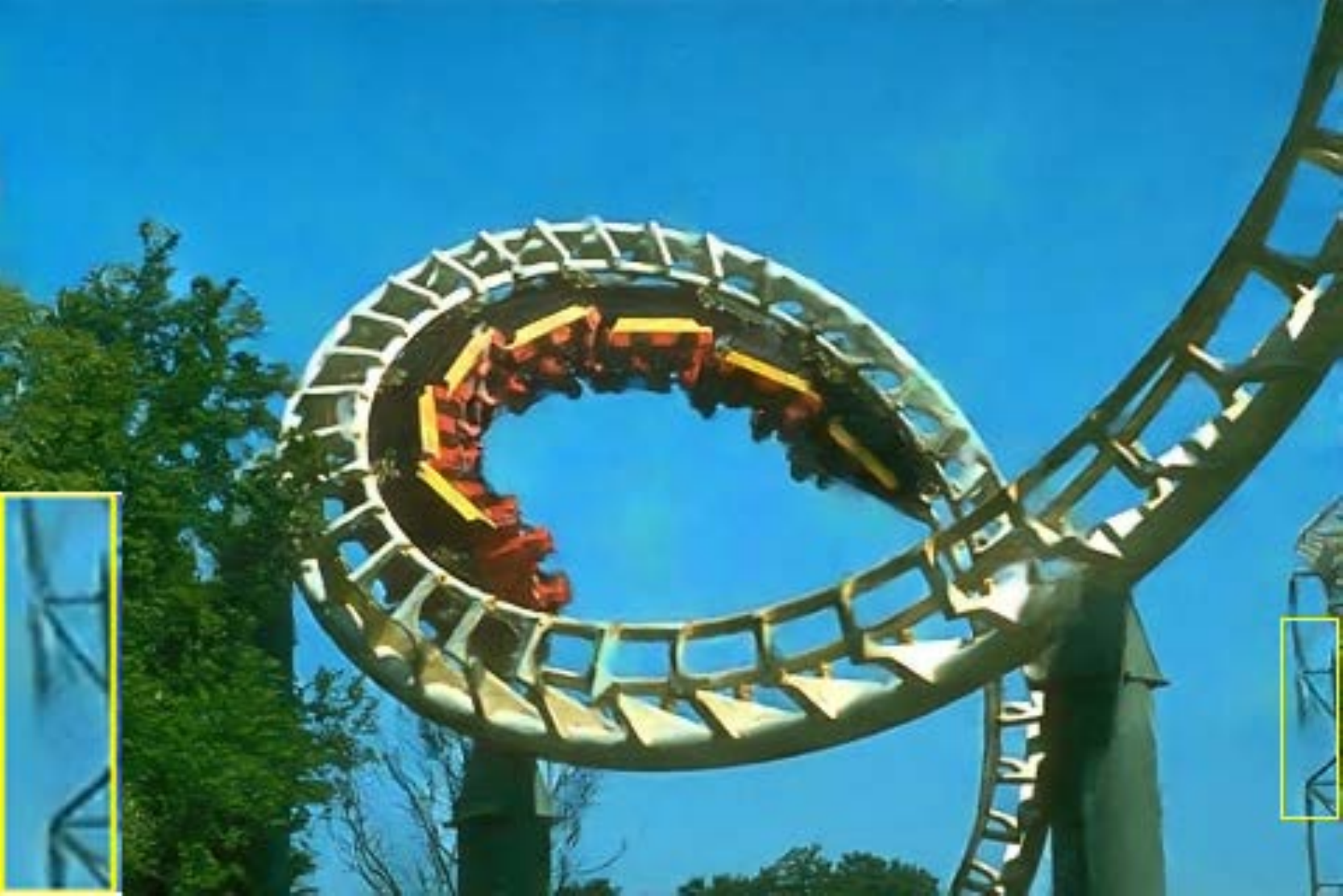}&\hspace{-4mm}
\includegraphics[width = 0.104\linewidth]{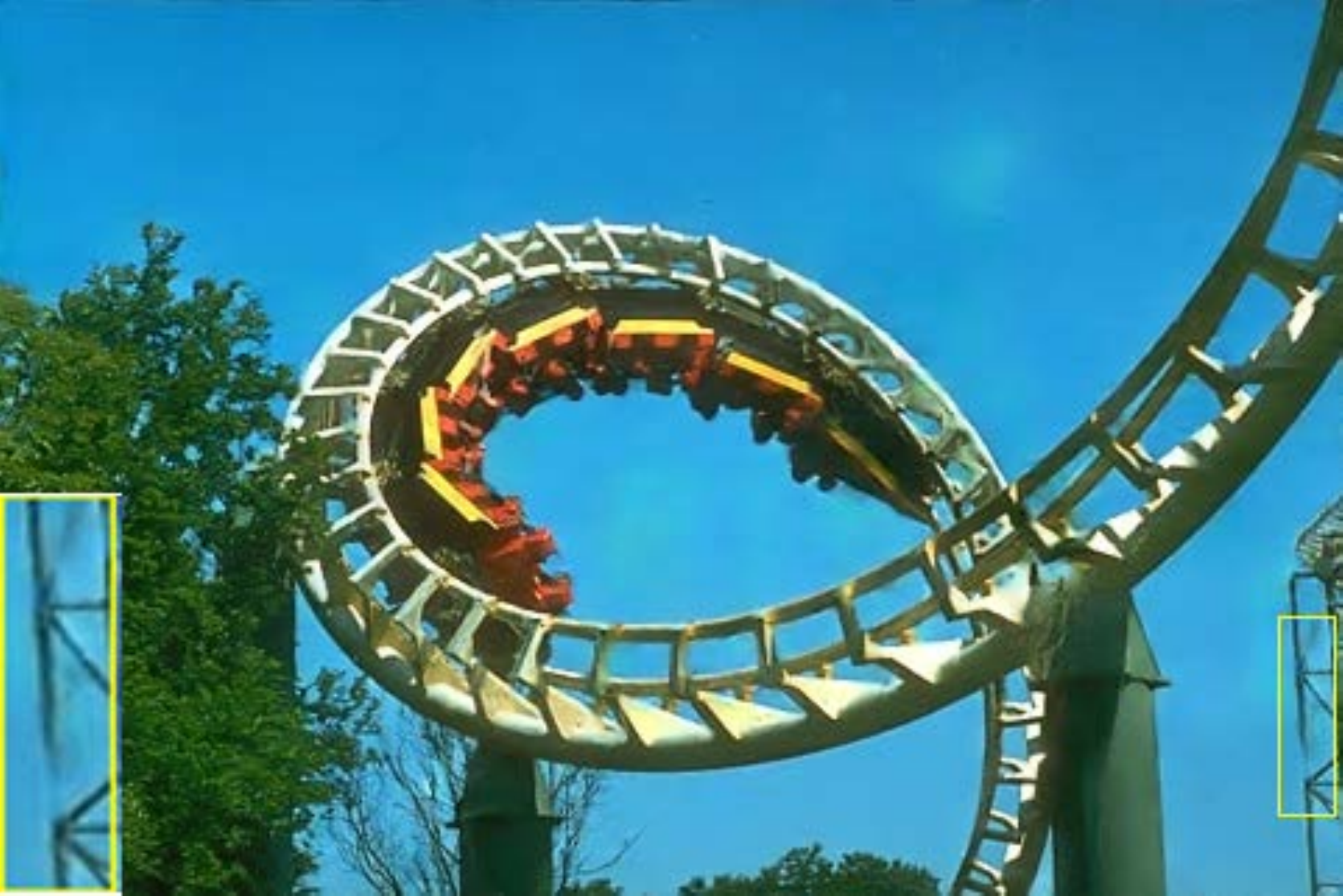}&\hspace{-4mm}
\includegraphics[width = 0.104\linewidth]{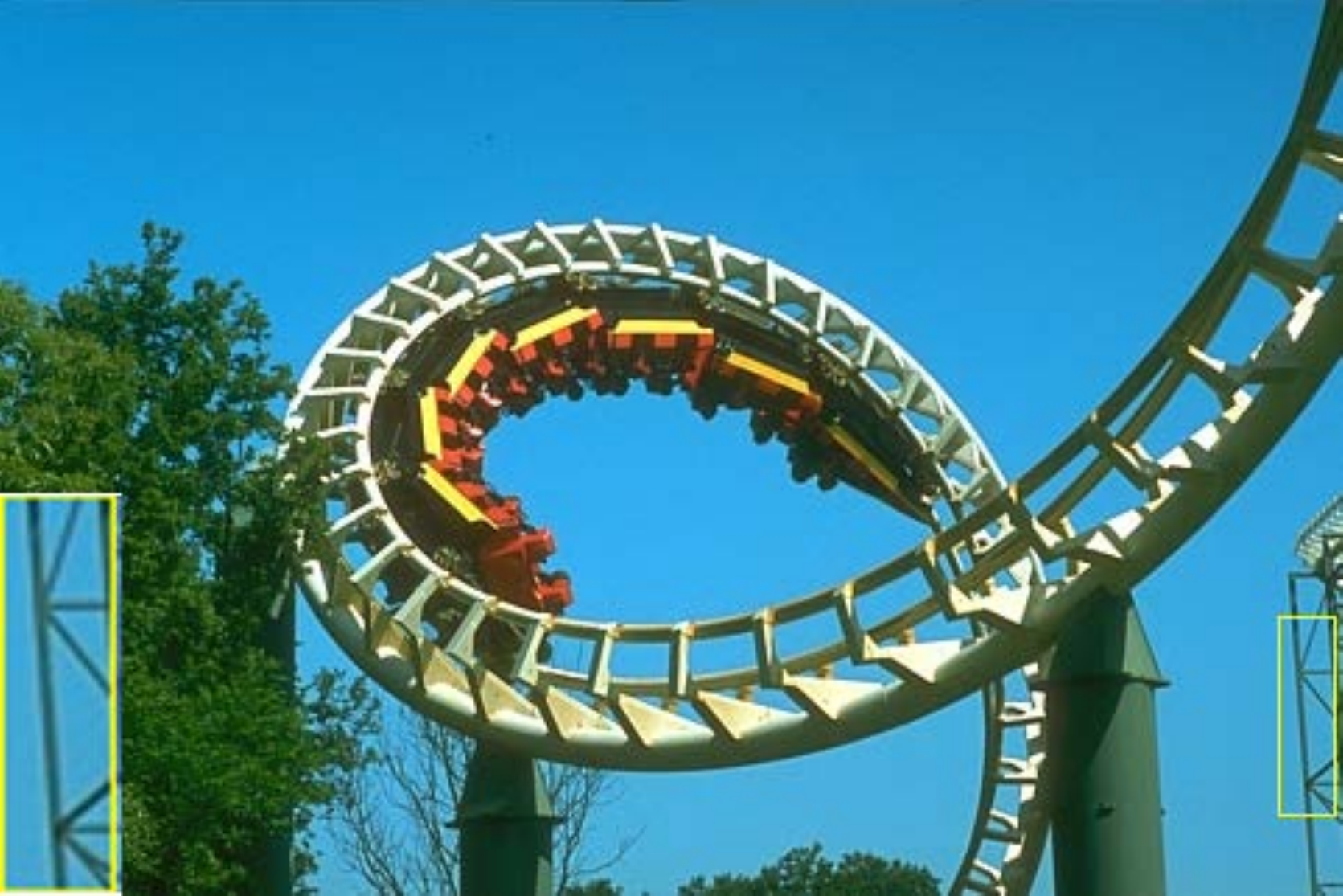}
\\
(a) Input  &\hspace{-4mm} (b) DCSFN&\hspace{-4mm} (c) MSPFN&\hspace{-4mm} (d) DRDNet&\hspace{-4mm} (e) RCDNet &\hspace{-4mm}(f) Syn2Real&\hspace{-4mm} (g) MPRNet&\hspace{-4mm}(h) Ours &\hspace{-4mm}(i) GT
\\
\end{tabular}
\end{center}
% \vspace{-2mm}
\caption{Comparison with state-of-the-art methods on synthetic datasets.
The proposed network is able to restore better texture.
}
\label{fig:deraining-syn-example}
% \vspace{-5mm}
\end{figure*}
%
%
%
% Compared with directly fine-tuning the model by KL-Loss without updating supervised image content that will get better along with the training process, we design an online-updated scheme that the supervised content images are updated along with the training epochs.
%A natural phenomenon is that the deraining results get better with the fine-tuning training epochs going.
%Hence, we design an online-updated scheme that the supervised content images are updated along with the training epochs.
%
%

%
\section{Experiments}
\label{sec: experiments}
We compare the proposed approach with 11 state-of-the-art methods (SOTAs), including
RESCAN~\cite{derain_rescan_li},
NLEDN~\cite{derain_nledn_li},
SSIR~\cite{Derain-cvpr19-semi},
PreNet~\cite{derain_prenet_Ren_2019_CVPR},
SpaNet~\cite{derain_2019_CVPR_spa},
DCSFN~\cite{mm20_wang_dcsfn},
MSPFN~\cite{cvpr20_jiang_mspfn},
DRDNet~\cite{drd_cvpr20_deng},
RCDNet~\cite{cvpr20_wang_rcdnet},
Syn2Real~\cite{cvpr20_syn2real}, and
MPRNet~\cite{derain_mprnet_cvpr21}, on five widely used synthetic datasets and a real-world dataset.
\subsection{Datasets and Evaluation Criteria}
\subsubsection{Synthetic dataset}
We use Rain200H~\cite{derain_jorder_yang}, Rain200L~\cite{derain_jorder_yang},
Rain1200 \cite{derain_zhang_did},
Rain1400~\cite{derain_ddn_fu}, and
Rain12~\cite{derain_lp_li} as the synthetic datasets for training and evaluation.
%which are used to train deraining models and evaluate the restored results.
{Rain200H} is the most challenging dataset, which has 1800 image pairs for training and 200 pairs for testing.
{Rain200L} is the easiest dataset with the same number of training and testing samples as Rain200H.
{Rain1200} has images with heavy, middle, and light rain, and there are 4000 training images and 400 testing images for each density.
{Rain1400} has 12600 training samples and 1400 testing samples.
Since {Rain12} only contains 12 testing samples, we use the model trained on Rain200H to test the deraining results.
We use Rain200H as the dataset for ablation study and analysis.

% \begin{table}[!h]
% \centering
% \caption{Details about synthetic datasets.
% }
% \scalebox{0.76}{
% \begin{tabular}{cccccccc}
% \hline
%  Datasets  & Rain200H& Rain200L& Rain1200 & Rain1400&Rain12 &\\
% \hline
% Train  & 1800 & 1800 & 12000 & 12600 & 0    \\
% \hline
% Test  & 200 &  200 & 1200 &1400& 12   \\
% \hline
% \end{tabular}}
% \label{tab:datasets.}
% \end{table}
%
\subsubsection{Real-world dataset}
%
% Yang~\etal.~\cite{derain_jorder_yang}, Li~\etal.~\cite{deraining-benchmark-analysis}, and Wang~\etal.~\cite{mm20_wang_jdnet} collect a large body of real-world images.
%In, they collect a large body of real-world images.
% We use the images provided by~\citet{derain_jorder_yang},~\citet{ deraining-benchmark-analysis}, and~\citet{mm20_wang_jdnet} to evaluate the deraining performance on real-world data.
%
\citet{derain_jorder_yang}, \citet{deraining-benchmark-analysis}, and \citet{mm20_wang_jdnet} provide a large body of real-world rainy images. We use them to evaluate the deraining results on real-world data.
\subsubsection{Evaluation criteria}
We use two widely used metrics, peak signal to noise ratio (PSNR)~\cite{PSNR_thu} and structural similarity index measure (SSIM)~\cite{SSIM_wang} to evaluate the quality of restored images on synthetic datasets.
As there are no ground-truth for real-world rainy images, we only compare the results visually.
% We use NIQE~\cite{NIQE} to evaluate the restored results the real-world dataset,
%
\subsection{Implementation Details}
We set the number of channels of each convolutional layer except the last one as $20$, and LeakyReLU with $\alpha = 0.2$ is used after each convolutional layer except for the last one.
For the last layer, we use $3 \times 3$ convolution without any activation function in $\mathcal{B}$, $\mathcal{M}$, and $\mathcal{T}$.
%For the convolutional layer, the channel is $3$ without any activation function.
We randomly crop $128\times128$ image patches as input, and the batch size is set as $12$.
We use ADAM optimizer~\cite{adam} to train the network.
The initial learning rate is 0.0005, which will be divided by $10$ at the $300$-th and $400$-th epochs, and the model training terminates after $500$ epochs.
We set $\lambda=0.0001$, $\alpha_{1}=1$, $\alpha_{2}=1$,  $\alpha_{3}=1$,  $\beta_{1}=0.05$, and $\beta_{2}=0.001$.
We train the model for 30 epochs on the real-world dataset, i.e., $\mathrm{Epoch}_{\mathrm{Real}}$ = 30.
Our model is trained with four NVIDIA RTX TITAN GPUs on the Pytorch platform.
\subsection{Results and Analysis on Synthetic Datasets}
\subsubsection{Comparisons with SOTAs on Synthetic Datasets}
%
% Tab.~\ref{tab: the results in synthetic datasets} summarizes the results on five synthetic datasets.
% One can see that our  method achieves superior performance compared with eleven SOTAs on all datasets with with moderate number of parameters.
% %We furthermore provide one challenging rainy example
% As shown in Fig.~\ref{fig:deraining-syn-example},
% the proposed method is able to generate a clearer rain-free image with better texture, while other models either hand down some rain streaks or leave some artifacts.
% Note that the fully-supervised method DRDNet and the semi-supervised approach Syn2Real both fail in this case.
Tab.~\ref{tab: the results in synthetic datasets} reports the results of our method and SOTAs on five synthetic datasets.
We can see that our method achieves the best results on all tested datasets in PSNR and SSIM.
% In the most challenging Rain200H dataset, the PSNR and SSIM of the proposed method are increased by 0.036 dB and 0.67\% compared with the best baseline MPRNet respectively, and meanwhile the number of parameters is reduced by 44\%.
We further show some deraining results in Fig.~\ref{fig:deraining-syn-example}.
It can be observed that our method can restore better details and textures and obtain clearer background images, while other approaches hand down some rain streaks or lose some details.

%to restore rain-free images.
%
\subsubsection{Analysis on BiSCSM}
\begin{figure}[!t]
\begin{center}
\begin{tabular}{ccc}
\includegraphics[width = 0.32\linewidth]{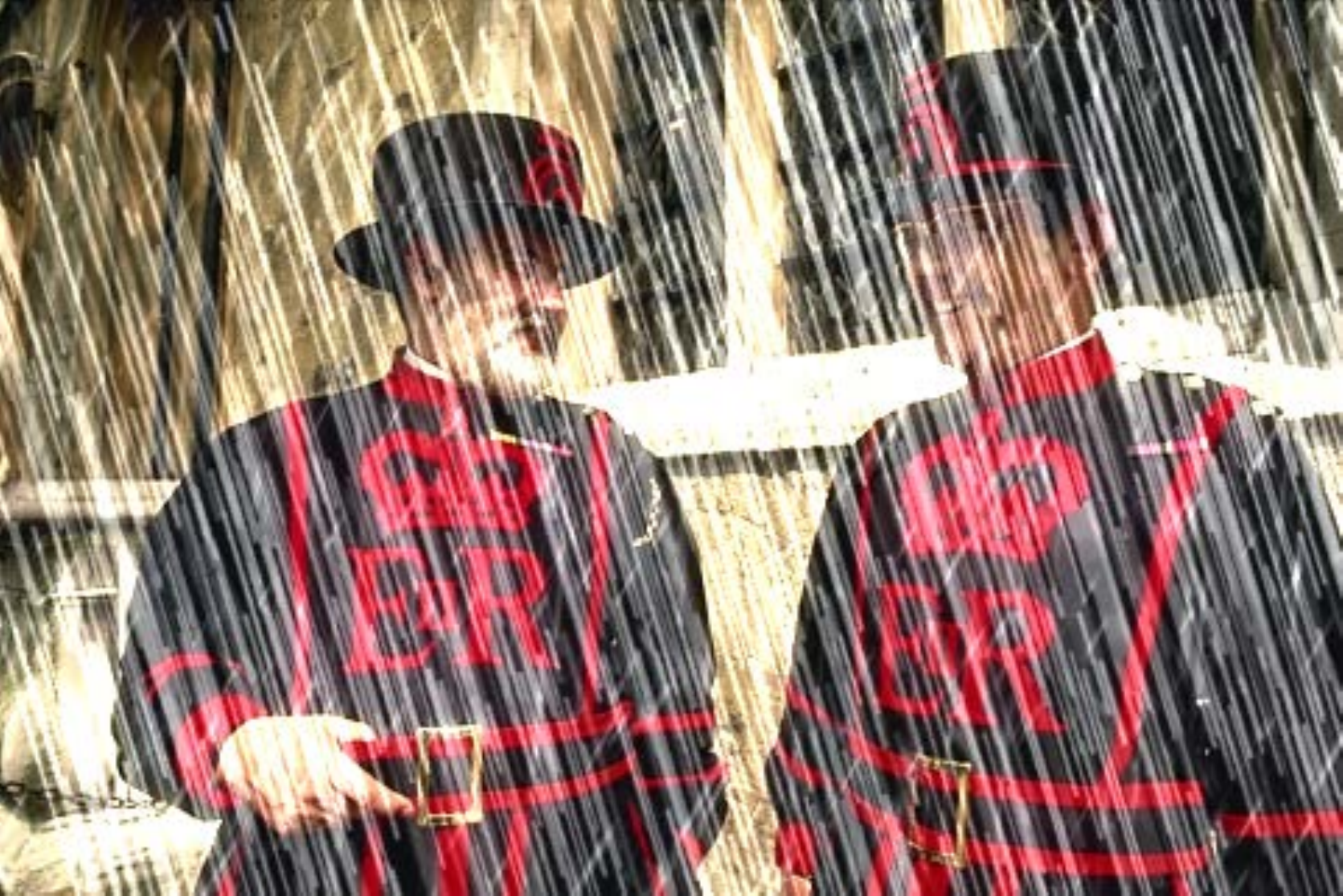} &\hspace{-4mm}
\includegraphics[width = 0.32\linewidth]{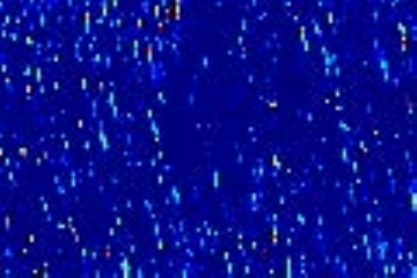} &\hspace{-4mm}
\includegraphics[width = 0.32\linewidth]{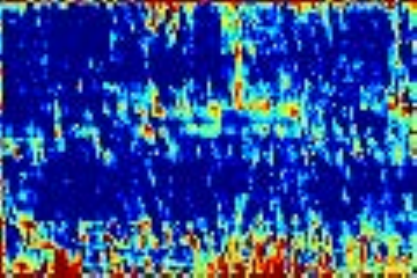}
\\
(a) Input &\hspace{-4mm} (b) Before &\hspace{-4mm} (c) After
\end{tabular}
\end{center}
% \vspace{-2mm}
\caption{Visualization of the feature maps before and after BiSCSM.
}
\label{fig:Before and After denote the visualization feature map before and after BiSCSM.}
% \vspace{-2mm}
\end{figure}

We analyze the effect of different components of BiSCSM and report the results in Tab.~\ref{tab:The analysis on basic component of the network.}.
Note that the method ($M_{1}$) does not generate better results if we remove BiSCSM from the encoder-decoder framework.
Furthermore, we replace the Down-to-Up and Up-to-Down operations in BiSCSM by element-wise summation ($M_{2}$) and $1 \times 1$ convolution ($M_{3}$) to fuse features at different scales.
We find that both operations do not perform well compared to the proposed ones, demonstrating the effectiveness of the mining modules.
%can learn more useful features for better deraining.
%
Fig.~\ref{fig:Before and After denote the visualization feature map before and after BiSCSM.}
%presents the visualization results of
visualizes the feature maps before and after BiSCSM.
One can see that the features are significantly enhanced after BiSCSM. %which thus facilitates the rain streaks removal.
\begin{table}[!t]%\footnotesize
\scalebox{0.67}{
\centering
\begin{tabular}{lccccccc}
\hline
Experiments& $M_{1}$ &$M_{2}$ & $M_{3}$       & $M_{4}$     &$M_{5}$      &$M_{6}$& $M_{7}$ (Ours)            \\
\hline
Sum     &   \XSolidBrush & \CheckmarkBold       & \XSolidBrush    & \XSolidBrush &\XSolidBrush  & \XSolidBrush &\XSolidBrush                  \\
Conv$_{1 \times 1}$  &\XSolidBrush &\XSolidBrush & \CheckmarkBold &\XSolidBrush & \XSolidBrush &\XSolidBrush &\XSolidBrush       \\
Down-to-Up   & \XSolidBrush &\XSolidBrush& \XSolidBrush & \CheckmarkBold&\XSolidBrush  &\XSolidBrush&\XSolidBrush                           \\
Up-to-Down   &\XSolidBrush &\XSolidBrush       & \XSolidBrush            & \XSolidBrush            & \CheckmarkBold&\XSolidBrush&\XSolidBrush\\
BiSCSM      & \XSolidBrush &\XSolidBrush      &\XSolidBrush              &\XSolidBrush             &\XSolidBrush         &\CheckmarkBold&  \CheckmarkBold                       \\
Position&\XSolidBrush &\CheckmarkBold&\CheckmarkBold& \CheckmarkBold &\CheckmarkBold &\XSolidBrush&        \CheckmarkBold                   \\
\hline
PSNR $\uparrow$ & 29.583&29.625  & 29.541       & 29.393     &  29.828      & 29.953 &  \textbf{29.985}     \\
\hline
SSIM $\uparrow$ &0.9168 &0.9177   & 0.9156       & 0.9172     & 0.9201    &  0.9218   &  \textbf{0.9218}  \\\hline
\end{tabular}}
% \vspace{-2mm}
\caption{Ablation study of BiSCSM.
		\CheckmarkBold and \XSolidBrush denote that the corresponding component is adopted and not adopted, respectively.}
\label{tab:The analysis on basic component of the network.}
\end{table}
We also test the single-directional mining modules, i.e., using Down-to-Up ($M_{4}$) or Up-to-Down ($M_{5}$) mining module alone, and the results show the bidirectional manner is better and can mine richer features for better deraining ($M_{4}$ and $M_{5}$ vs. $M_{7}$).
Finally, we observe that position awareness can improve the deraining performance ($M_{6}$ vs. $M_{7}$).
\begin{table}[!t]%\footnotesize
	\scalebox{0.86}{
		\centering
		\begin{tabular}{lccccc}
\hline
Experiments& $R_{1}$ &$R_{2}$ & $R_{3}$       & $R_{4}$     &$R_{5}$ (Ours)    \\
\hline
$1/2$ Scale& \XSolidBrush   & \CheckmarkBold       & \XSolidBrush    & \XSolidBrush &   \CheckmarkBold         \\
$1/4$ Scale&\XSolidBrush  & \XSolidBrush       & \CheckmarkBold             & \XSolidBrush&   \CheckmarkBold    \\
Full Scale   &\XSolidBrush   & \XSolidBrush     &\XSolidBrush              & \CheckmarkBold            &\XSolidBrush  \\
\hline
PSNR $\uparrow$& 29.824&29.879  & 29.839       & 29.970     &  \textbf{29.985}           \\
\hline
SSIM $\uparrow$&0.9195 &0.9209   & 0.9204       & 0.9211     & \textbf{0.9218}     \\
\hline
\end{tabular}}
% \vspace{-2mm}
\caption{Ablation study of the multi-scale compact constraints.}
\label{tab:Results on different types of multi-scale compact constraint.}
% \vspace{-4mm}
\end{table}
\subsubsection{Analysis on MSCC}
Compared with deep models that have no constraints on intermediate layers, our proposed MSCC can make the deep network more compact and learn more useful features.
Tab.~\ref{tab:Results on different types of multi-scale compact constraint.} shows the results of different types of constraints.
We observe that the model without any contraints ($R_{1}$) performs the worst, while the results get better as we add different constraints and reach the best when both $1/2$ ($R_{2}$) and $1/4$ ($R_{3}$) scale constraints are added.

We also consider replacing the $1/2$ and $1/4$ scale images with full-scale images to constrain the network.
The comparison between $R_{4}$ and $R_{5}$ demonstrates that full-scale images are less effective than the multi-scale images, which is probably due to the better structures of Laplacian pyramid images.
\subsubsection{Ablation Study on Collaborative Learning}
We conduct an ablation study of the proposed collaborative network in Tab.~\ref{tab:Results on different types of networks.}. %reports the results of different types of networks.
The results show that the model achieves the best performance when the sub-networks $\mathcal{T}$ and $\mathcal{M}$ are learned with $\mathcal{B}$ in a  collaborative manner.
%learning manner are able to help better restore rain-free images.
% One is deserved to mention that we cascade the three-stream sub-networks to form a longer network model and whether the collaborative learning is better than the cascaded pattern or not.
%One may wonder to know that what will be the results if we cascade the three-stream sub-networks to form a longer network model?
% is deserved to mention that we cascade the three-stream sub-networks to form a longer network model and whether the collaborative learning is better than the cascaded pattern or not.

We also consider a case by cascading the three sub-networks to form a deep network model.
Specifically, we cascade the network modules $\mathcal{N}_{11}$, $\mathcal{N}_{12}$, $\mathcal{N}_{13}$, $\mathcal{N}_{21}$, $\mathcal{N}_{22}$, and $\mathcal{N}_{31}$ and ensure the cascaded network has roughly the same number of parameters as the collaborative network. We find that the cascaded network does not perform as well as the collaborative network, which further demonstrates the effectiveness of the proposed collaborative learning manner.
%also further verifies the proposed collaborative learning manner is indeed a better learning scheme.
%We hope this learning manner can boost low-level vision development.

\begin{table}[!t]
\centering
\scalebox{0.68}{
\begin{tabular}{c|c|c|c|c}
\hline
   & w/o $\mathcal{T}$ \& w/o $\mathcal{M}$ & w/o $\mathcal{T}$ \& w/ $\mathcal{M}$& Cascaded  & w/ $\mathcal{T}$ \& w/ $\mathcal{M}$ (Ours)   \\
\hline
PSNR $\uparrow$ &  29.563& 29.355& 29.642 &\textbf{29.985}  \\
\hline
SSIM $\uparrow$ &0.9158&  0.9146& 0.9188 & \textbf{0.9218}  \\
\hline
\end{tabular}}
% \vspace{-2mm}
\caption{Ablation study on collaborative learning.
}

\label{tab:Results on different types of networks.}
% \vspace{-2mm}
\end{table}
\begin{figure}[!t]
\begin{center}
\begin{tabular}{cc}
\includegraphics[width = 0.499\linewidth]{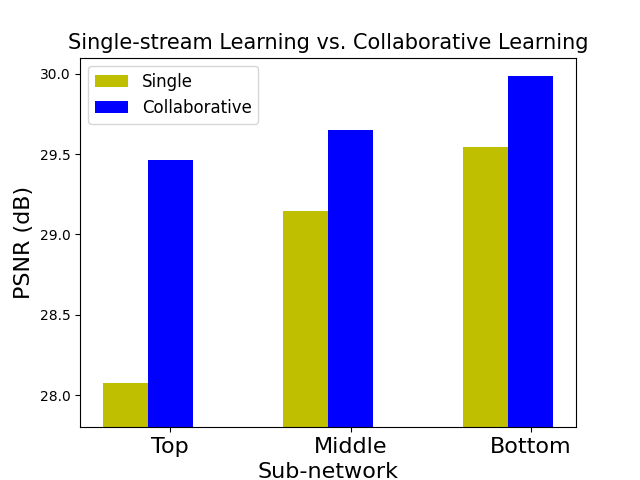} &\hspace{-5mm}
\includegraphics[width = 0.499\linewidth]{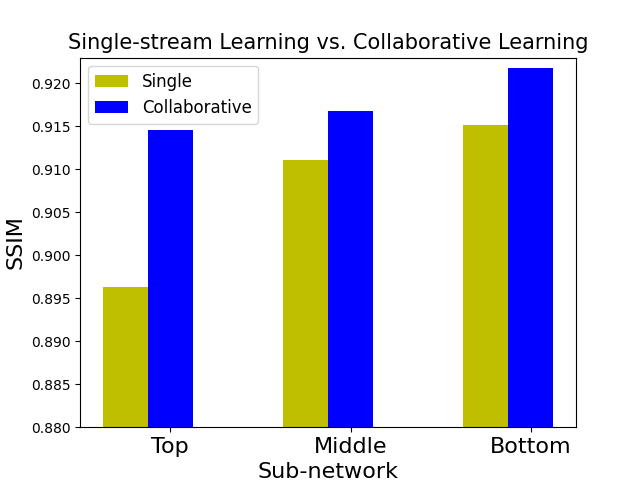}
\\
(a) PSNR &\hspace{-4mm} (b) SSIM
\end{tabular}
\end{center}
% \vspace{-2mm}
\caption{Comparison between single-stream learning and collaborative learning.
}
\label{fig: Comparative results between single learning and collaborative progressive learning.}
% \vspace{-4mm}
\end{figure}
\begin{figure}[!t]
\begin{center}
\begin{tabular}{cc}
\includegraphics[width = 0.499\linewidth]{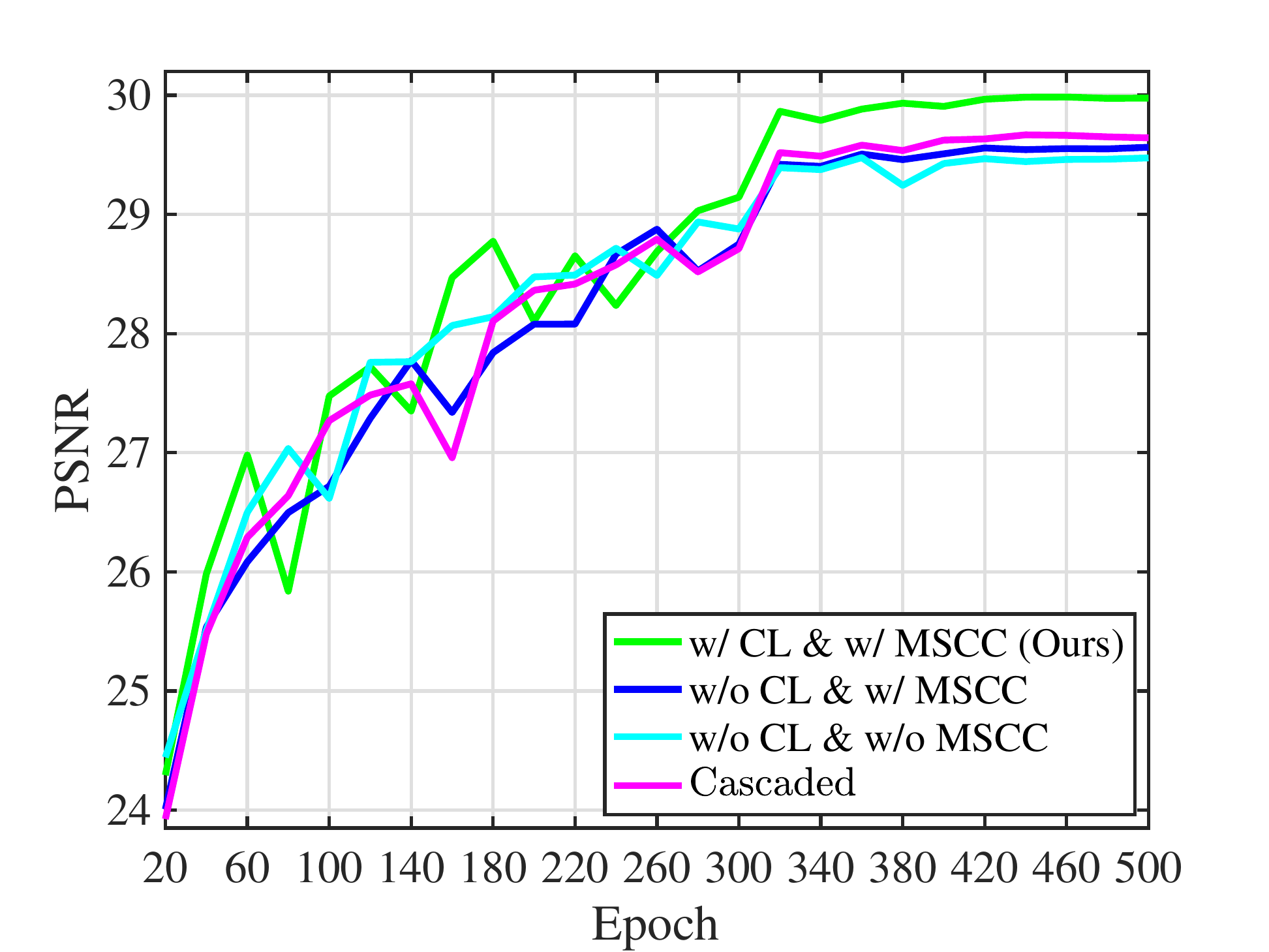} &\hspace{-5mm}
\includegraphics[width = 0.499\linewidth]{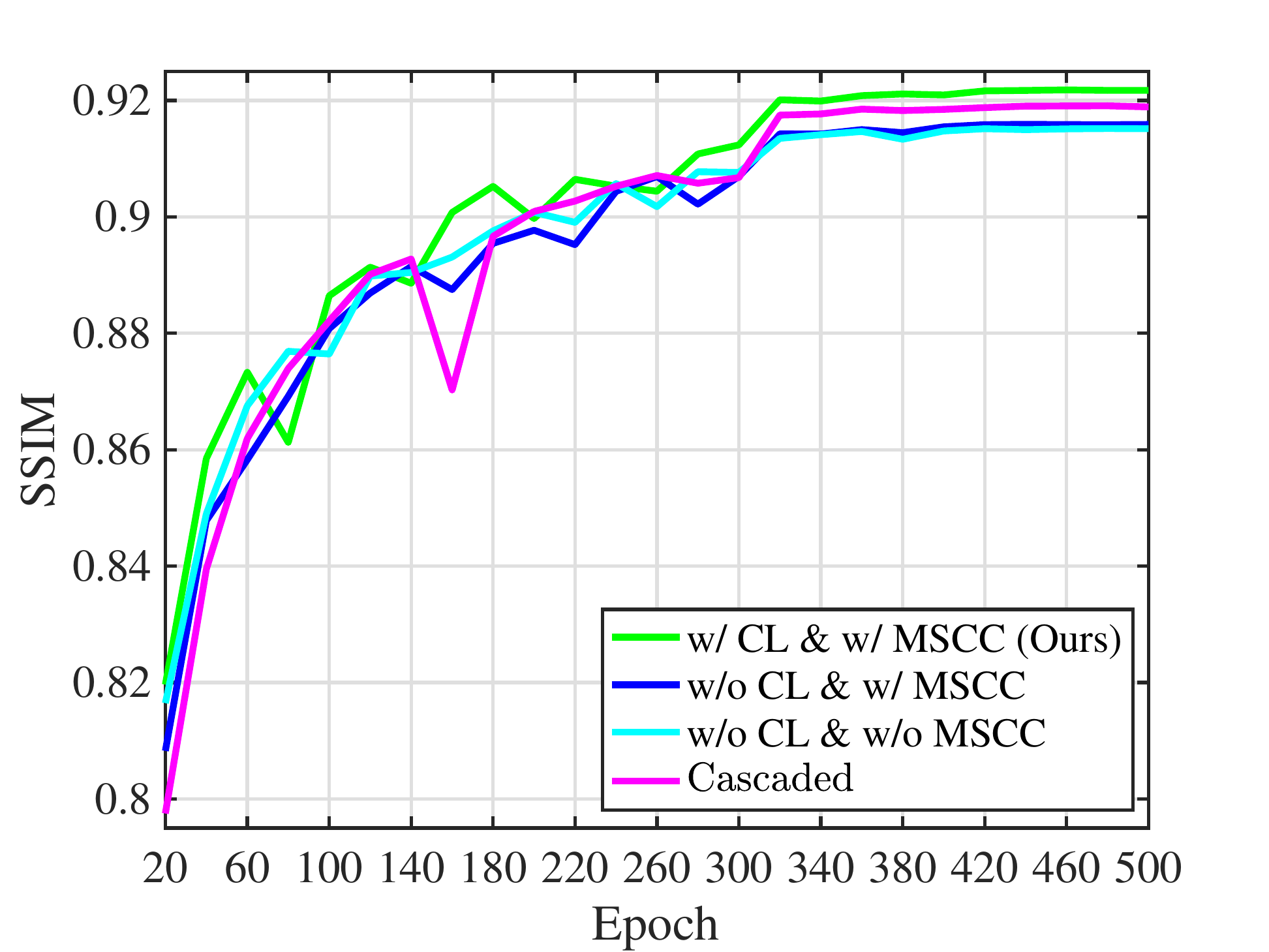}
\\
(a) PSNR &\hspace{-4mm} (b) SSIM
\end{tabular}
\end{center}
% \vspace{-2mm}
\caption{Further ablation study on collaborative learning (CL) and multi-scale compact constraints (MSCC).
}
\label{fig: Results on collaborative progressive learning and multi-scale compact constraint.}
% \vspace{-4mm}
\end{figure}
\begin{figure*}[!t]%\footnotesize
\vspace{-1mm}
\begin{center}
\begin{tabular}{ccccccccc}
% \includegraphics[width = 0.118\linewidth]{real/real-input-1_.png} &\hspace{-4mm}
% \includegraphics[width = 0.118\linewidth]{real/real-dcsfn-1_.png} &\hspace{-4mm}
% \includegraphics[width = 0.118\linewidth]{real/real-mspfn-1_.png} &\hspace{-4mm}
% \includegraphics[width = 0.118\linewidth]{real/real-drd-1_.png} &\hspace{-4mm}
% \includegraphics[width = 0.118\linewidth]{real/real-rcd-1_.png}&\hspace{-4mm}
% \includegraphics[width = 0.118\linewidth]{real/real-syn2real-1_.png}&\hspace{-4mm}
% \includegraphics[width = 0.118\linewidth]{real/real-mpr-1_.png}&\hspace{-4mm}
% \includegraphics[width = 0.118\linewidth]{real/real-ours-1_.png}
% \\
\includegraphics[width = 0.118\linewidth]{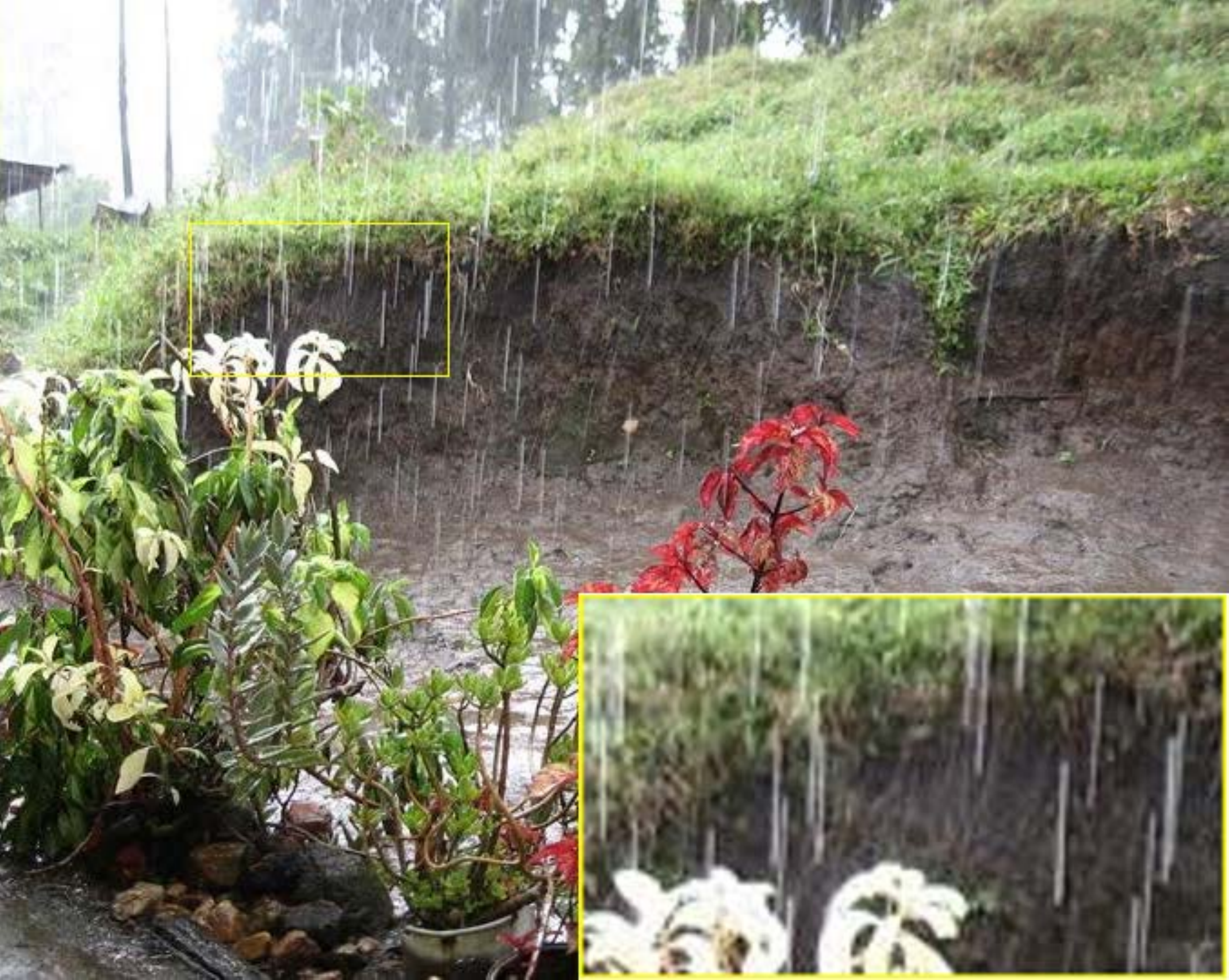} &\hspace{-4mm}
\includegraphics[width = 0.118\linewidth]{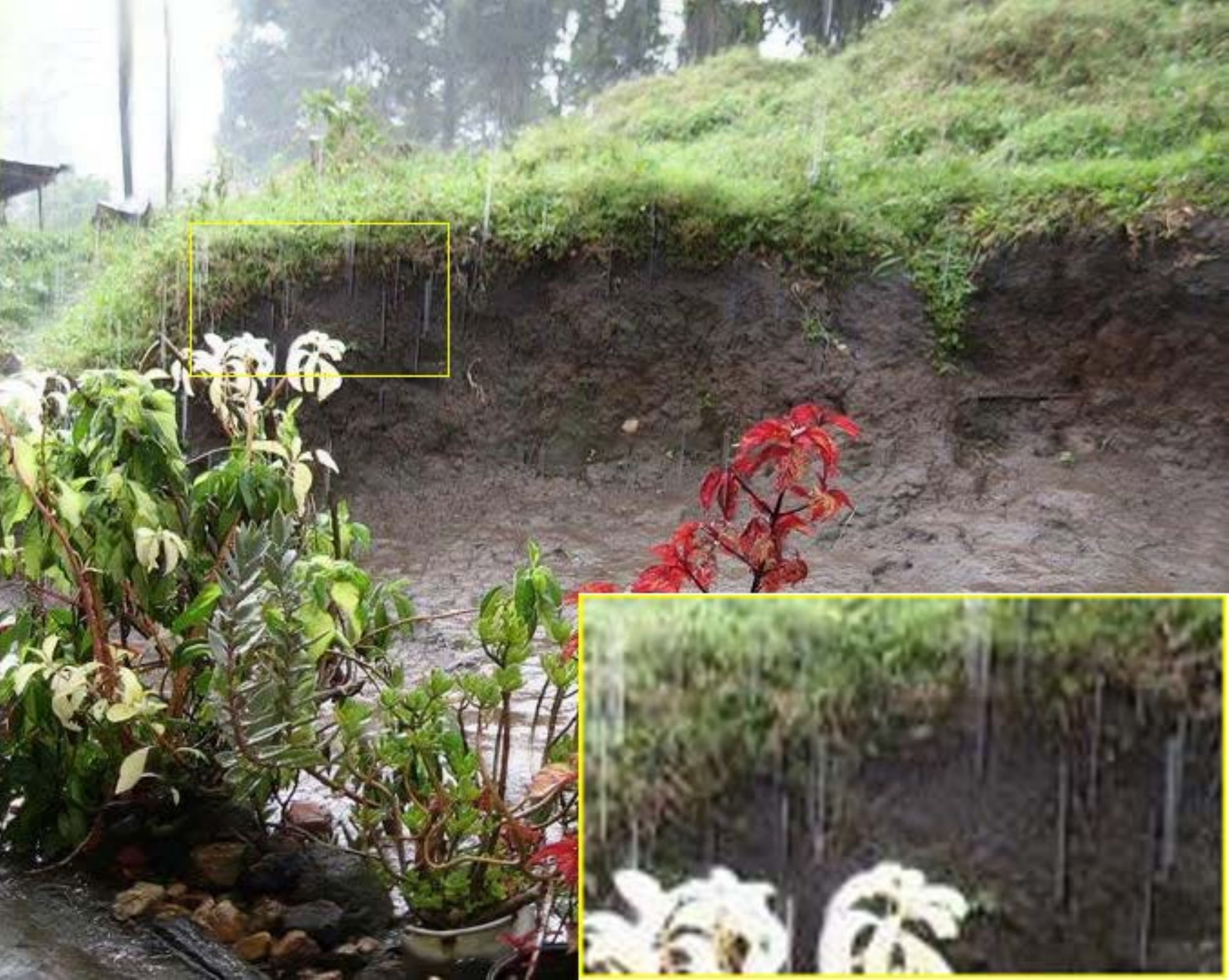} &\hspace{-4mm}
\includegraphics[width = 0.118\linewidth]{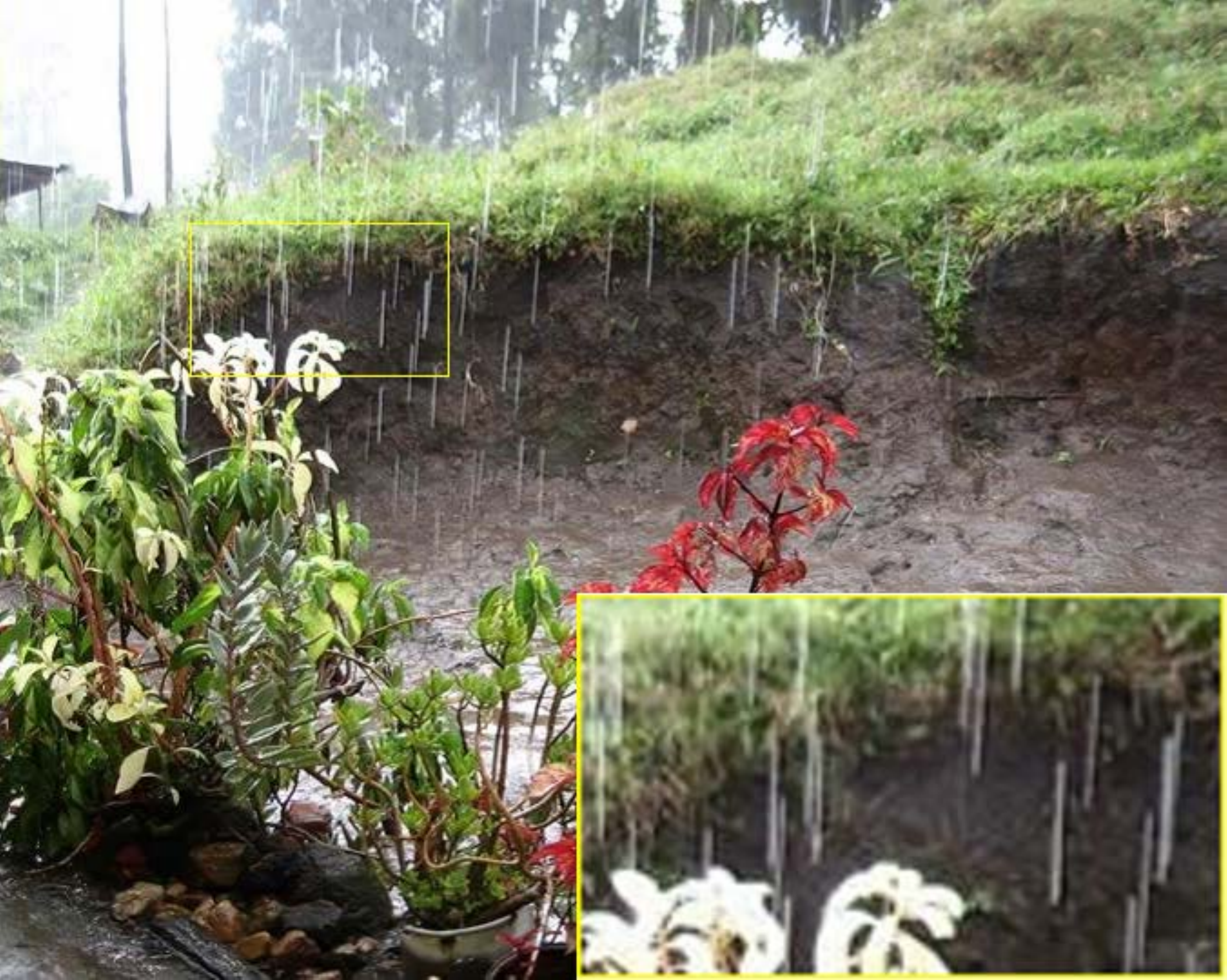} &\hspace{-4mm}
\includegraphics[width = 0.118\linewidth]{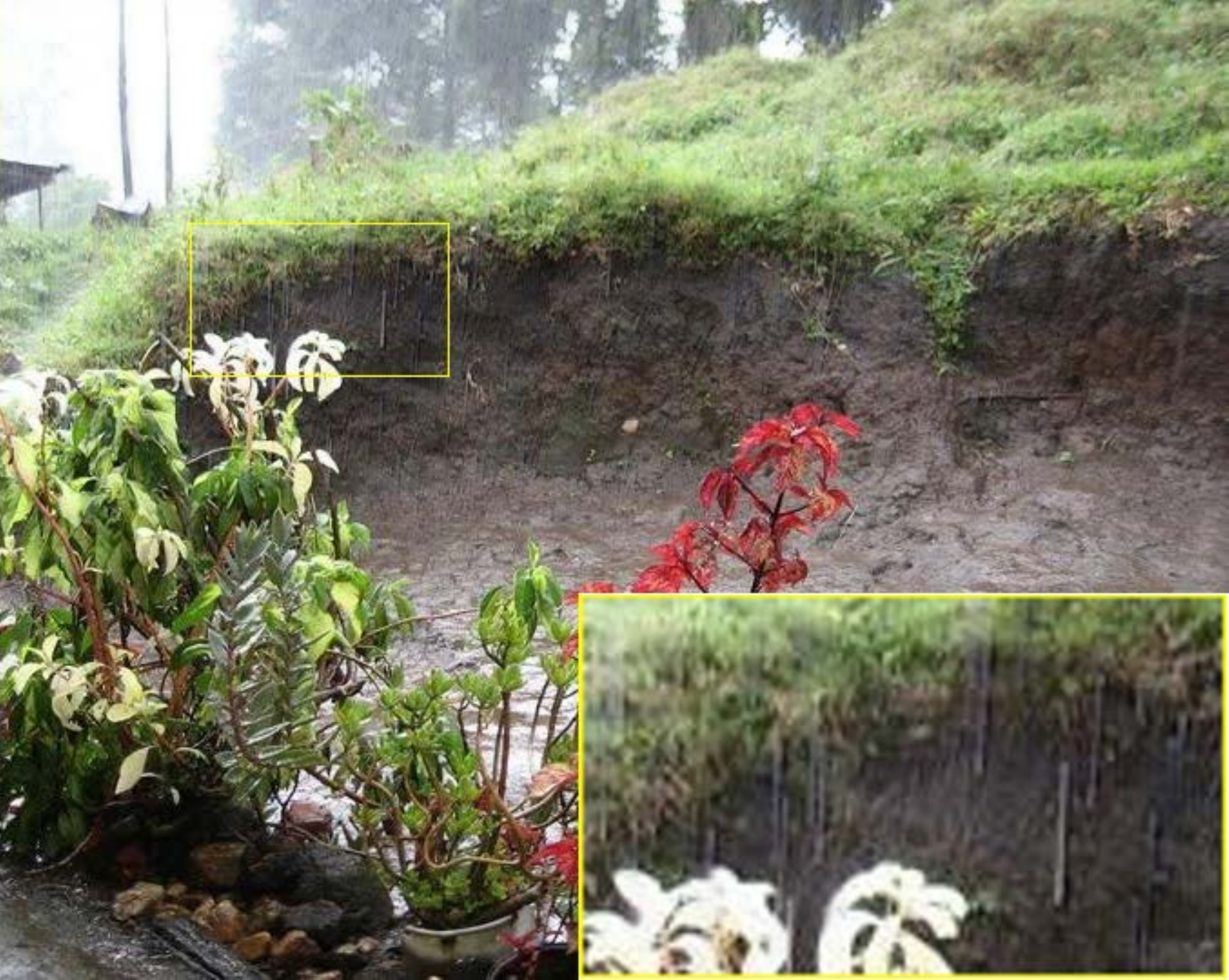} &\hspace{-4mm}
\includegraphics[width = 0.118\linewidth]{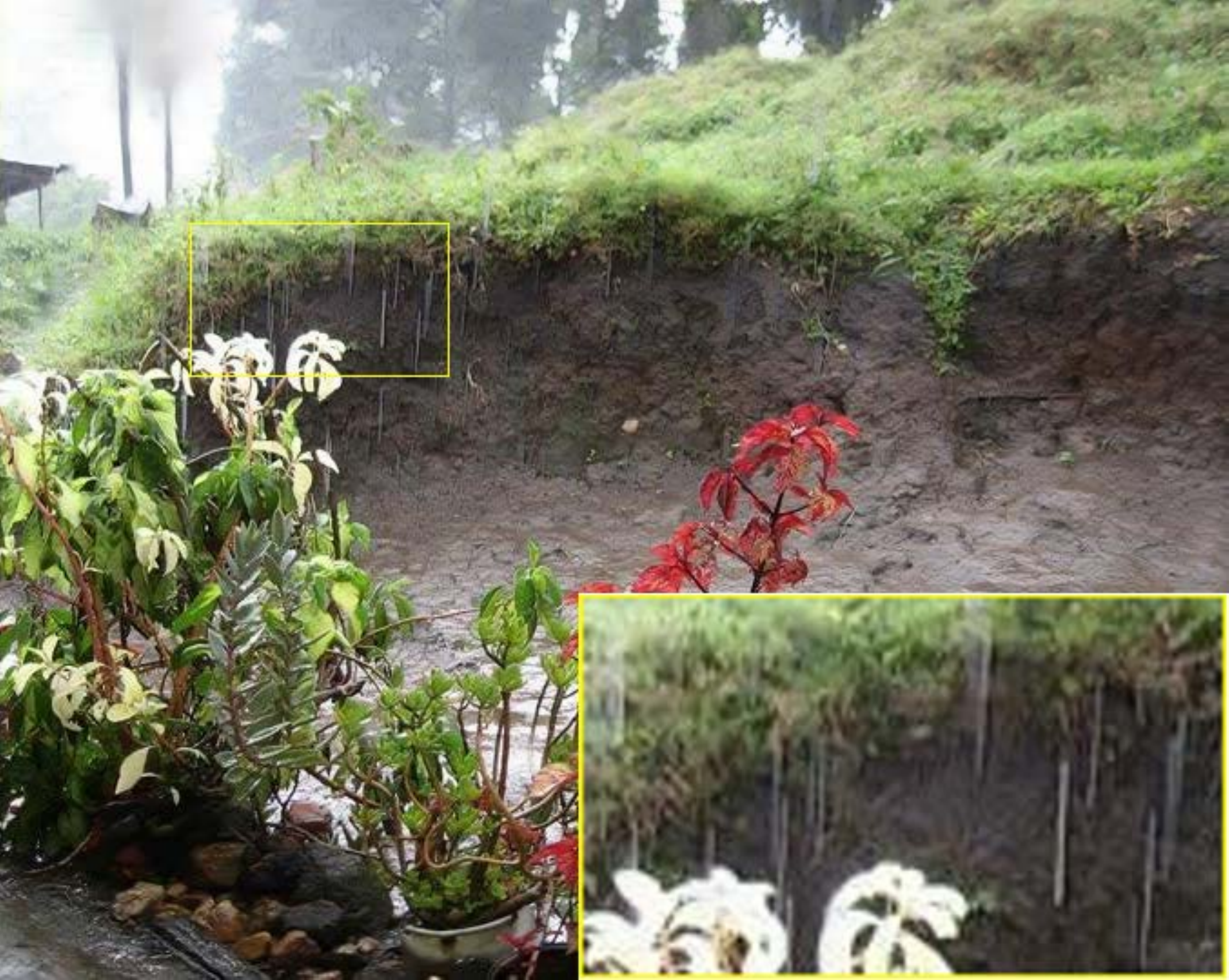}&\hspace{-4mm}
\includegraphics[width = 0.118\linewidth]{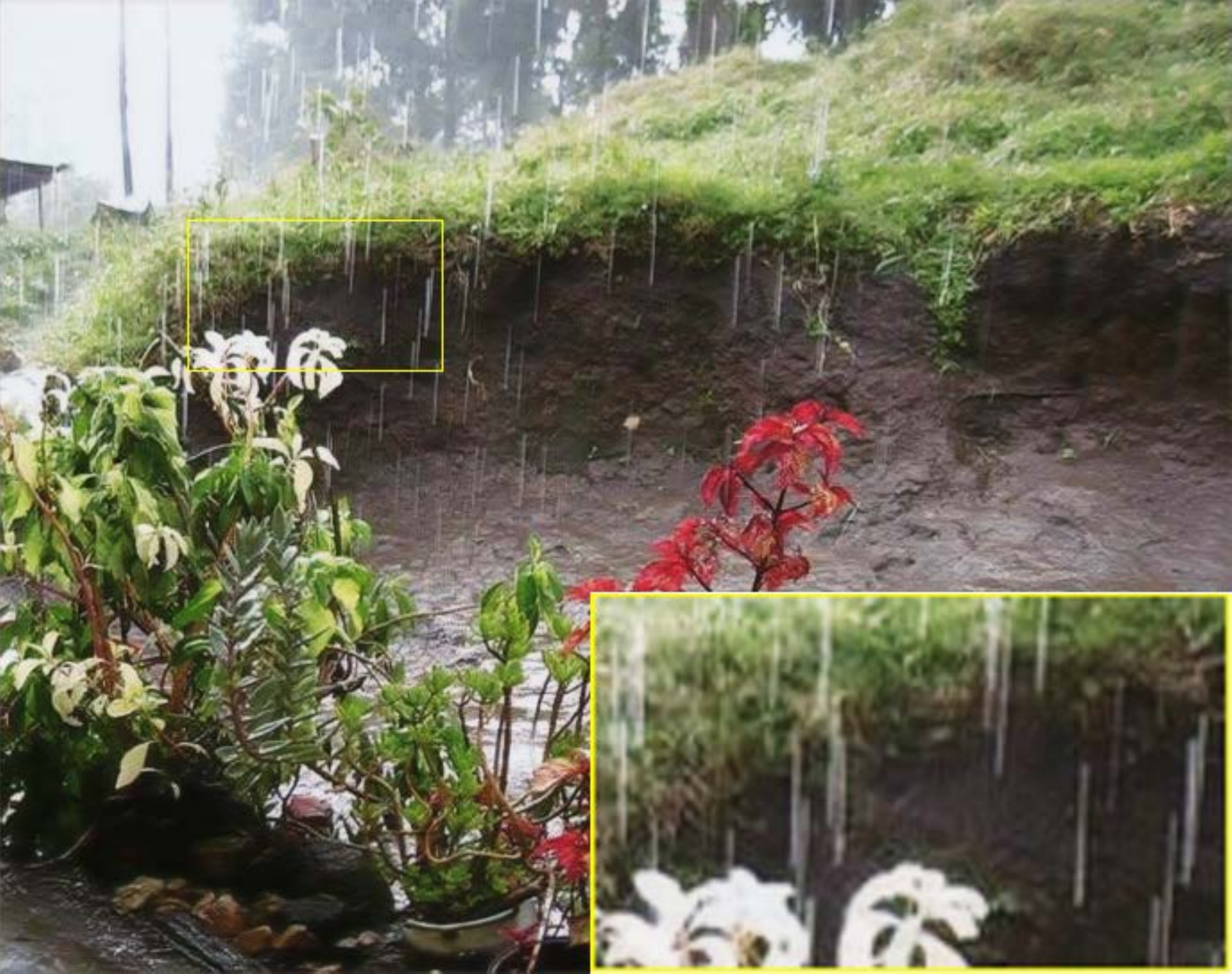}&\hspace{-4mm}
\includegraphics[width = 0.118\linewidth]{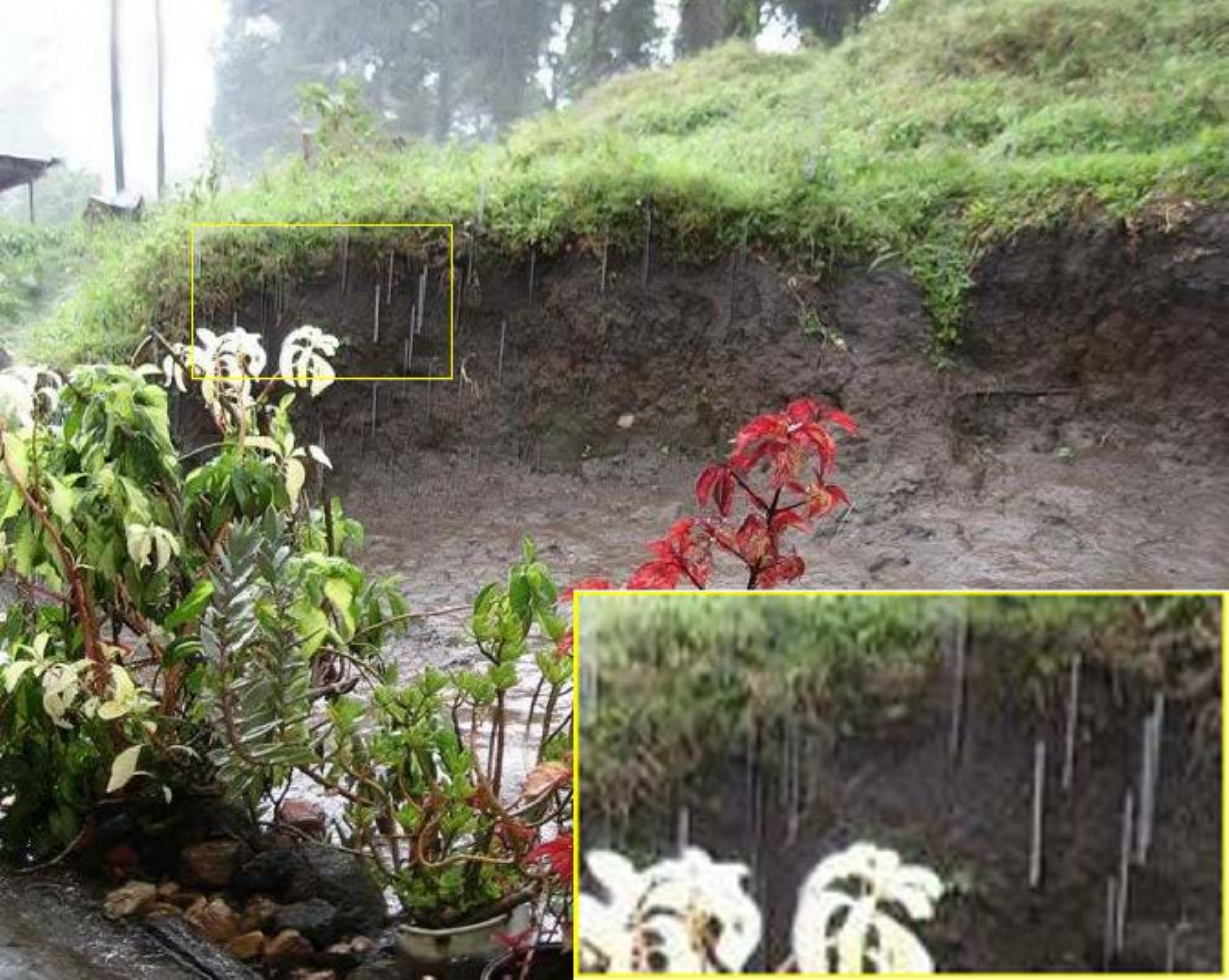}&\hspace{-4mm}
\includegraphics[width = 0.118\linewidth]{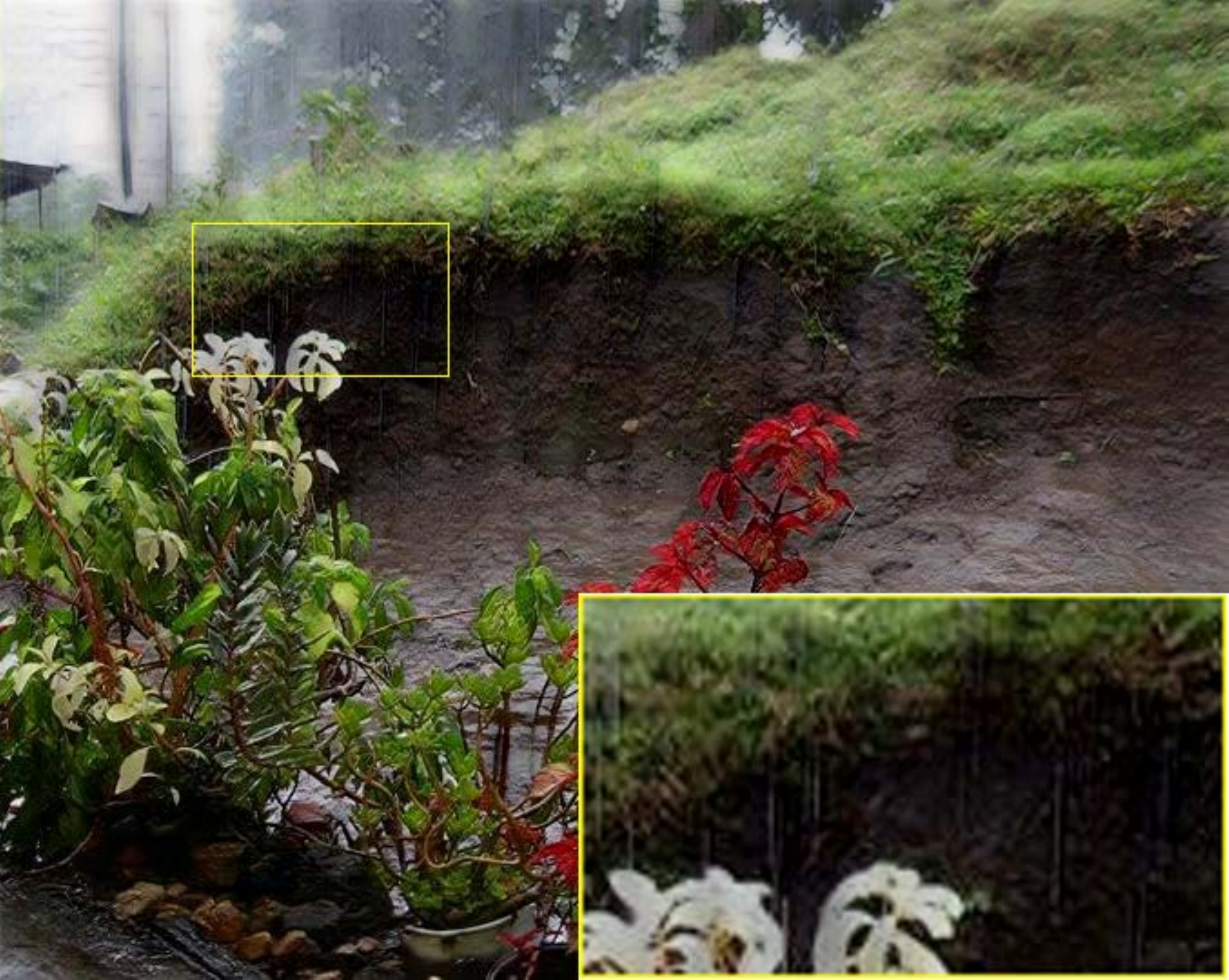}
\\
\includegraphics[width = 0.118\linewidth]{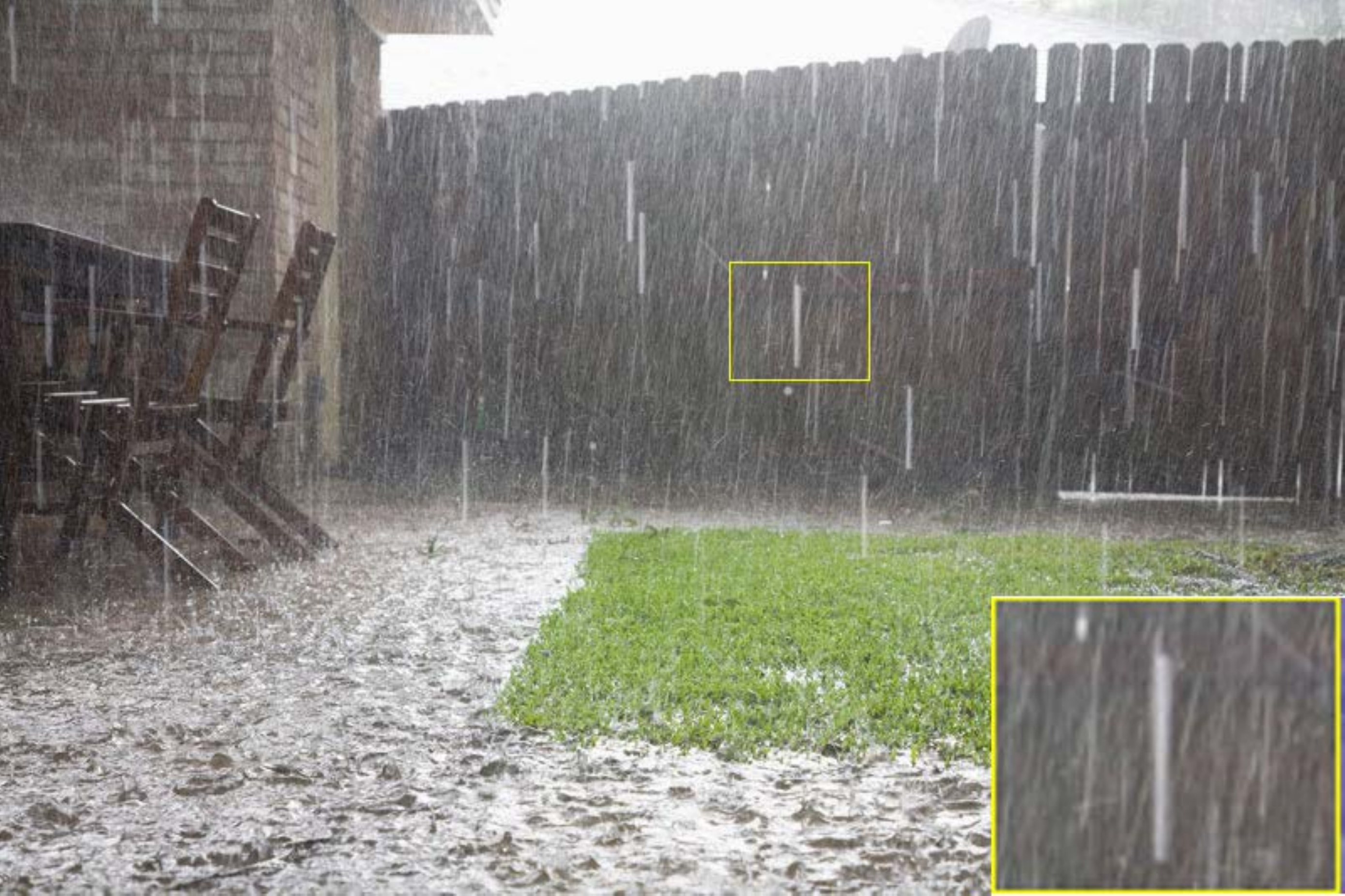} &\hspace{-4mm}
\includegraphics[width = 0.118\linewidth]{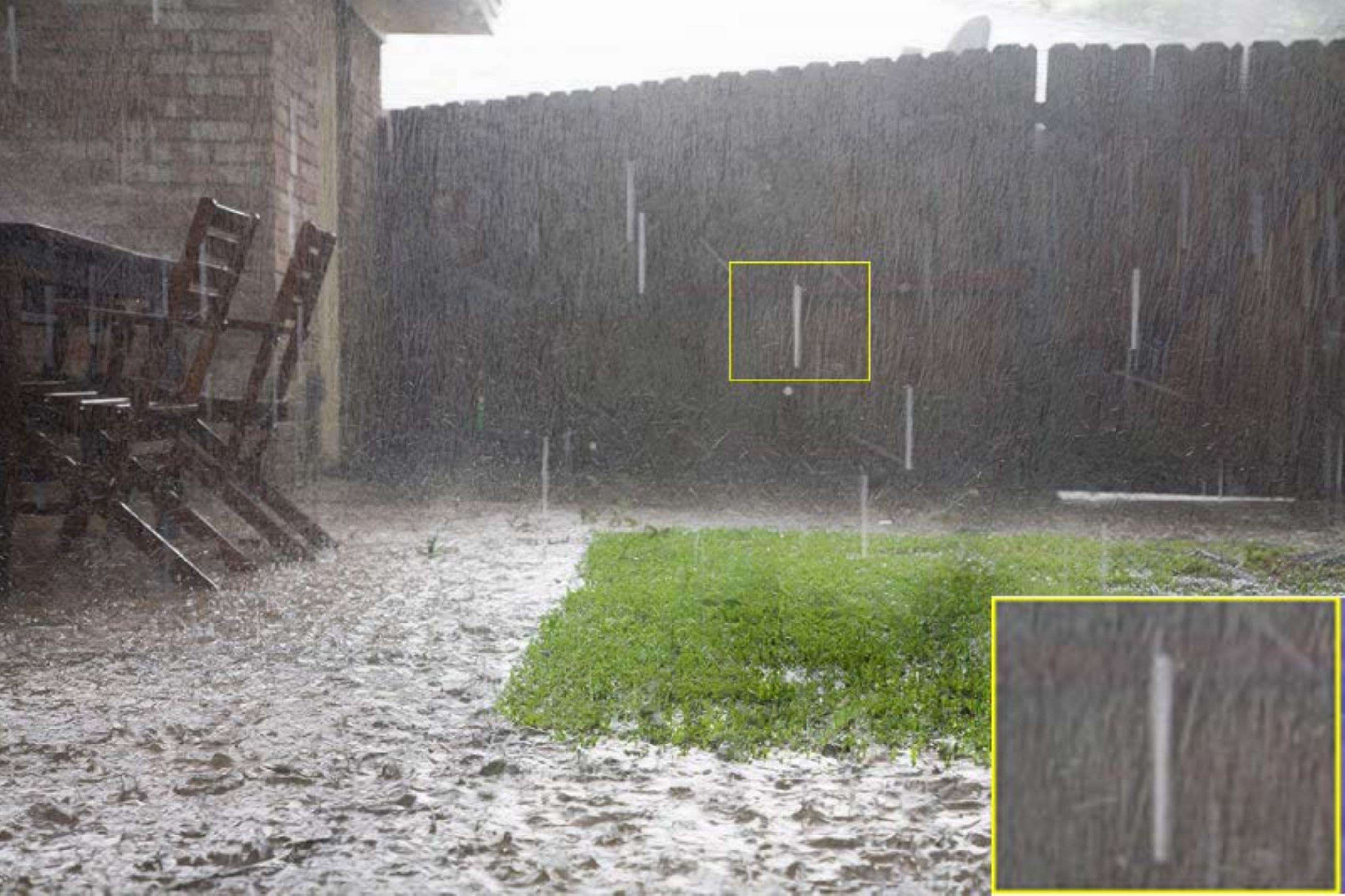} &\hspace{-4mm}
\includegraphics[width = 0.118\linewidth]{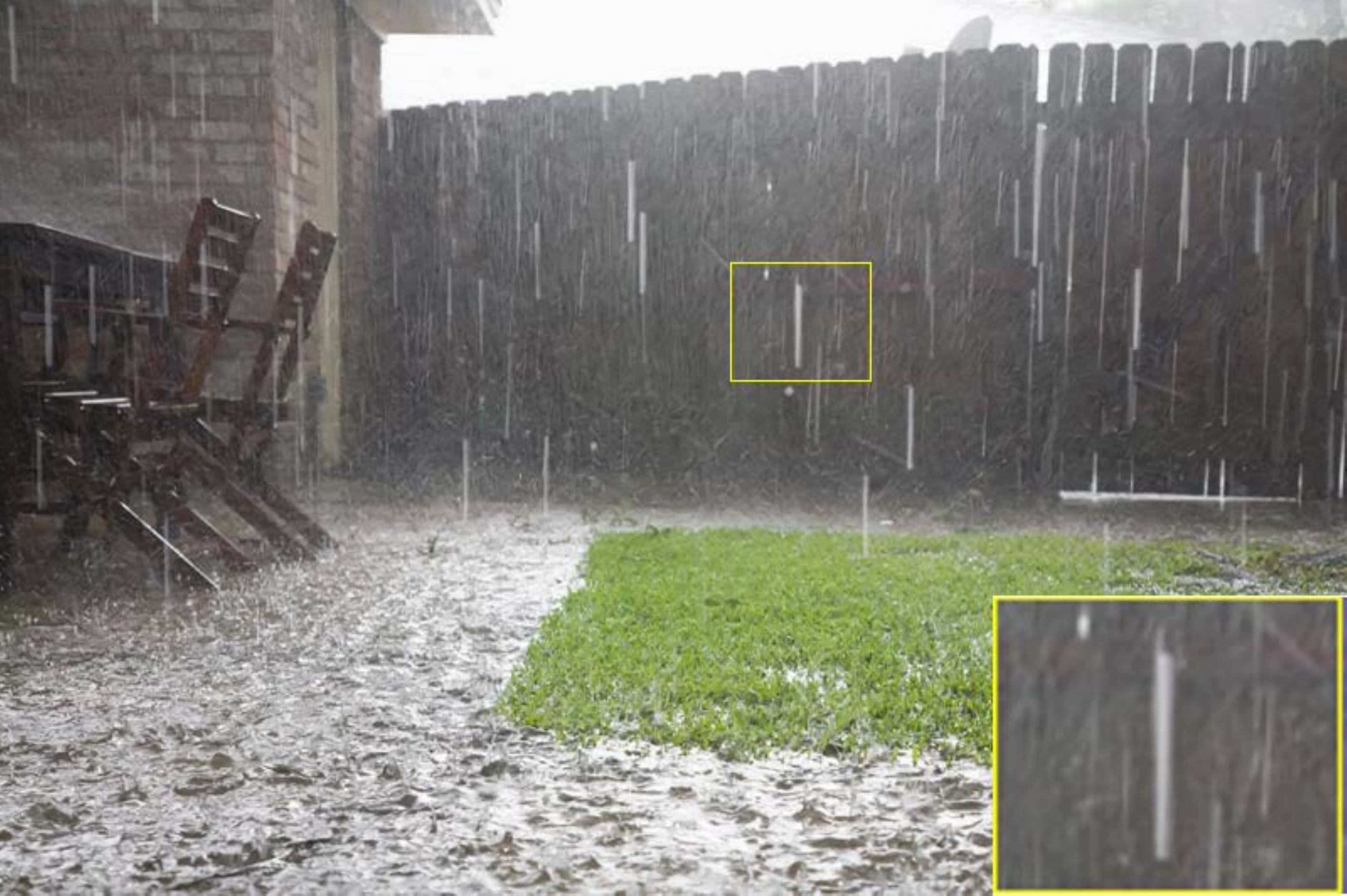} &\hspace{-4mm}
\includegraphics[width = 0.118\linewidth]{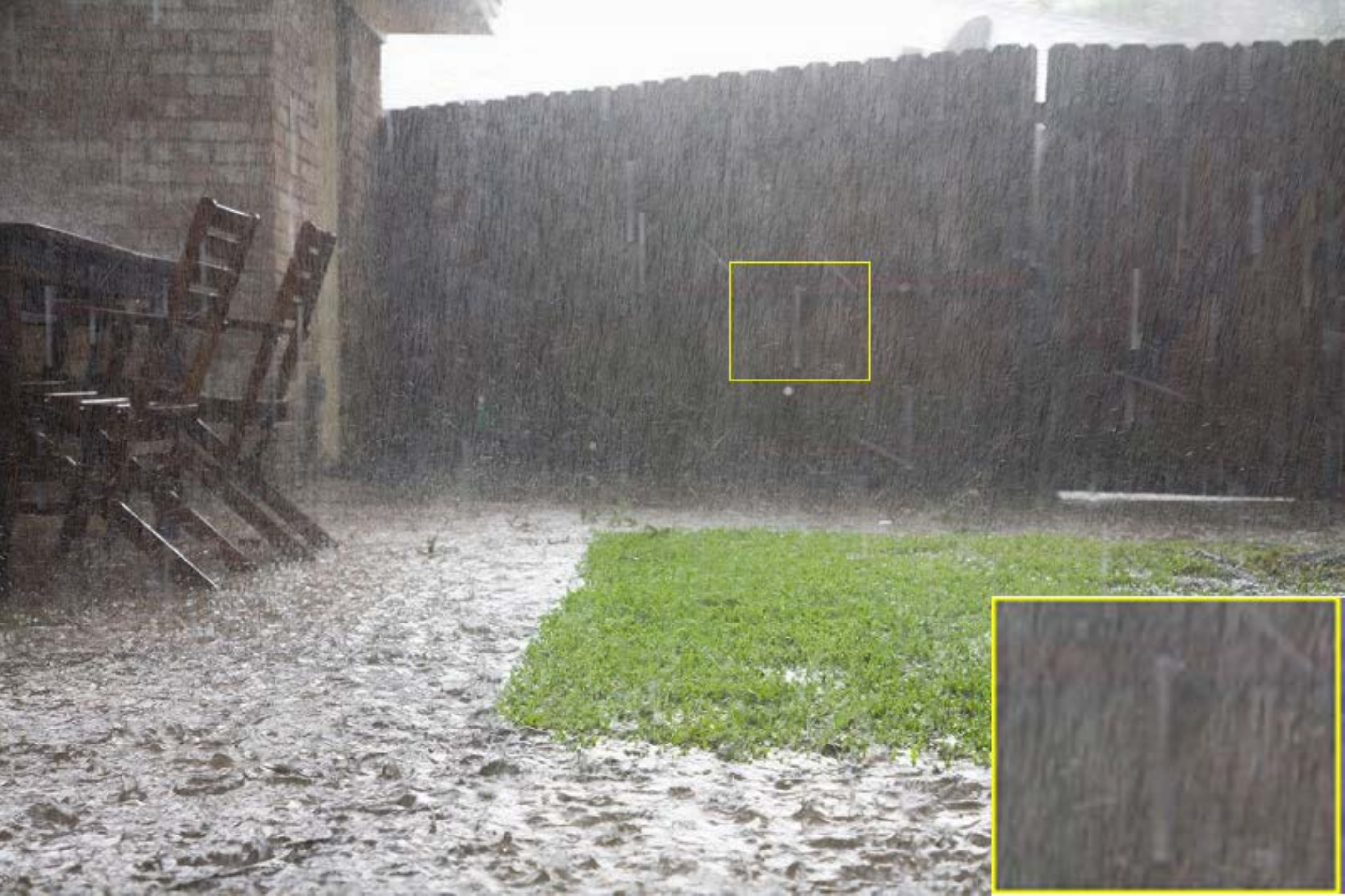} &\hspace{-4mm}
\includegraphics[width = 0.118\linewidth]{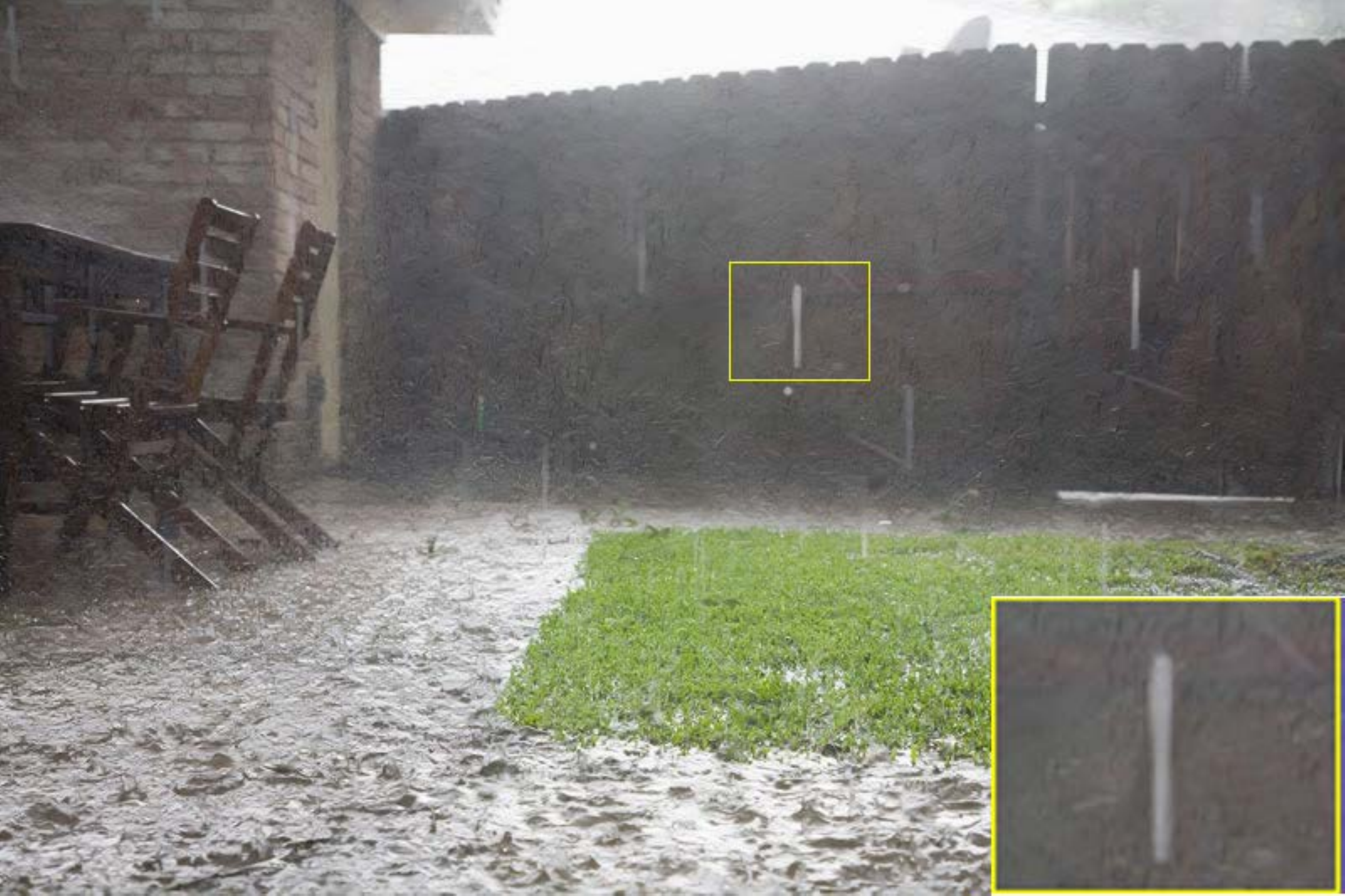}&\hspace{-4mm}
\includegraphics[width = 0.118\linewidth]{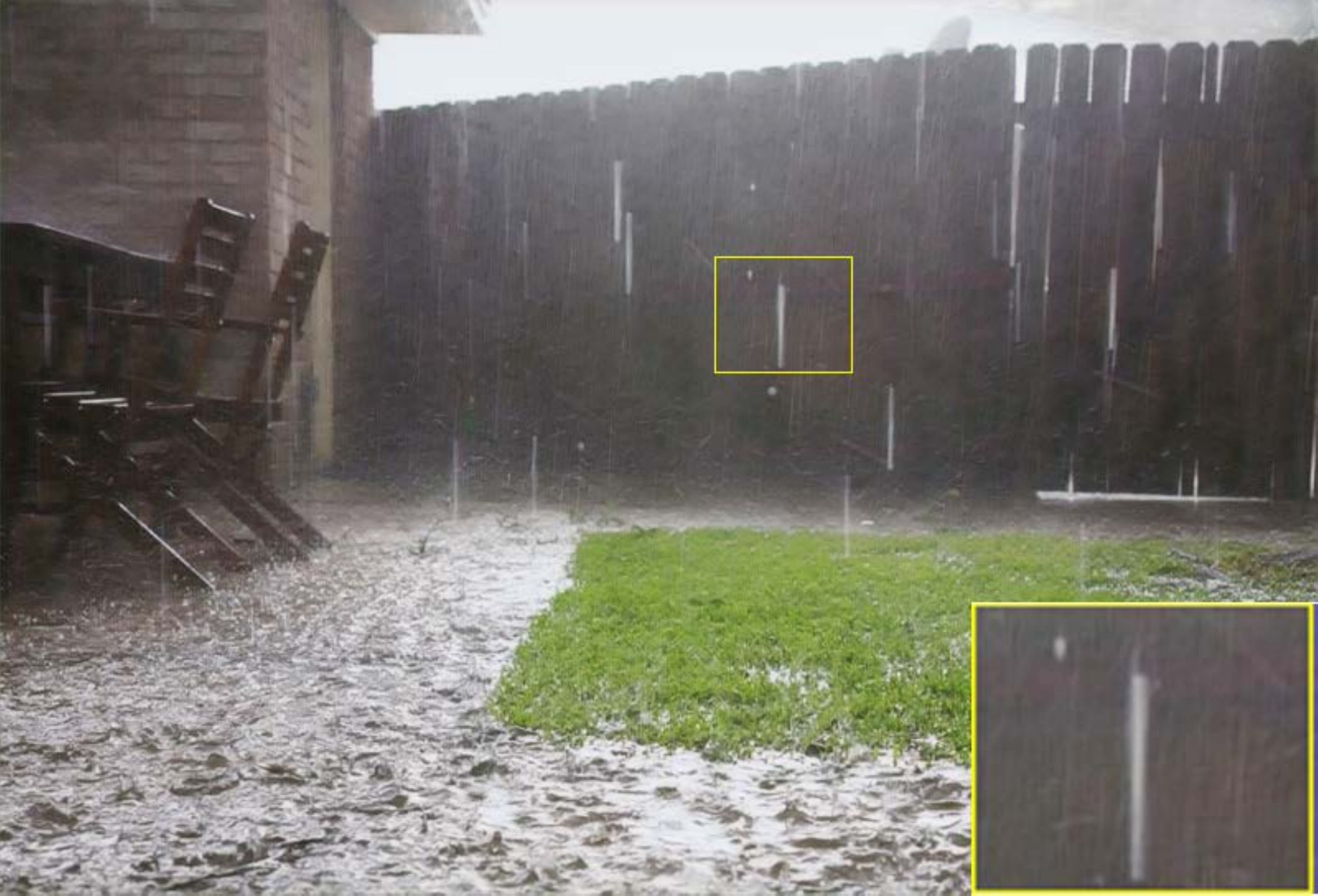}&\hspace{-4mm}
\includegraphics[width = 0.118\linewidth]{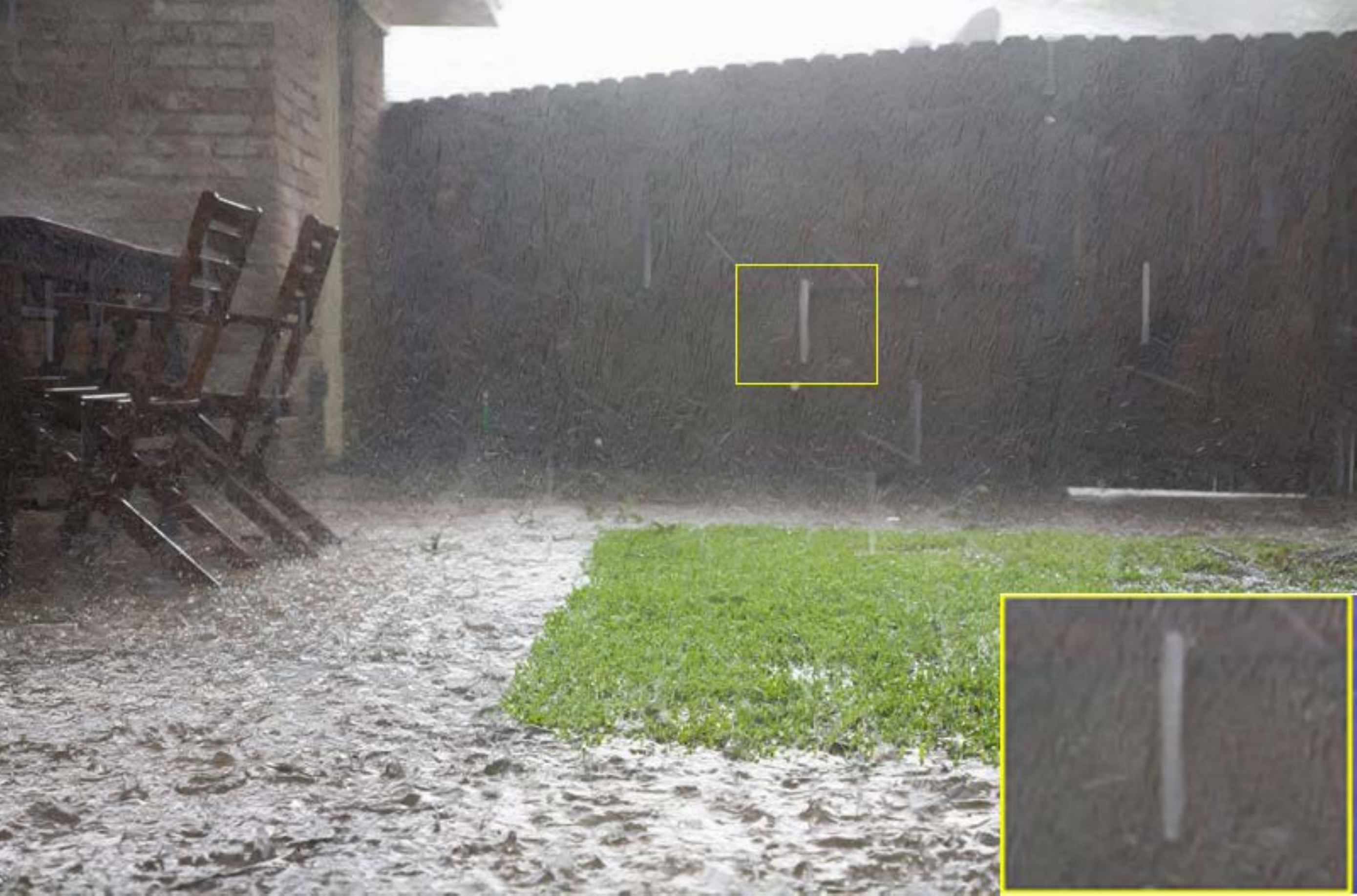}&\hspace{-4mm}
\includegraphics[width = 0.118\linewidth]{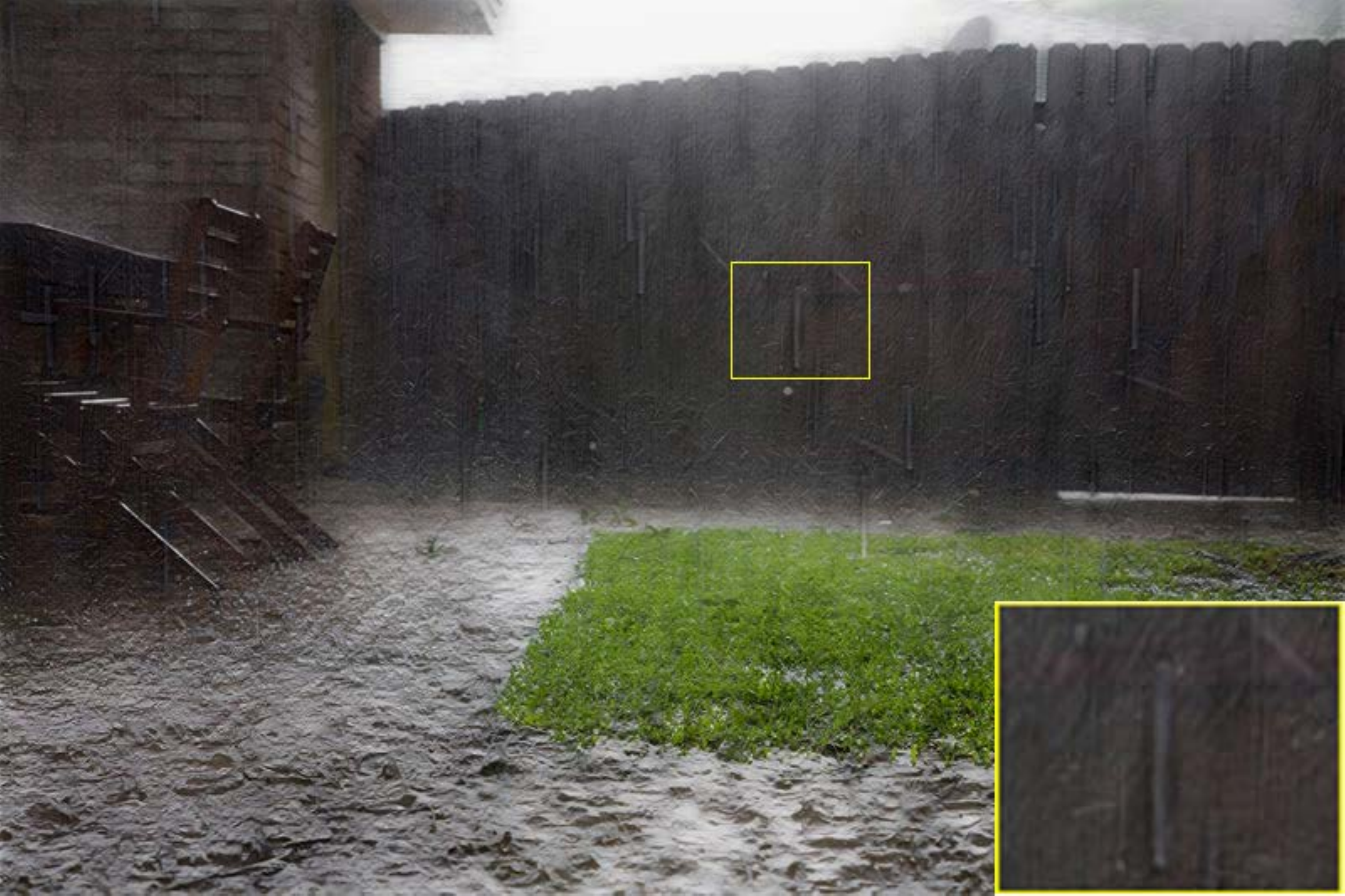}
\\
\includegraphics[width = 0.118\linewidth]{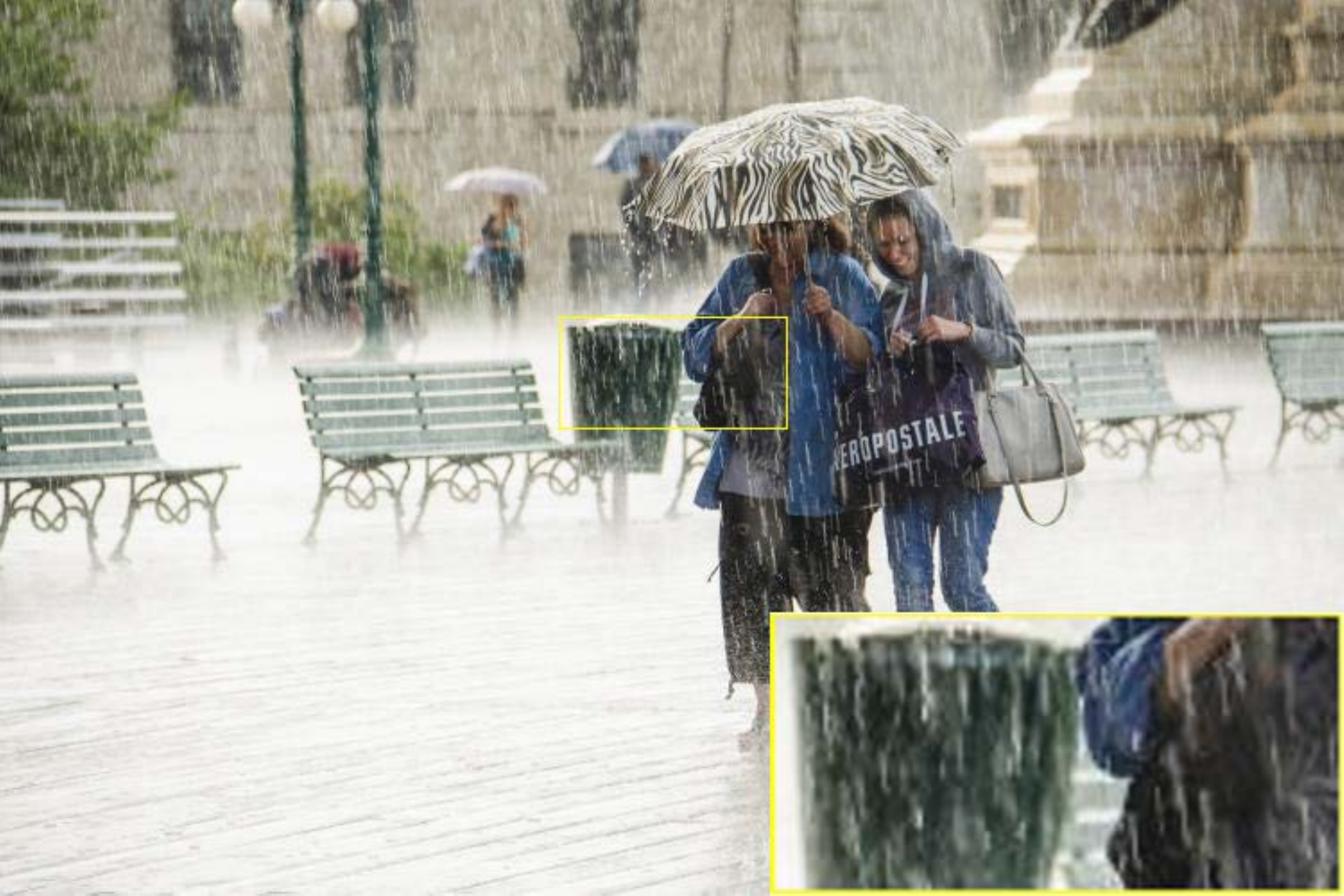} &\hspace{-4mm}
\includegraphics[width = 0.118\linewidth]{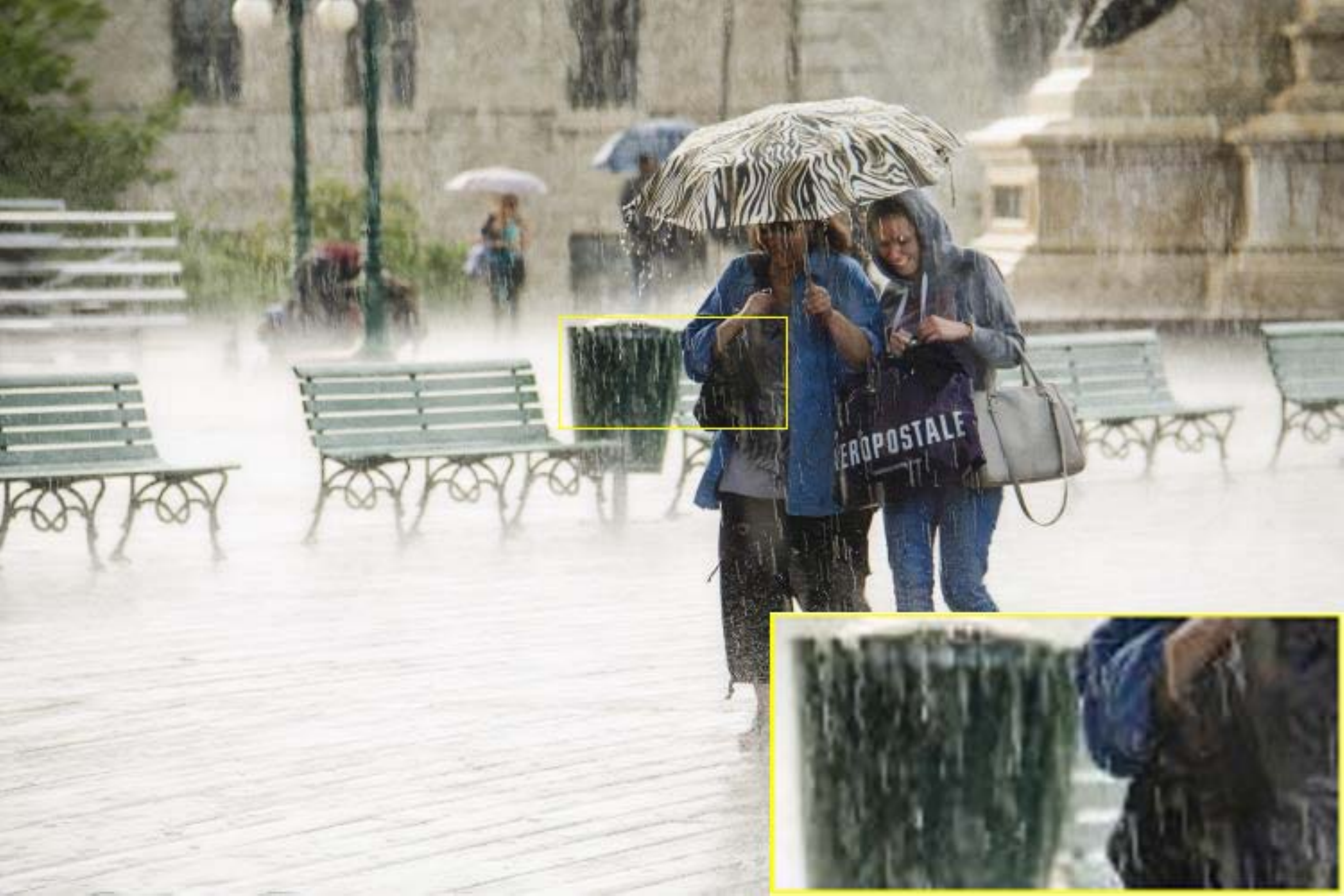} &\hspace{-4mm}
\includegraphics[width = 0.118\linewidth]{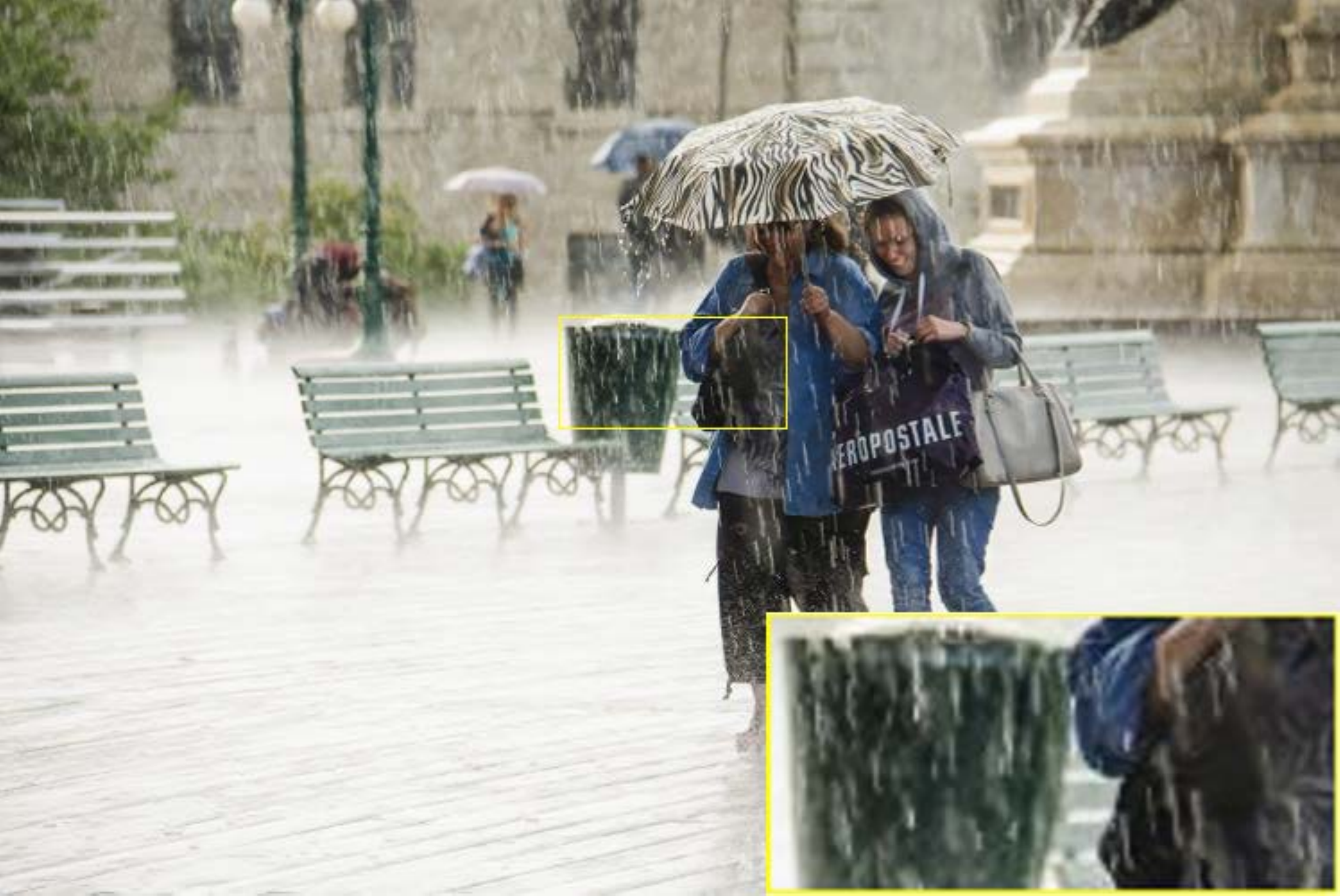} &\hspace{-4mm}
\includegraphics[width = 0.118\linewidth]{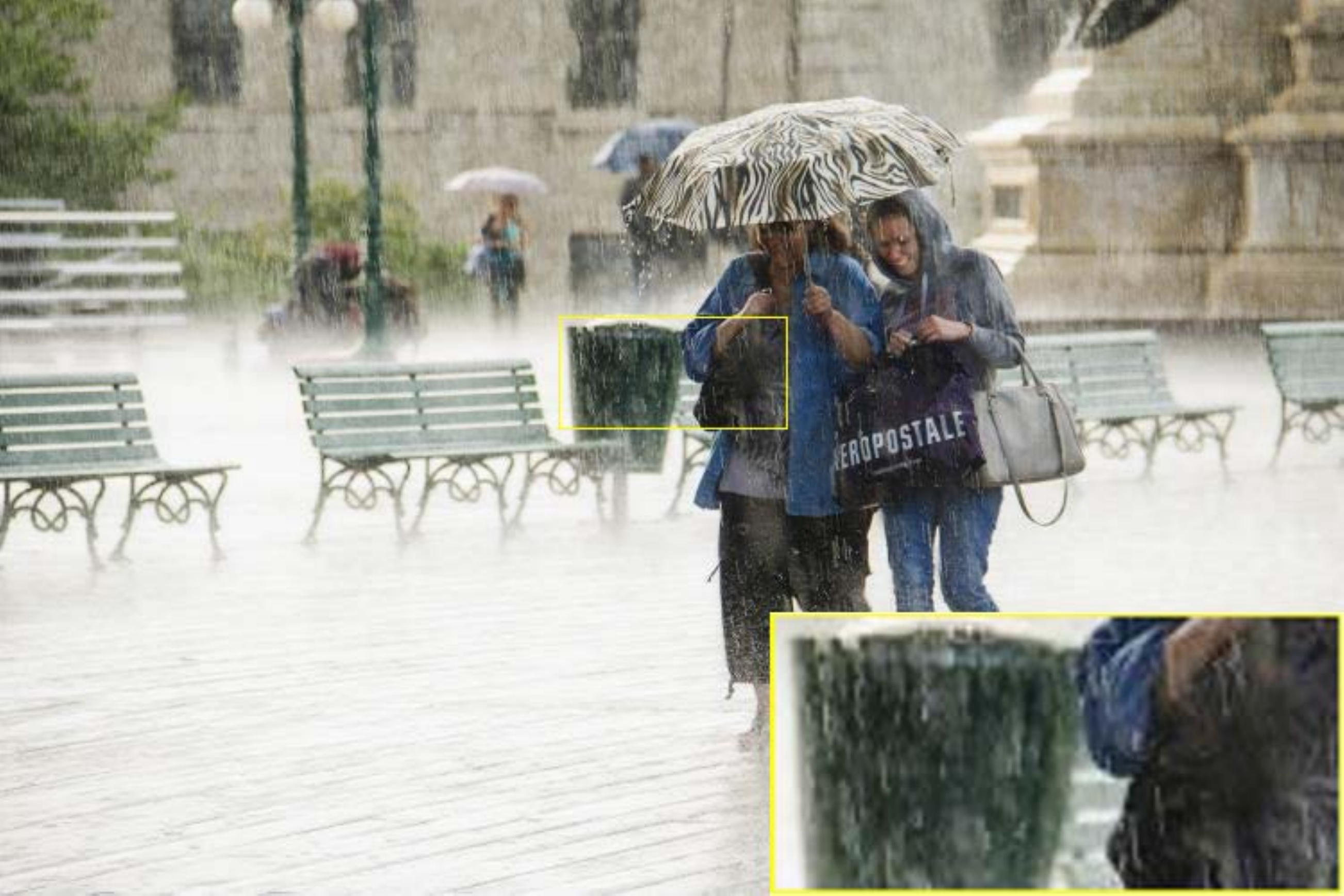} &\hspace{-4mm}
\includegraphics[width = 0.118\linewidth]{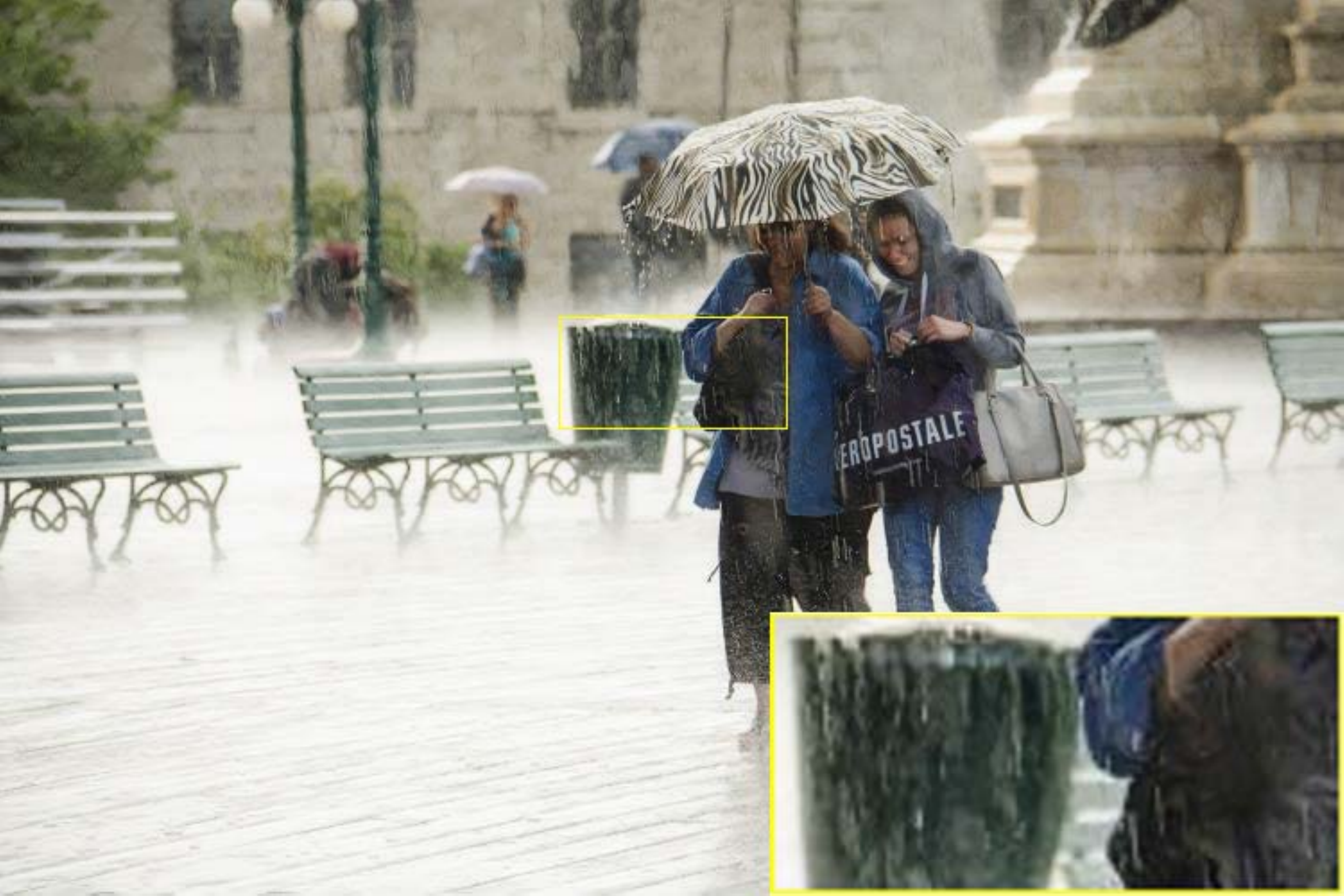}&\hspace{-4mm}
\includegraphics[width = 0.118\linewidth]{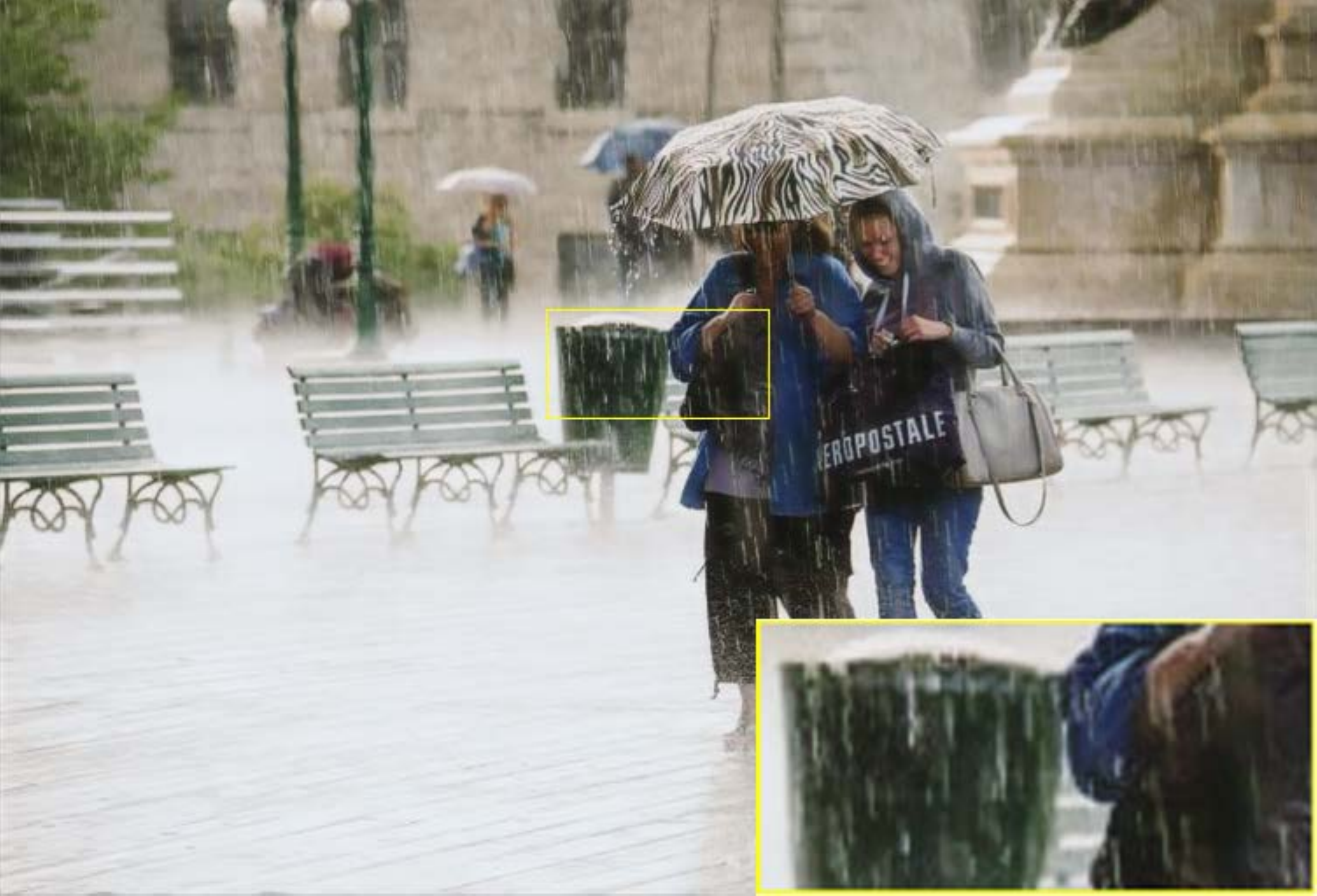}&\hspace{-4mm}
\includegraphics[width = 0.118\linewidth]{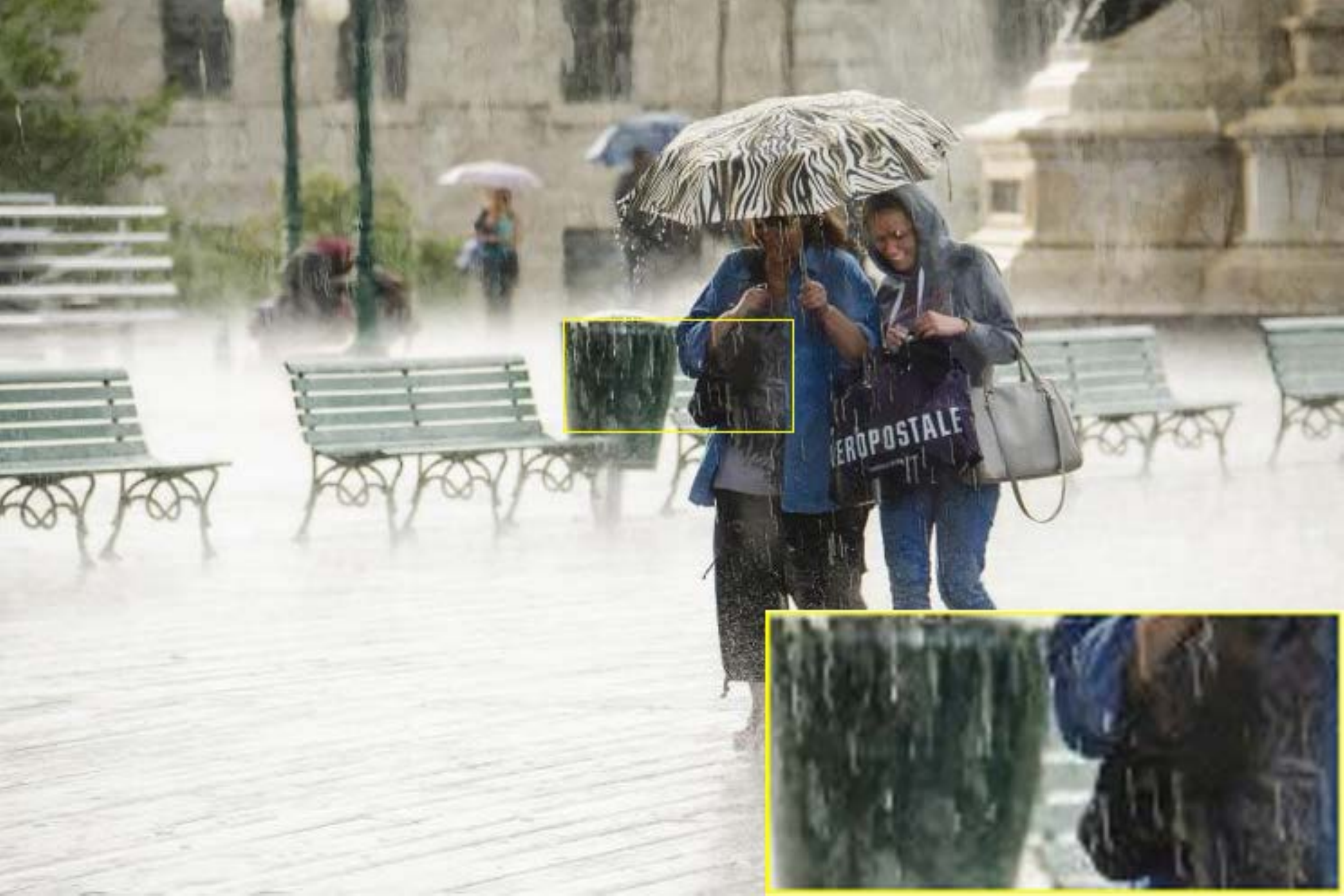}&\hspace{-4mm}
\includegraphics[width = 0.118\linewidth]{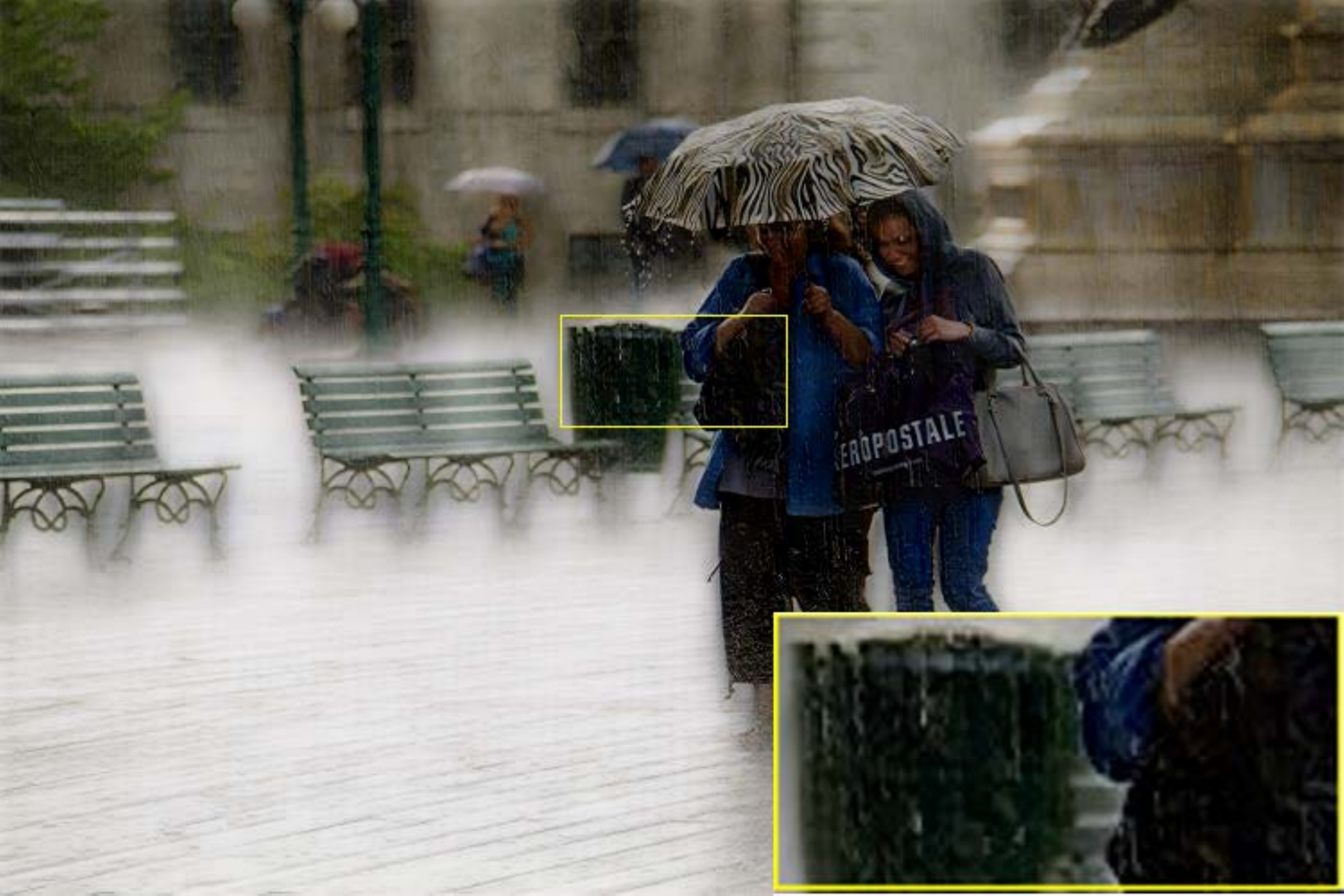}
\\
(a) Input  &\hspace{-4mm} (b) DCSFN&\hspace{-4mm} (c) MSPFN&\hspace{-4mm} (d) DRDNet&\hspace{-4mm} (e) RCDNet &\hspace{-4mm}(f) Syn2Real&\hspace{-4mm} (g) MPRNet&\hspace{-4mm}(h) Ours
\\
\end{tabular}
\end{center}
% \vspace{-2mm}
\caption{Comparisons with state-of-the-art methods on real-world images. Our proposed online-update learning approach is able to deal with various rainy conditions to better remove rain streaks and even haze and recover clearer images.
}
\label{fig:deraining-real-example}
% \vspace{-2mm}
\end{figure*}

\subsubsection{Comparison between Single-stream Learning and Collaborative Learning}
In Fig.~\ref{fig: Comparative results between single learning and collaborative progressive learning.}, we provide the comparison of the performance of each sub-network trained in single-stream learning and collaborative learning manners for image deraining. Single-stream learning means training the sub-networks $\mathcal{T}$, $\mathcal{M}$, and $\mathcal{B}$ independently. The results show that the sub-networks trained by collaborative learning preform much better than by single-stream learning. It can be also observed that the sub-network $\mathcal{T}$ trained by collaborative learning generates comparable results as the sub-network $\mathcal{B}$ trained by single-stream learning, while the size of $\mathcal{T}$ is two-thirds of $\mathcal{B}$.
\subsubsection{Further Ablation Study on Collaborative Learning and Multi-scale Compact Constraints}
Fig.~\ref{fig: Results on collaborative progressive learning and multi-scale compact constraint.} provides a further ablation study of collaborative learning and multi-scale compact constraints, where the learning curves of different variants are plotted.
The results show that both collaborative learning and multi-scale compact constraints are useful for improving the deraining performance. It can also be observed that without either collaborative learning or multi-scale compact constraints, the models perform worse than the cascaded network, which further demonstrates the effectiveness of the proposed external and internal learning manners for the deraining task.
\subsection{Results and Analysis on Real-world Datasets}
\begin{figure}[!t]
\begin{center}
\begin{tabular}{cccc}
\includegraphics[width = 0.24\linewidth]{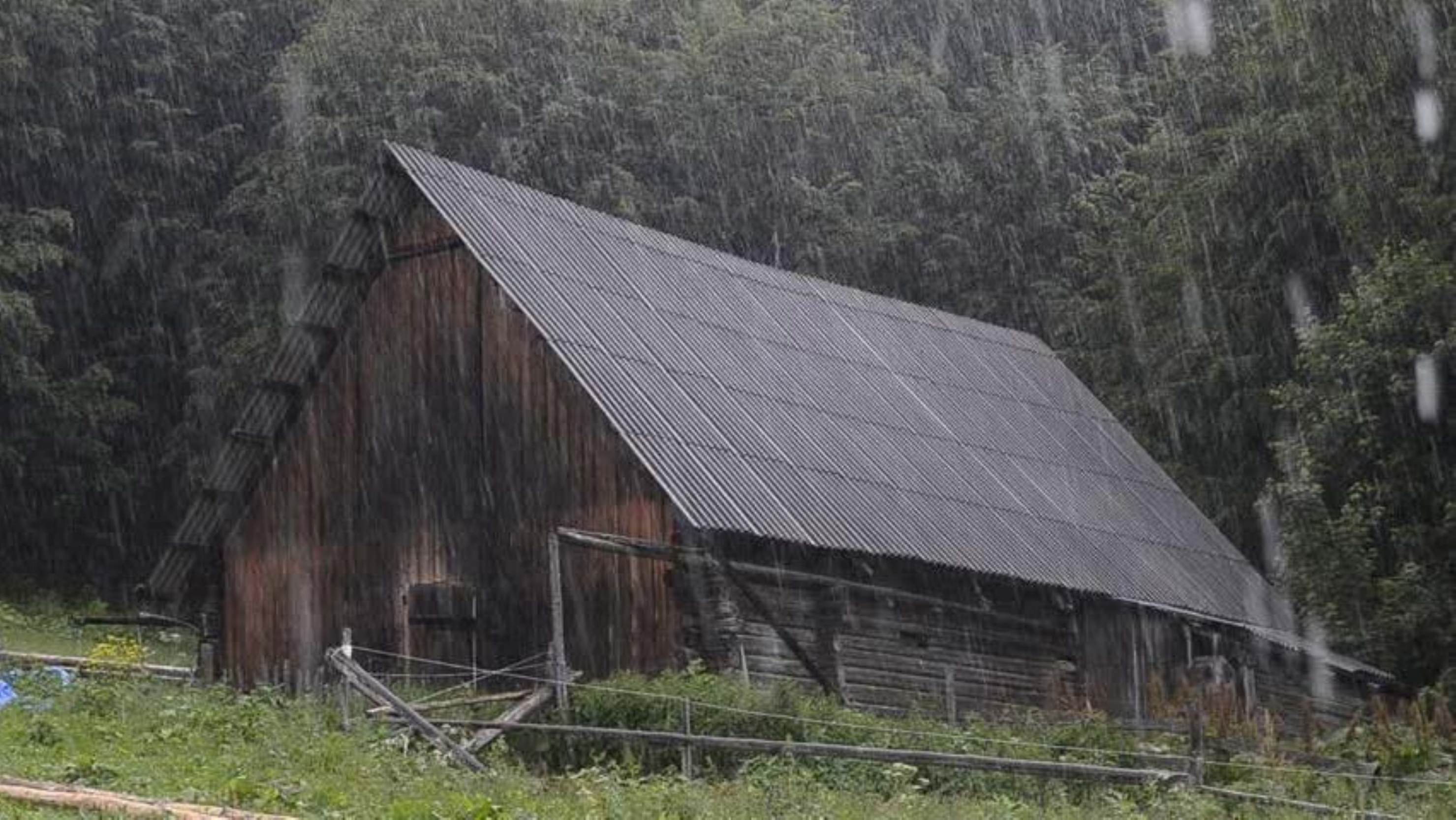} &\hspace{-4.5mm}
\includegraphics[width = 0.24\linewidth]{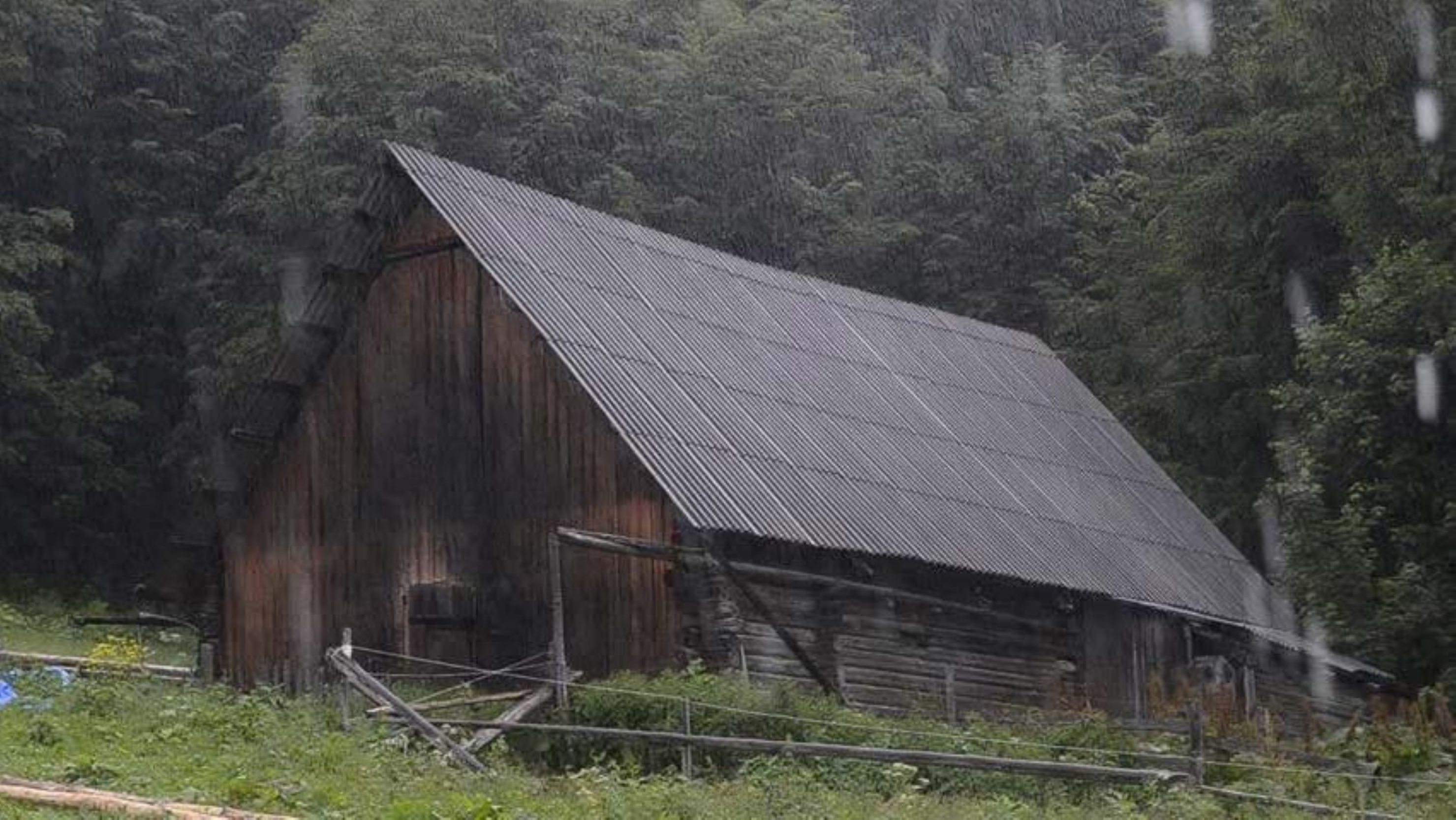} &\hspace{-4.5mm}
\includegraphics[width = 0.24\linewidth]{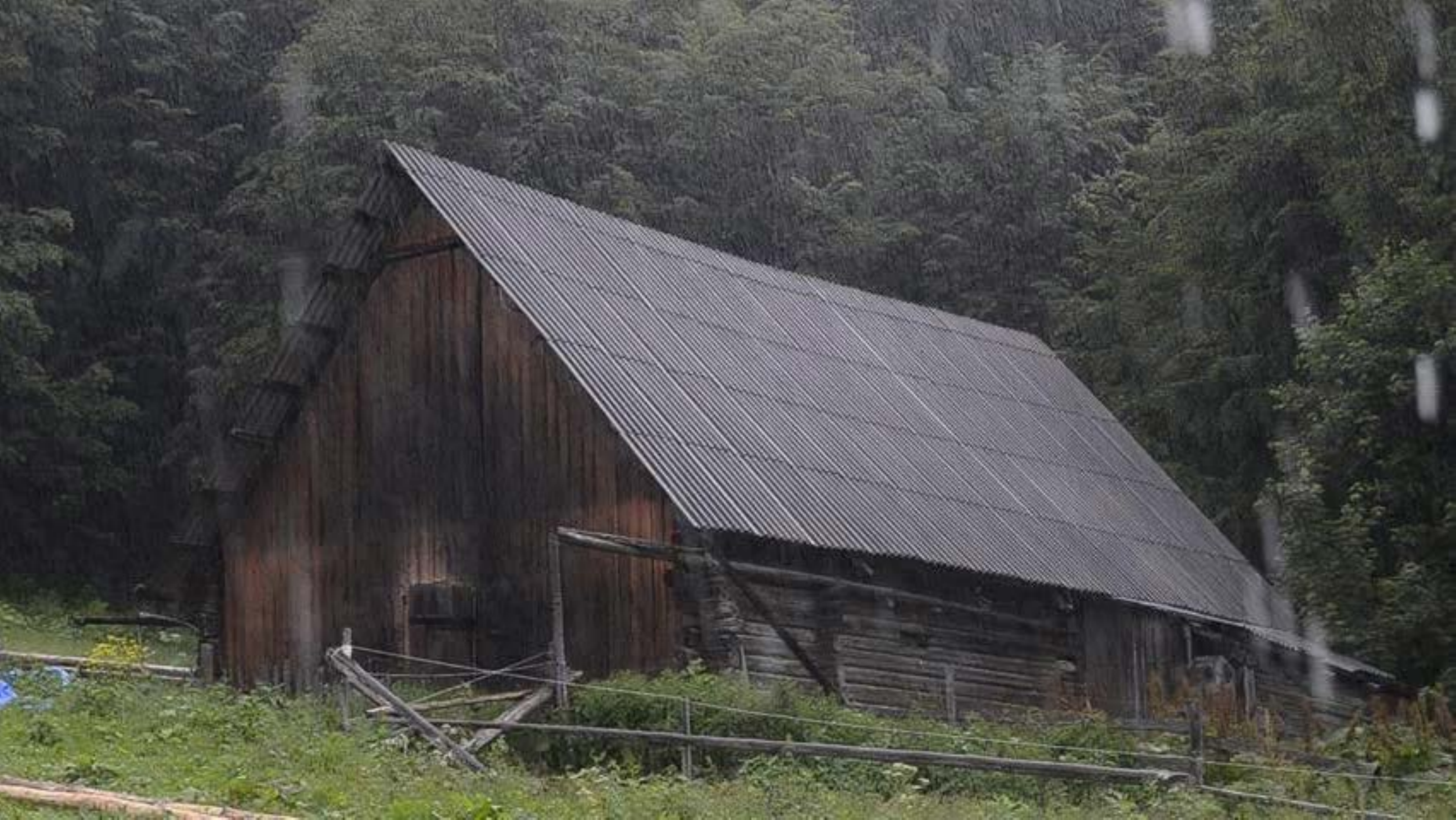} &\hspace{-4.5mm}
\includegraphics[width = 0.24\linewidth]{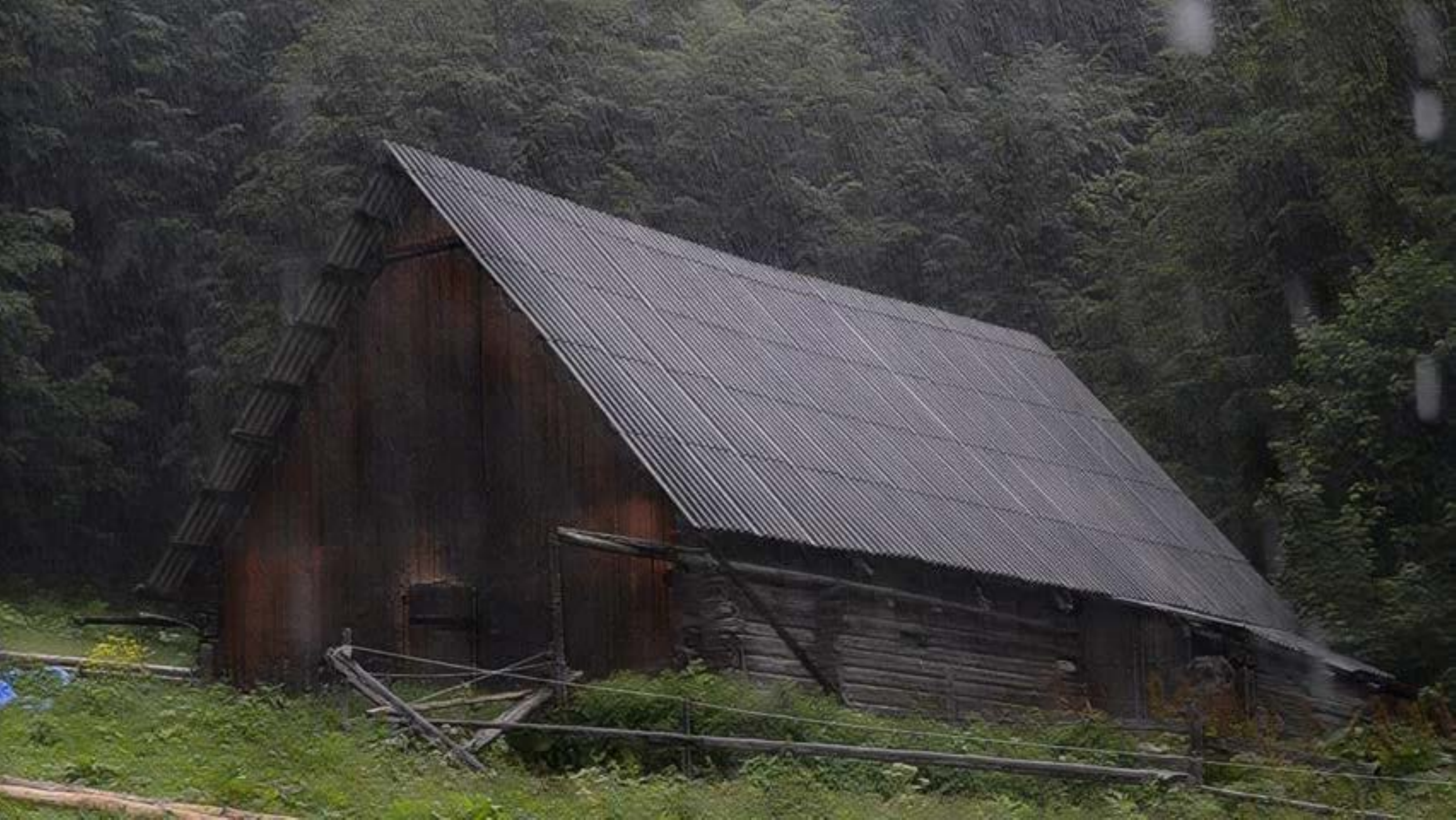}
\\
(a)  &\hspace{-4mm} (b)  &\hspace{-4.5mm} (c)  &\hspace{-4mm} (d)
\\
\end{tabular}
\end{center}
% \vspace{-2mm}
\caption{Ablation study of the proposed online-update learning approach.
(a) Input. (b) Only trained on synthetic data. (c) Directly fine-tuned on real-world images. (d) Online-update learning on real-world images.
}
\label{fig: Results on fine-tune manner.}
\end{figure}
\subsubsection{Comparisons with SOTAs on Real-world Images}
%
% Fig.~\ref{fig:deraining-real-example} presents the deraining results of our method and SOTAs on several challenging real-world images. One can see that our model is able to recover cleaner images with fewer rain streaks and finer details than others.

We further demonstrate the effectiveness of our method on the real-world dataset by  comparing with state-of-the-art methods.
Fig.~\ref{fig:deraining-real-example} presents the deraining results of several challenging cases. It can be observed that our method produces cleaner and clearer deraining results than others, demonstrating its effectiveness in removing rain streaks of real-world rainy images.
%while other methods always leave some  rain streaks.
%
\subsubsection{Effectiveness of the Online-update Learning Approach}
Fig.~\ref{fig: Results on fine-tune manner.} presents the ablation study of the online-update learning manner. Compared with only using synthetic data to train the model (Fig.~\ref{fig: Results on fine-tune manner.}(b)) or directly fine-tuning the model on real-world images without updating the pseudo ground truth (Fig.~\ref{fig: Results on fine-tune manner.}(c)), our proposed online-update learning manner (Fig.~\ref{fig: Results on fine-tune manner.}(d)) is able to further improve the deraining performance on real-world images, demonstrating its effectiveness.
%Hence, we can conclude that the proposed online-update learning scheme is a better training manner for real-world image deraining.
%
\subsubsection{Generality of the Online-update Learning Approach}
\begin{figure}[!t]
\begin{center}
\begin{tabular}{cccc}
\includegraphics[width = 0.24\linewidth]{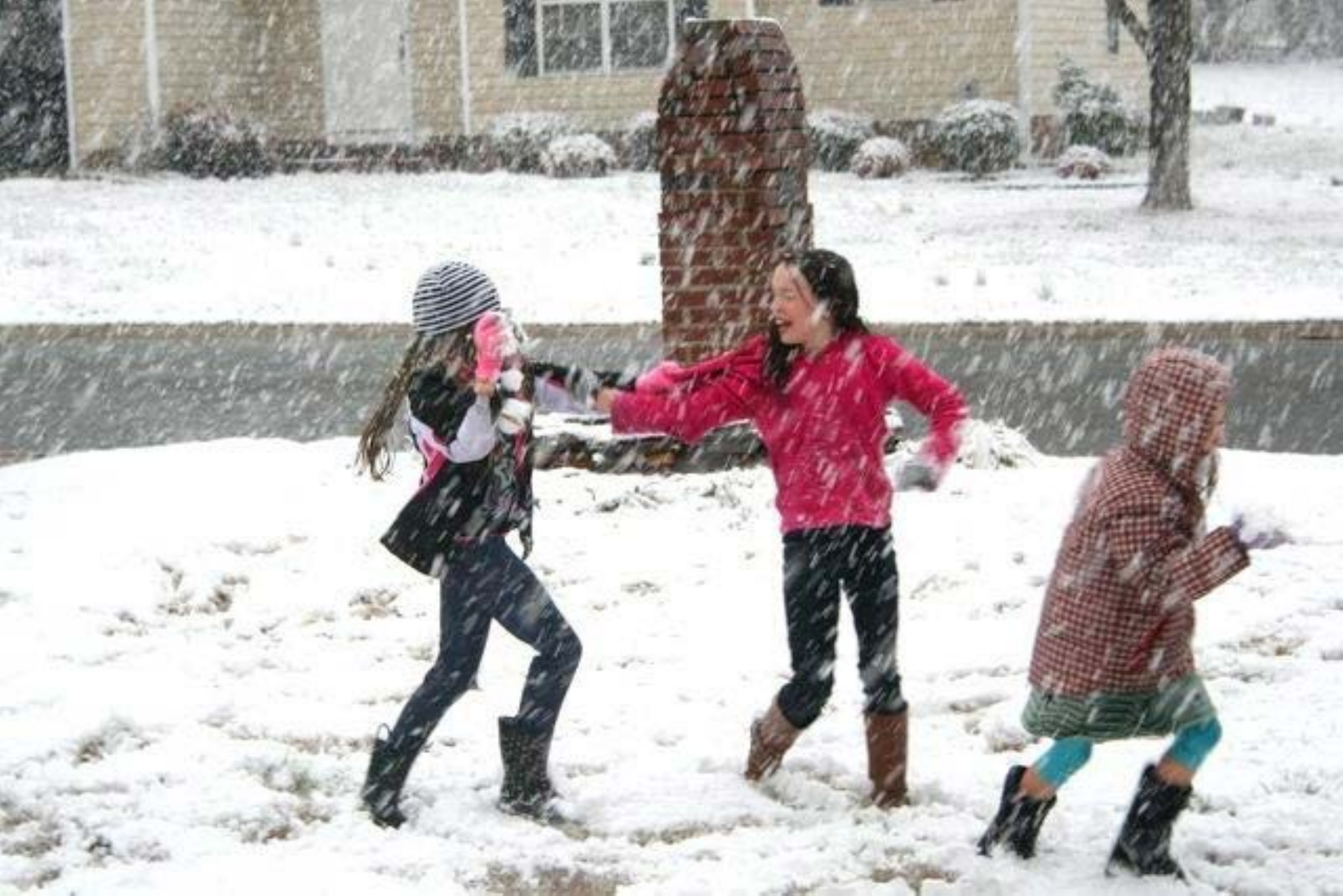} &\hspace{-4.5mm}
\includegraphics[width = 0.24\linewidth]{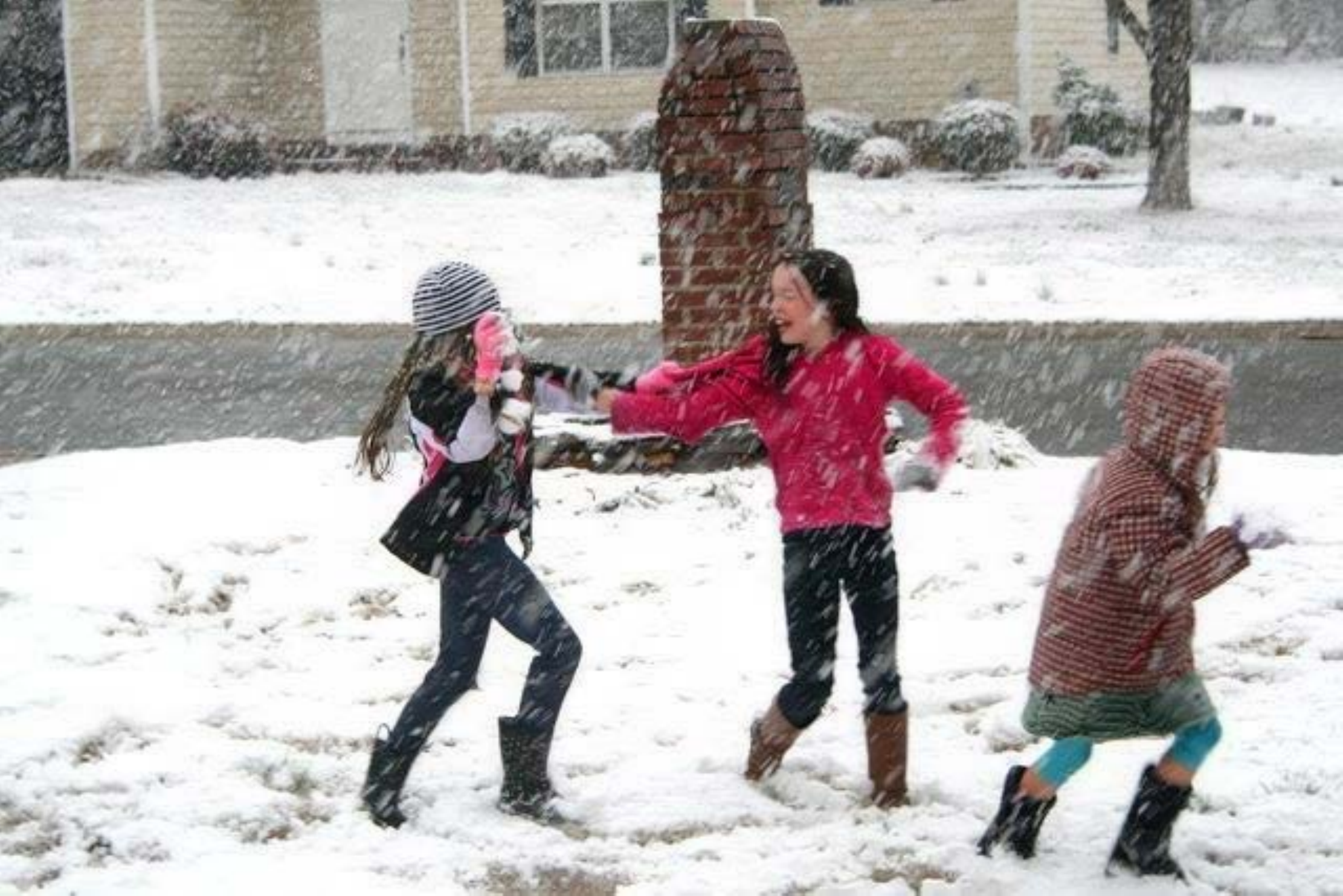} &\hspace{-4.5mm}
\includegraphics[width = 0.24\linewidth]{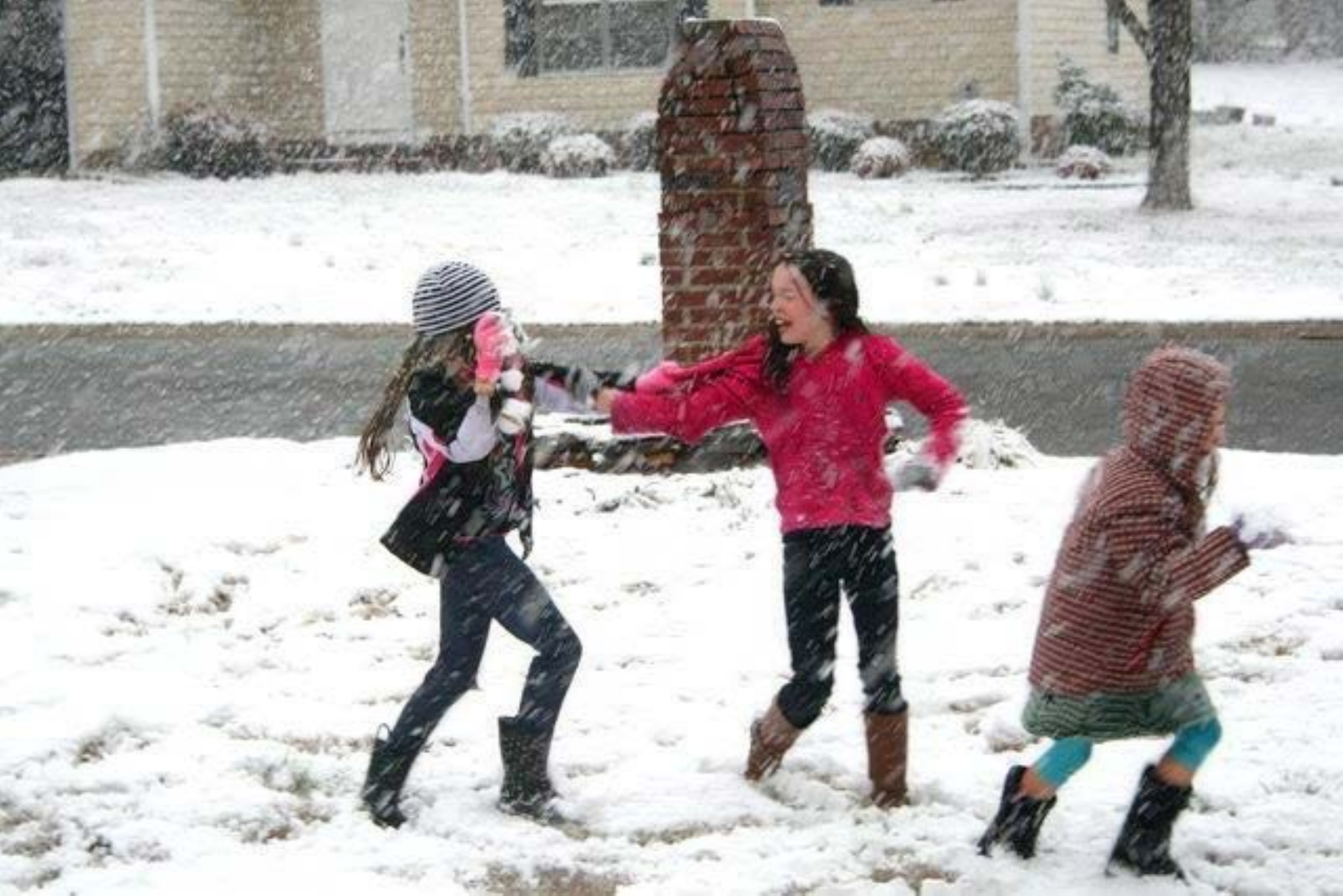} &\hspace{-4.5mm}
\includegraphics[width = 0.24\linewidth]{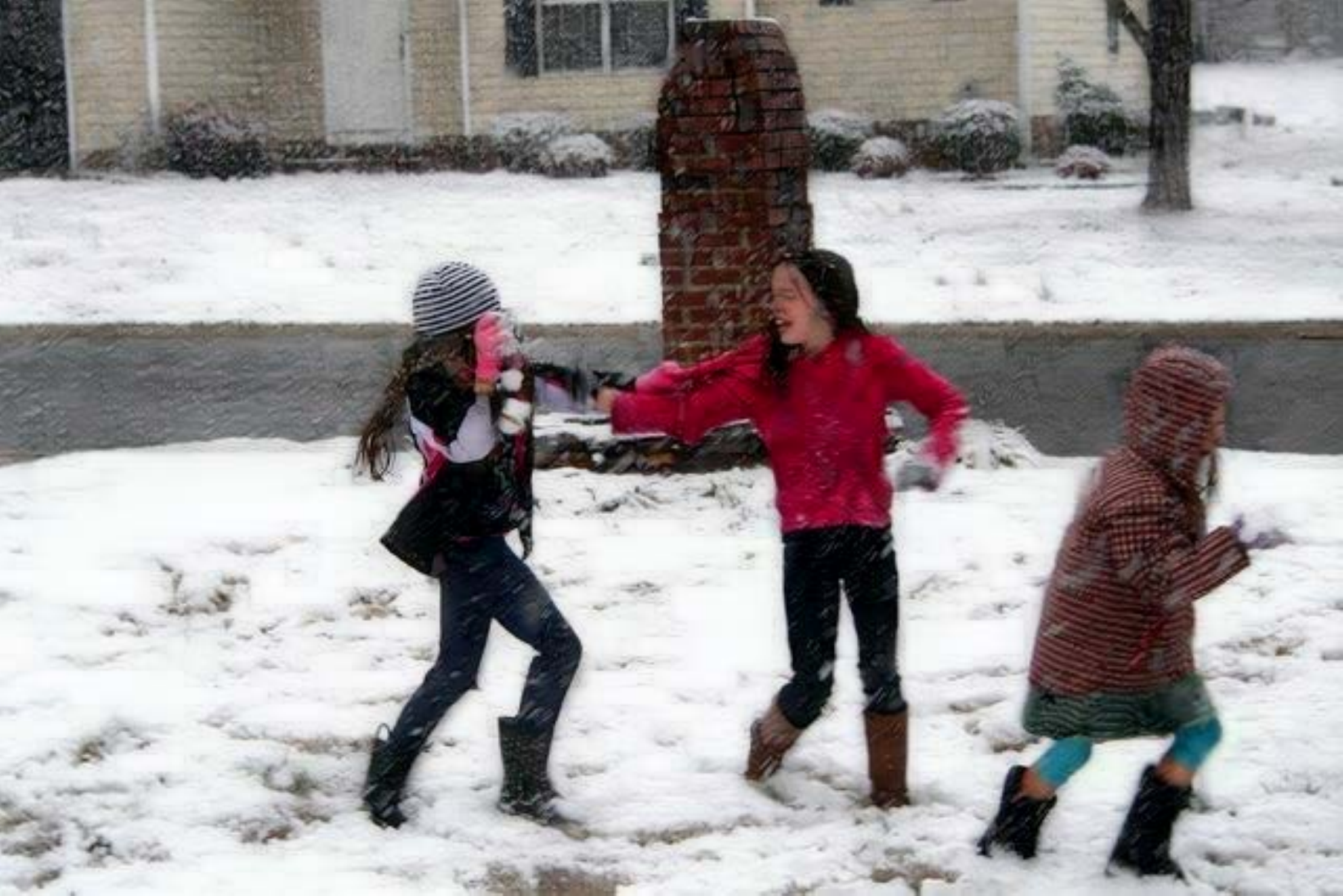}
\\
(a)  &\hspace{-4mm} (b)  &\hspace{-4.5mm} (c)   &\hspace{-4.5mm} (d)
\\
\end{tabular}
\end{center}
% \vspace{-2mm}
%\caption{Results on different learning manners of DCSFN~\cite{mm20_wang_dcsfn}.
\caption{Applying the proposed online-update learning approach on DCSFN~\cite{mm20_wang_dcsfn}.
(a) Input. (b) Only trained on synthetic data. (c) Directly fine-tuned on real-world images. (d) Online-update learning on real-world images.
}
\label{fig: Results on fine-tune manner of DCSFN.}
% \vspace{-5mm}
\end{figure}

To further demonstrate the effectiveness of the online-update learning approach for real-world image deraining, we apply it on a state-of-the-art method, DCSFN~\cite{mm20_wang_dcsfn}.
Similar to Fig.~\ref{fig: Results on fine-tune manner.}, Fig.~\ref{fig: Results on fine-tune manner of DCSFN.} presents the results on a real-world rainy image by different ways of learning.
It can be seen that the proposed online-update learning approach can be successfully applied to DCSFN to significantly improve its deraining performance on real-world images. Since the online-update learning approach is generic, we believe it be applied to many other existing methods to improve their performance in real-world image deraining.

\section{Conclusion}
%TODO!
In this paper, we have proposed a high-order collaborative network with multi-scale compact constraints to control the learning process in an external and internal manner for image deraining.
 We have further developed a bidirectional scale-content similarity mining module to learn useful features at different scales in a down-to-up and up-to-down way to facilitate rain streaks removal.
Finally, to improve the deraining performance on real-world images, we have proposed an effective online-update learning approach to fine-tune the deraining model on real-world rainy images in a self-supervised manner.
Extensive experiments show that the proposed model outperforms state-of-the-art methods on five public synthetic datasets and one real-world dataset.
% \vspace{-3mm}
\section*{Acknowledgements}
We would like to thank the anonymous reviewers for their
helpful comments.
This work was supported by the Fundamental Research Funds for the Central Universities (No. 30920041109),
and the Grant of DaSAIL Project P0030935 funded by PolyU/UGC.
%novel method to solve the single image deraining problem by designing a high-order collaborative learning manner and multi-scale compact constraints, which have been demonstrated can improve the deraining performance by controlling the learning process in external and internal manners.
% has been verified that can boost the deraining performance from external network manner.
% The proposed multi-scale compact constraint controls the feature spaces of a high-order collaborative network towards generate richer rainy features, which has been demonstrated to be effective for image deraining.
%The purposed bidirectional scale-content similarity mining as a fusion module can learn useful features at different scales in a down-to-up and up-to-down manner to facilitate rain removal.
%
%The proposed online-updated learning algorithm has been verified that it can improve the real-world image deraining performance with a large margin.
%
%Our extensive analysis and discussion have demonstrated that the proposed deraining algorithm outperforms eleven state-of-the-art approaches on five public datasets and one real-world dataset.

\bibliography{ijcai20}

\begin{thebibliography}{33}
\providecommand{\natexlab}[1]{#1}

\bibitem[{Chen and Hsu(2013)}]{derain_lowrank}
Chen, Y.; and Hsu, C. 2013.
\newblock A Generalized Low-Rank Appearance Model for Spatio-temporally
  Correlated Rain Streaks.
\newblock In \emph{{IEEE} {ICCV}}, 1968--1975.

\bibitem[{Deng et~al.(2020)Deng, Wei, Wang, Feng, Liang, Xie, Wang, and
  Wang}]{drd_cvpr20_deng}
Deng, S.; Wei, M.; Wang, J.; Feng, Y.; Liang, L.; Xie, H.; Wang, F.~L.; and
  Wang, M. 2020.
\newblock Detail-recovery Image Deraining via Context Aggregation Networks.
\newblock In \emph{{IEEE} {CVPR}}, 14548--14557.

\bibitem[{Fu et~al.(2017)Fu, Huang, Zeng, Huang, Ding, and
  Paisley}]{derain_ddn_fu}
Fu, X.; Huang, J.; Zeng, D.; Huang, Y.; Ding, X.; and Paisley, J. 2017.
\newblock Removing Rain from Single Images via a Deep Detail Network.
\newblock In \emph{{IEEE} {CVPR}}, 1715--1723.

\bibitem[{Hu et~al.(2019)Hu, Fu, Zhu, and Heng}]{derain_cvpr19_hu}
Hu, X.; Fu, C.; Zhu, L.; and Heng, P. 2019.
\newblock Depth-Attentional Features for Single-Image Rain Removal.
\newblock In \emph{{IEEE} {CVPR}}, 8022--8031.

\bibitem[{Huynh-Thu and Ghanbari(2008)}]{PSNR_thu}
Huynh-Thu, Q.; and Ghanbari, M. 2008.
\newblock Scope of validity of PSNR in image/video quality assessment.
\newblock \emph{Electronics Letters}, 44(13): 800--801.

\bibitem[{Jiang et~al.(2020)Jiang, Wang, Yi, Chen, Huang, Luo, Ma, and
  Jiang}]{cvpr20_jiang_mspfn}
Jiang, K.; Wang, Z.; Yi, P.; Chen, C.; Huang, B.; Luo, Y.; Ma, J.; and Jiang,
  J. 2020.
\newblock Multi-Scale Progressive Fusion Network for Single Image Deraining.
\newblock In \emph{{IEEE} {CVPR}}, 8343--8352.

\bibitem[{Kang, Lin, and Fu(2012)}]{derain_id_kang}
Kang, L.; Lin, C.; and Fu, Y. 2012.
\newblock Automatic Single-Image-Based Rain Streaks Removal via Image
  Decomposition.
\newblock \emph{{IEEE} TIP}, 21(4): 1742--1755.

\bibitem[{Kingma and Ba(2015)}]{adam}
Kingma, D.~P.; and Ba, J. 2015.
\newblock Adam: {A} Method for Stochastic Optimization.
\newblock In \emph{ICLR}.

\bibitem[{Li et~al.(2018{\natexlab{a}})Li, He, Zhang, Chang, Dong, and
  Lin}]{derain_nledn_li}
Li, G.; He, X.; Zhang, W.; Chang, H.; Dong, L.; and Lin, L. 2018{\natexlab{a}}.
\newblock Non-locally Enhanced Encoder-Decoder Network for Single Image
  De-raining.
\newblock In \emph{{ACM} {MM}}, 1056--1064.

\bibitem[{Li et~al.(2019)Li, Araujo, Ren, Wang, Tokuda, Junior, Cesar{-}Junior,
  Zhang, Guo, and Cao}]{deraining-benchmark-analysis}
Li, S.; Araujo, I.~B.; Ren, W.; Wang, Z.; Tokuda, E.~K.; Junior, R.~H.;
  Cesar{-}Junior, R.; Zhang, J.; Guo, X.; and Cao, X. 2019.
\newblock Single Image Deraining: {A} Comprehensive Benchmark Analysis.
\newblock In \emph{{IEEE} {CVPR}}, 3838--3847.

\bibitem[{Li et~al.(2018{\natexlab{b}})Li, Wu, Lin, Liu, and
  Zha}]{derain_rescan_li}
Li, X.; Wu, J.; Lin, Z.; Liu, H.; and Zha, H. 2018{\natexlab{b}}.
\newblock Recurrent Squeeze-and-Excitation Context Aggregation Net for Single
  Image Deraining.
\newblock In \emph{ECCV}, 262--277.

\bibitem[{Li et~al.(2016)Li, Tan, Guo, Lu, and Brown}]{derain_lp_li}
Li, Y.; Tan, R.~T.; Guo, X.; Lu, J.; and Brown, M.~S. 2016.
\newblock Rain Streak Removal Using Layer Priors.
\newblock In \emph{{IEEE} {CVPR}}, 2736--2744.

\bibitem[{Luo, Xu, and Ji(2015)}]{derain_dsc_luo}
Luo, Y.; Xu, Y.; and Ji, H. 2015.
\newblock Removing Rain from a Single Image via Discriminative Sparse Coding.
\newblock In \emph{{IEEE} {ICCV}}, 3397--3405.

\bibitem[{Mei et~al.(2020{\natexlab{a}})Mei, Fan, Zhang, Yu, Zhou, Liu, Fu,
  Huang, and Shi}]{mei2020pyramid}
Mei, Y.; Fan, Y.; Zhang, Y.; Yu, J.; Zhou, Y.; Liu, D.; Fu, Y.; Huang, T.~S.;
  and Shi, H. 2020{\natexlab{a}}.
\newblock Pyramid Attention Networks for Image Restoration.
\newblock \emph{arXiv preprint arXiv:2004.13824}.

\bibitem[{Mei et~al.(2020{\natexlab{b}})Mei, Fan, Zhou, Huang, Huang, and
  Shi}]{sr_cross_scale_mining_Mei}
Mei, Y.; Fan, Y.; Zhou, Y.; Huang, L.; Huang, T.~S.; and Shi, H.
  2020{\natexlab{b}}.
\newblock Image Super-Resolution With Cross-Scale Non-Local Attention and
  Exhaustive Self-Exemplars Mining.
\newblock In \emph{{IEEE} {CVPR}}, 5689--5698.

\bibitem[{Mei et~al.(2020{\natexlab{c}})Mei, Fan, Zhou, Huang, Huang, and
  Shi}]{Mei_sisr_mining}
Mei, Y.; Fan, Y.; Zhou, Y.; Huang, L.; Huang, T.~S.; and Shi, H.
  2020{\natexlab{c}}.
\newblock Image Super-Resolution With Cross-Scale Non-Local Attention and
  Exhaustive Self-Exemplars Mining.
\newblock In \emph{{IEEE} {CVPR}}, 5689--5698.

\bibitem[{Pan et~al.(2021)Pan, Dong, Liu, Zhang, Ren, Tang, Tai, and
  Yang}]{PhysicsGAN-pan}
Pan, J.; Dong, J.; Liu, Y.; Zhang, J.; Ren, J. S.~J.; Tang, J.; Tai, Y.; and
  Yang, M. 2021.
\newblock Physics-Based Generative Adversarial Models for Image Restoration and
  Beyond.
\newblock \emph{{IEEE} TPAMI}, 43(7): 2449--2462.

\bibitem[{Pan et~al.(2018)Pan, Liu, Sun, Zhang, Liu, Ren, Li, Tang, Lu, Tai,
  and Yang}]{dualcnn-pan}
Pan, J.; Liu, S.; Sun, D.; Zhang, J.; Liu, Y.; Ren, J. S.~J.; Li, Z.; Tang, J.;
  Lu, H.; Tai, Y.; and Yang, M. 2018.
\newblock Learning Dual Convolutional Neural Networks for Low-Level Vision.
\newblock In \emph{{IEEE} CVPR}, 3070--3079.

\bibitem[{Ren et~al.(2019)Ren, Zuo, Hu, Zhu, and
  Meng}]{derain_prenet_Ren_2019_CVPR}
Ren, D.; Zuo, W.; Hu, Q.; Zhu, P.; and Meng, D. 2019.
\newblock Progressive Image Deraining Networks: {A} Better and Simpler
  Baseline.
\newblock In \emph{{IEEE} {CVPR}}, 3937--3946.

\bibitem[{Wang et~al.(2020{\natexlab{a}})Wang, Wu, Su, and
  Chen}]{mm20_wang_jdnet}
Wang, C.; Wu, Y.; Su, Z.; and Chen, J. 2020{\natexlab{a}}.
\newblock Joint Self-Attention and Scale-Aggregation for Self-Calibrated
  Deraining Network.
\newblock In \emph{ACM MM}, 2517--2525.

\bibitem[{Wang et~al.(2020{\natexlab{b}})Wang, Xing, Wu, Su, and
  Chen}]{mm20_wang_dcsfn}
Wang, C.; Xing, X.; Wu, Y.; Su, Z.; and Chen, J. 2020{\natexlab{b}}.
\newblock {DCSFN:} Deep Cross-scale Fusion Network for Single Image Rain
  Removal.
\newblock In \emph{ACM MM}, 1643--1651.

\bibitem[{Wang et~al.(2019{\natexlab{a}})Wang, Zhang, Su, Wu, Yao, and
  Wang}]{spic-wang-derain}
Wang, C.; Zhang, M.; Su, Z.; Wu, Y.; Yao, G.; and Wang, H. 2019{\natexlab{a}}.
\newblock Learning a multi-level guided residual network for single image
  deraining.
\newblock \emph{Signal Process. Image Commun.}, 78: 206--215.

\bibitem[{Wang et~al.(2020{\natexlab{c}})Wang, Xie, Zhao, and
  Meng}]{cvpr20_wang_rcdnet}
Wang, H.; Xie, Q.; Zhao, Q.; and Meng, D. 2020{\natexlab{c}}.
\newblock A Model-Driven Deep Neural Network for Single Image Rain Removal.
\newblock In \emph{{IEEE} {CVPR}}, 3100--3109.

\bibitem[{Wang et~al.(2019{\natexlab{b}})Wang, Yang, Xu, Chen, Zhang, and
  Lau}]{derain_2019_CVPR_spa}
Wang, T.; Yang, X.; Xu, K.; Chen, S.; Zhang, Q.; and Lau, R. W.~H.
  2019{\natexlab{b}}.
\newblock Spatial Attentive Single-Image Deraining With a High Quality Real
  Rain Dataset.
\newblock In \emph{{IEEE} {CVPR}}, 12270--12279.

\bibitem[{Wang et~al.(2004)Wang, Bovik, Sheikh, and Simoncelli}]{SSIM_wang}
Wang, Z.; Bovik, A.~C.; Sheikh, H.~R.; and Simoncelli, E.~P. 2004.
\newblock Image quality assessment: from error visibility to structural
  similarity.
\newblock \emph{{IEEE} TIP}, 13(4): 600--612.

\bibitem[{Wang et~al.(2021)Wang, Wang, Su, and Chen}]{icassp-wang-derain}
Wang, Z.; Wang, C.; Su, Z.; and Chen, J. 2021.
\newblock Dense Feature Pyramid Grids Network for Single Image Deraining.
\newblock In \emph{{IEEE} ICASSP}, 2025--2029.

\bibitem[{Wei et~al.(2019)Wei, Meng, Zhao, Xu, and Wu}]{Derain-cvpr19-semi}
Wei, W.; Meng, D.; Zhao, Q.; Xu, Z.; and Wu, Y. 2019.
\newblock Semi-Supervised Transfer Learning for Image Rain Removal.
\newblock In \emph{{IEEE} {CVPR}}, 3877--3886.

\bibitem[{Yang et~al.(2017)Yang, Tan, Feng, Liu, Guo, and
  Yan}]{derain_jorder_yang}
Yang, W.; Tan, R.~T.; Feng, J.; Liu, J.; Guo, Z.; and Yan, S. 2017.
\newblock Deep Joint Rain Detection and Removal from a Single Image.
\newblock In \emph{{IEEE} {CVPR}}, 1685--1694.

\bibitem[{Yasarla, Sindagi, and Patel(2020)}]{cvpr20_syn2real}
Yasarla, R.; Sindagi, V.~A.; and Patel, V.~M. 2020.
\newblock Syn2Real Transfer Learning for Image Deraining Using Gaussian
  Processes.
\newblock In \emph{{IEEE} {CVPR}}, 2723--2733.

\bibitem[{Zamir et~al.(2021)Zamir, Arora, Khan, Hayat, Khan, Yang, and
  Shao}]{derain_mprnet_cvpr21}
Zamir, S.~W.; Arora, A.; Khan, S.; Hayat, M.; Khan, F.~S.; Yang, M.; and Shao,
  L. 2021.
\newblock Multi-Stage Progressive Image Restoration.
\newblock In \emph{{IEEE} {CVPR}}.

\bibitem[{Zhang and Patel(2017)}]{derain_zhang_Sparse_and_Low-Rank}
Zhang, H.; and Patel, V.~M. 2017.
\newblock Convolutional Sparse and Low-Rank Coding-Based Rain Streak Removal.
\newblock In \emph{{IEEE} {WACV}}, 1259--1267.

\bibitem[{Zhang and Patel(2018)}]{derain_zhang_did}
Zhang, H.; and Patel, V.~M. 2018.
\newblock Density-Aware Single Image De-Raining Using a Multi-Stream Dense
  Network.
\newblock In \emph{{IEEE} {CVPR}}, 695--704.

\bibitem[{Zhu et~al.(2020)Zhu, Wang, Zhang, Su, and Zhao}]{icme2020_zhu_phy}
Zhu, H.; Wang, C.; Zhang, Y.; Su, Z.; and Zhao, G. 2020.
\newblock Physical Model Guided Deep Image Deraining.
\newblock In \emph{{IEEE} {ICME}}, 1--6.

\end{thebibliography}
\end{document}